\documentclass[10pt, letter, onecolumn]{arxiv}
\usepackage[utf8]{inputenc}
\usepackage{textgreek}
\usepackage{kantlipsum, lipsum}
\usepackage{dm-colors}
\usepackage{amsmath}
\usepackage{pstricks, pst-node}
\usepackage{verbatim}
\usepackage{multirow}
\usepackage{rotating}
\usepackage{scalerel}
\usepackage{booktabs}
\usepackage{enumitem}
\usepackage{xspace}
\usepackage{comment}
\usepackage[breakable]{tcolorbox}
\usepackage{bm}
\usepackage{bbm} 
\usepackage{mathtools}
\usepackage{soul}
\usepackage{epsfig}
\usepackage{graphicx}
\usepackage{amssymb}
\usepackage{colortbl}
\usepackage{csquotes}
\usepackage{setspace}
\usepackage{colortbl}
\usepackage{bbding}
\usepackage{threeparttable}
\usepackage{tabularx,ragged2e}
\usepackage{longtable}
\usepackage{placeins}
\usepackage{bbding}
\usepackage{algorithm}
\usepackage{algorithmic}
\usepackage{fancyvrb}
\definecolor{mainteal}{RGB}{0, 128, 128}
\definecolor{darkpastelgreen}{rgb}{0.13, 0.55, 0.13}
\definecolor{darkpastelred}{rgb}{0.55, 0.13, 0.13}
\usepackage[hang,flushmargin]{footmisc}

\usepackage{nameref}
\usepackage{varioref}
\usepackage{amssymb}%
\usepackage{lineno}
\usepackage{pifont}
\usepackage{xcolor}

\usepackage[pagebackref,breaklinks,colorlinks,
    citecolor=blue,
    filecolor=magenta,
    linkcolor=red,
    urlcolor=cyan]{hyperref}
    
\usepackage[noabbrev,capitalize]
{cleveref}
\usepackage{etoc}

\definecolor{reasoncolor}{HTML}{EE822F}
\definecolor{diagnosecolor}{HTML}{C81D31}
\definecolor{lookupcolor}{HTML}{1E5799}
\definecolor{matchcolor}{HTML}{3B82C4}
\definecolor{searchcolor}{HTML}{5BA3E0}
\graphicspath{{figures/}}
\definecolor{midgreen}{rgb}{0.0, 0.75, 0.0}
\definecolor{softblue}{HTML}{136783}

\newcommand{\ModelName}{ClinGeX}

\definecolor{goodgreen}{HTML}{00B050}
\definecolor{badred}{HTML}{FF0000}

\title{\Large{Advancing Clinical Genomic Analysis with a Skill-Centred Agent}}

\author[1,2]{Qiaoyu Zheng} 
\author[1,2]{Weike Zhao}
\author[3]{Yanjie Fan}
\author[4]{Ye Cao}
\author[5]{Mengqi Jiang}
\author[1]{Xiao Zhou}
\author[1]{Xiaoman Zhang}
\author[1]{Pengcheng Qiu}
\author[3]{Yongguo Yu}
\author[1]{Yanfeng Wang} 
\author[1,$\dag$]{Ya Zhang} 
\author[1,$\dag$]{Chaoyi Wu} 
\author[1,$\dag$]{Weidi Xie}
\affil[1]{\normalsize Shanghai Jiao Tong University, Shanghai, China \hspace{1cm}}

\affil[2]{\normalsize Shanghai AI Laboratory, Shanghai, China \authorcr \vspace{0.1cm}}

\affil[3]{\normalsize Xinhua Hospital affiliated to Shanghai Jiao Tong University School of Medicine, Shanghai, China \authorcr \vspace{0.1cm}}

\affil[4]{\normalsize Department of Obstetrics and Gynaecology, Faculty of Medicine, The Chinese University of Hong Kong \authorcr \vspace{0.1cm}}

\affil[5]{\normalsize Department of Radiation Oncology, Shanghai Chest Hospital, Shanghai Jiao Tong University School of Medicine, Shanghai, China \authorcr \vspace{0.1cm}}

\affil[$\dagger$]{\small
Corresponding authors. Email addresses:
\{three-world, ya\_zhang, wtzxxxwcy02, weidi\}@sjtu.edu.cn
}

\begin{document}

\begin{abstract}

Clinical genomic analysis requires experts to integrate evolving guidelines, databases, and computational evidence. As sequencing has become increasingly accessible, this complex and labor-intensive process has become a major bottleneck to the timely use of genomic information in clinical decision-making. To fill this gap, we developed \textbf{\ModelName{}}, a large-language-model agent that composes 105 curated executable skills through task-specific retrieval and dependency-aware orchestration. On 223 temporally held-out benchmark cases, ClinGeX achieved 71.7\% macro-averaged task-level concordance with expert reference classifications, compared with 45.7\% for the strongest comparator. In a three-condition, counterbalanced human-AI collaboration study, 45 clinical genomic professionals assessed 135 clinically derived cases under routine (no AI assistance), DeepSeek-assisted and ClinGeX-assisted conditions. ClinGeX assistance increased raw case-level concordance across criterion-level variant analysis, genomic interpretation in rare disease and molecular oncology decision-making (51.1\%, 57.8\%, 64.4\% versus 15.6\%, 31.1\%, 20.0\% in comparator conditions) and reduced reported interpretation time in all scenarios. During an exploratory one-month real-world non-interventional assessment, 11 experienced professionals submitted feedback on 96 self-selected analysis episodes and reported frequent conclusion adoption and time savings. These findings support executable domain skills as a promising approach to augment specialist genomic interpretation.

\end{abstract}

\maketitle


\section{INTRODUCTION}

\textit{Genomic analysis} has become an increasingly central component of modern precision medicine, particularly in rare disease and cancer~\cite{Riess2024Genomes,Smedley2021RareDisease,Zehir2017Cancer}. Advances in high-throughput sequencing have enabled the systematic characterization of genetic variation at unprecedented scale, facilitating molecular diagnosis, disease classification, patient stratification, and the identification of targeted therapeutic opportunities~\cite{Goodwin2016NGS,Ashley2016PrecisionMedicine,
Smedley2021RareDisease,Zehir2017Cancer}. Yet, as sequencing has become increasingly accessible, the major bottleneck in genomic medicine has shifted from data generation to the interpretation
of genetic variation and the derivation of clinically actionable insights~\cite{Muir2016ScalingComputation,Sboner2011RealCost,RiveraMunoz2018ClinGen,Richards2015ACMG}, a challenge famously framed as the ``\$1,000 genome, \$100,000 analysis''~\cite{Mardis2010Genome}.

This bottleneck arises from three challenges: (\textit{i}) \textbf{fragmented evidence}: relevant knowledge is distributed across heterogeneous and often siloed sources, including variant databases, biomedical literature, clinical guidelines, and patient phenotypes; (\textit{ii}) \textbf{expertise-intensive reasoning}: assessing variant significance demands substantial domain knowledge and contextual judgment involving disease mechanisms, inheritance patterns, phenotypic context, and evidence strength; and (\textit{iii}) \textbf{dynamic analytical workflows}: different patients and questions entail adaptive combinations of variant annotation, phenotype matching, evidence retrieval, pathogenicity assessment, and result synthesis. As a result, although conventional bioinformatics tools can automate individual steps~\cite{McLaren2016VEP,Smedley2015Exomiser}, performing a comprehensive genomic analysis still relies heavily on experts to integrate disparate information and synthesize actionable conclusions, limiting its scalability.

Recently, large language model~(LLM)-based agents have emerged as a promising paradigm for automating similar knowledge-intensive clinical tasks by combining reasoning with information retrieval and tool use~\cite{Chen2025LLMAgents,Goodell2025ClinicalTools}.  Bringing this paradigm to genomic analysis, however, is non-trivial. The field relies on a vast and continuously evolving ecosystem of professional guidelines, databases, and computational tools, each with distinct formats, interfaces, and usage protocols~\cite{Borchert2021KnowledgeBases,Jin2024GeneGPT}. Systematically translating these heterogeneous resources into standardized, agent-operable capabilities within a unified, scalable ``agent harness'' is therefore a central design problem~\cite{meng2026agentharness}. Crucially, such a harness must preserve the flexibility of LLMs to dynamically compose workflows for diverse clinical questions, while ensuring that individual analyses follow established domain-specific procedures and standards.

\begin{figure}[!t]
    \centering
    \includegraphics[width=1\linewidth]{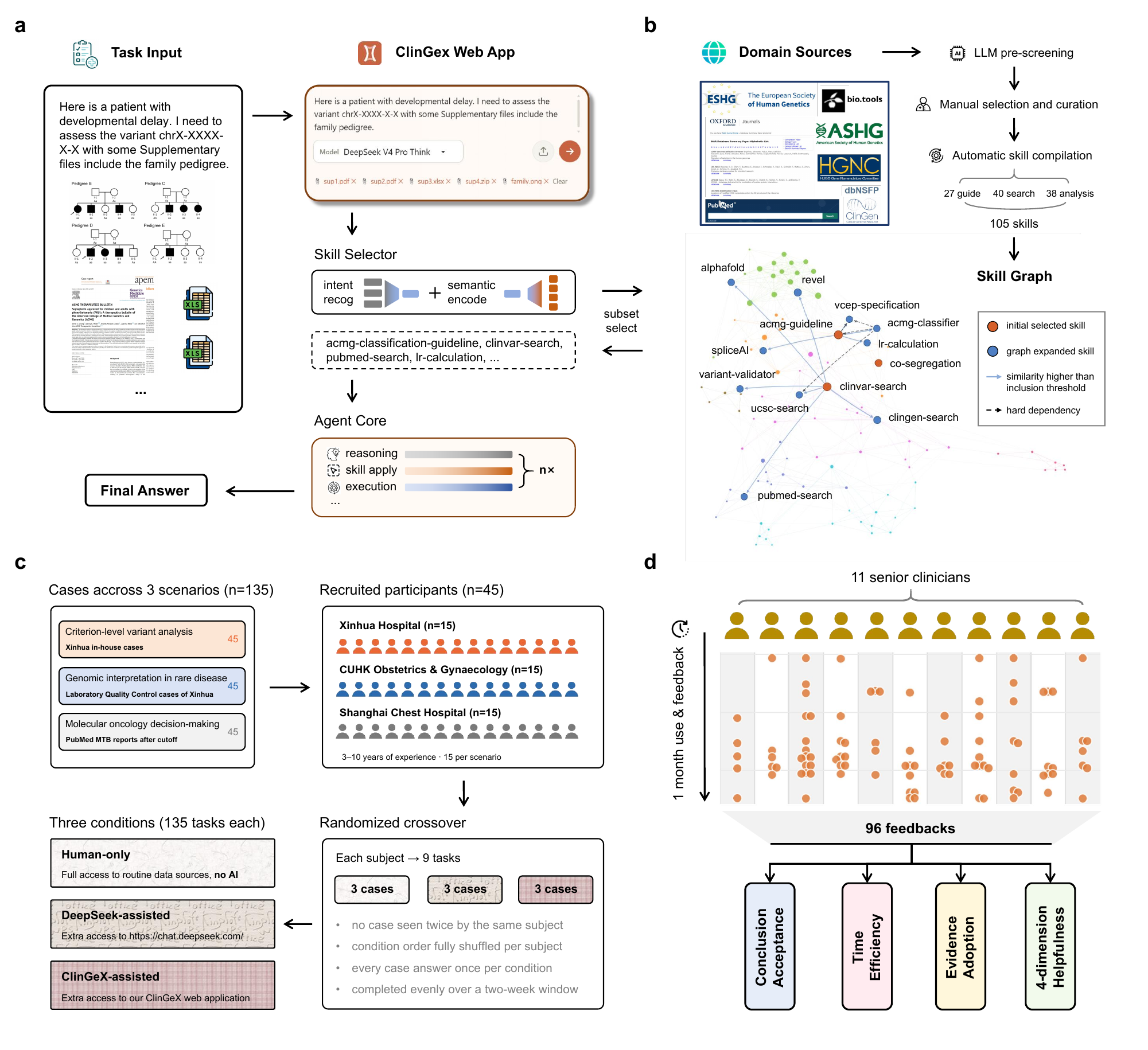}
    \caption{\textbf{Overview of \ModelName{} and its evaluation.}
    \textbf{a.} Task-oriented workflow. A clinical-genomics request and any accompanying text, image, PDF or other file types are submitted through the ClinGeX web application. The skill selector identifies the analytical needs, retrieves a task-specific subset of skills and their prerequisites, and passes them to the agent core for iterative planning, execution and evidence synthesis. 
    \textbf{b.} Construction of the skill library. Candidate resources were identified through a broad domain scan, prioritised by LLM-assisted screening, reviewed by clinical and domain experts, and compiled and tested as 105 executable skills: 27 guide, 40 search and 38 analysis skills. Skills are connected by functional-relatedness and directed prerequisite edges. 
    \textbf{c.} Controlled human-AI study. 45 readers evaluated 135 clinically derived cases across criterion-level variant analysis, genomic interpretation in rare disease and molecular oncology decision-making. 
    \textbf{d.} Exploratory one-month deployment, where 11 clinical-genomics professionals submitted structured feedback on 96 completed analysis episodes.}
    \label{fig:systemoverview}
\end{figure}


Here, we present \textbf{\ModelName{}}, a skill-centred agentic framework designed to support and accelerate clinical genomic analysis across heterogeneous inputs, including free text, pedigree images, clinical documents and sequencing data. \ModelName{} comprises \textbf{105 specialized skills}, each encoding an atomic operation in clinical genomics. These skills span five recurring categories: guideline following, entity mapping, data-source lookup, numerical calculation and model-based prediction. In the deployed system, they are grouped by operational role into \textbf{three functional classes}—guide, search and analysis. Together, they transform heterogeneous genomic resources into standardized capabilities that an LLM agent can combine for different clinical questions. 
Methodologically, we developed the skill library and system workflow through \textbf{six stages}: broad scanning of domain resources; LLM-assisted prioritization and expert selection; iterative skill development and testing; graph-based organization of related skills and their prerequisites; task-specific retrieval and progressive disclosure of skill instructions; and integration of multimodal inputs into an online interface. \textbf{At inference}, this organization enables the system to start from skills matched to a request, add their prerequisites and expose detailed instructions only when needed, limiting irrelevant context while preserving flexible workflows. This process converts established guidelines, databases and analytical tools into executable skills, while allowing \ModelName{} to retrieve and compose only the capabilities required for each analysis (Fig.~\ref{fig:systemoverview}b).


To evaluate both the performance and clinical utility of \ModelName{}, we conducted two independent studies. We assessed its autonomous analytical performance across 223 cases against OpenAI DeepResearch~\cite{OpenAI2025DeepResearch}, representing a state-of-the-art commercial general-purpose research system, and Biomni~\cite{Huang2026Biomni}, a leading open-source biomedical agent from academia. \ModelName{} substantially outperformed both systems, achieving 71.7\% concordance with expert consensus, 26.0 percentage points above the strongest comparator; moreover, 74.9\% of its reports were judged directly adoptable without substantial revision by senior physicians, exceeding both comparators by more than 40 percentage points. \textbf{More importantly}, to determine whether these capabilities could translate into meaningful support for clinicians, we conducted a randomized multicentre study involving \textbf{45} clinical genomic professionals from three medical centres and \textbf{135} retrospectively collected real-world cases spanning \textbf{criterion-level variant analysis}, \textbf{genomic interpretation in rare disease}, and \textbf{molecular oncology decision-making} (Fig.~\ref{fig:systemoverview}c). DeepSeek\footnote{Official website: \url{https://chat.deepseek.com/}}~\cite{Guo2025DeepSeekR1} was included as an accessible and representative example of widely used general-purpose LLM assistants. \ModelName{}-assisted clinicians achieved significantly higher case-level concordance across all three scenarios, reaching \textbf{51.1\%}, \textbf{57.8\%}, and \textbf{64.4\%}, compared with 22.2\%, 28.9\%, and 31.1\% for clinicians-only and 15.6\%, 31.1\%, and 20.0\% with DeepSeek assistance. \ModelName{} assistance also improved evidence identification and reduced the median time for criterion-level variant analysis \textbf{from 60 to 15 minutes}. Together, these findings support \ModelName{} as a potential ``Co-Clinician''~\cite{Schubert2025AIEducation,OSullivan2026ComplexCardiology}, augmenting rather than replacing expert genomic decision-making.

Finally, to preliminarily assess its utility in routine clinical-genomics work, we evaluated \ModelName{} over a prespecified one-month period of routine use, during which \textbf{11} experienced clinical genomics professionals from multiple medical centers used the system for self-directed, naturally occurring tasks and provided structured feedback on \textbf{96} completed analyses (Fig.~\ref{fig:systemoverview}d). Experts reported fully or partially adopting \ModelName{}'s conclusions in \textbf{94.8\%} of tasks, while time
savings of more than 15 minutes were reported for \textbf{77.3\%} of difficult tasks. Together, these complementary evaluations support the potential of a skill-centred, domain-grounded agent to improve the accuracy, efficiency, and practical scalability of clinical genomic analysis.

\section{Results}

\subsection{\ModelName{} Overview}

\ModelName{} is an LLM-driven agent for clinical genomic analysis. Through a chatbot interface (Fig.~\ref{fig:systemoverview}a), users can submit free-text requests together with optional multimodal materials, including pedigree images, clinical records and other file types. The system returns a free-text reports accompanied by a concise summary of the analytical workflow, evidence sources and key intermediate outputs to support traceability.

At its core, \ModelName{} uses a skill-centred architecture comprising 105 atomic clinical-genomics skills. For each request, a skill selector decomposes the task, retrieves relevant skills and expands the selection using semantic and prerequisite relationships encoded in a skill graph. The agent then executes the resulting task-specific workflow iteratively and synthesizes the accumulated evidence into its final response. Actions and tool outputs are retained in the execution record for review. Further details of the system architecture and development are provided in Methods Sec.~\ref{system_develop}.

We next evaluated \ModelName{} across three settings of increasing clinical realism: automated system benchmarks, a controlled human--AI collaboration study, and prospective assessment in real-world clinical-genomics use. Together, these evaluations examined standalone performance, effects on clinician performance, and perceived utility in routine workflows.

\begin{figure}[!h]
    \centering
    \includegraphics[width=1\linewidth]{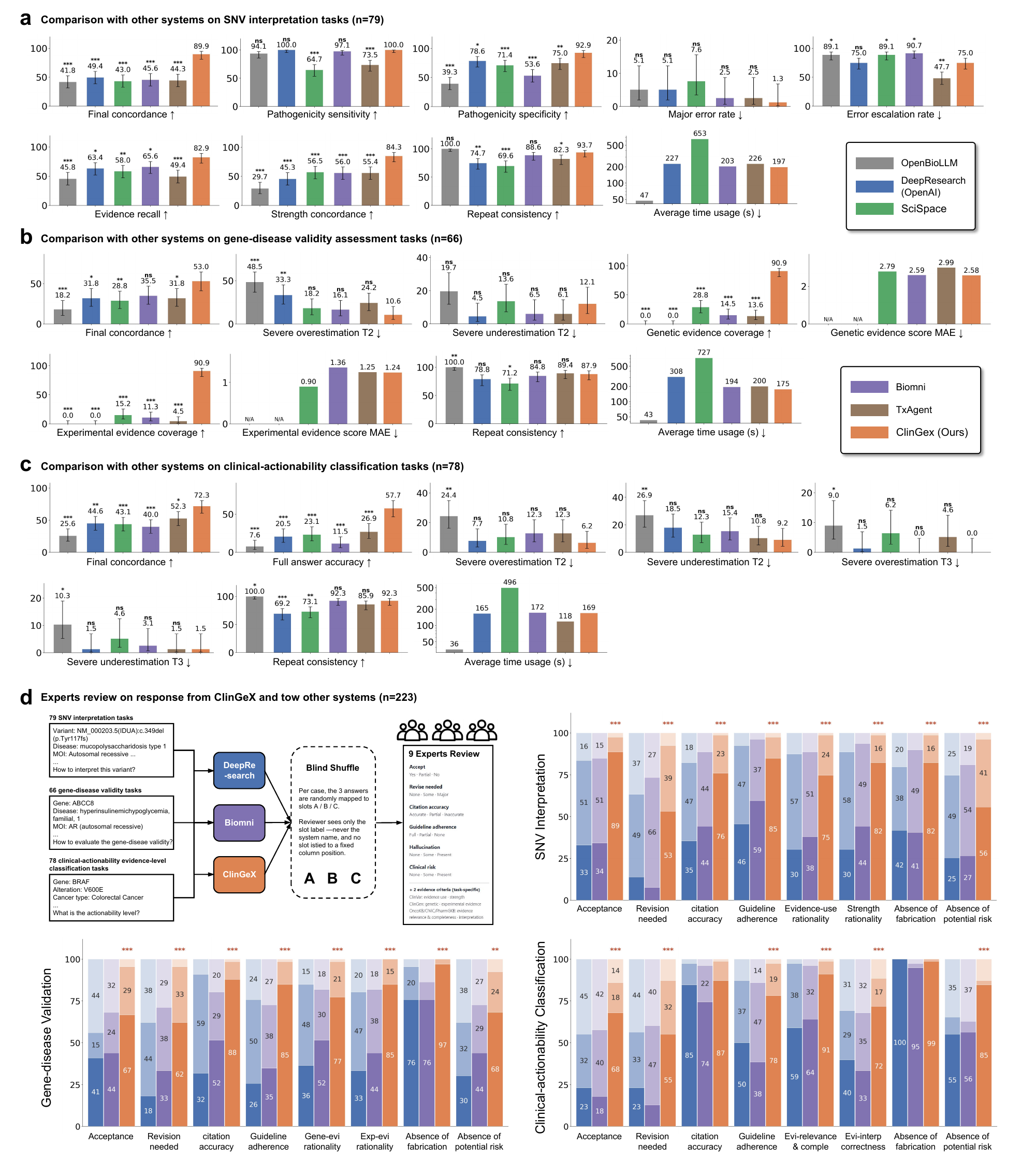}
    \caption{\textbf{System-level comparison of \ModelName{} with biomedical and general research systems.}
    \textbf{a}--\textbf{c}. Performance of OpenBioLLM, OpenAI DeepResearch, SciSpace, Biomni, TxAgent and ClinGeX on SNV interpretation (a; n = 79), gene-disease validity assessment (b; n = 66) and clinical-actionability classification (c; n = 78). Final concordance, repeat consistency and execution time are shown for all tasks; other measures are task specific. Arrows indicate the favourable direction and error bars denote 95\% confidence intervals. 
    \textbf{d}. Blinded assessment of the 669 reports generated by ClinGeX, Biomni and DeepResearch for the 223 benchmark cases. For each case, the three reports were de-identified, randomly labelled and rated on three-level ordinal dimensions by one of nine domain experts; each reviewer assessed all three systems for their assigned cases. Colours denote the three rating categories.}
    \label{fig:comparison}
\end{figure}

\subsection{System-level Analysis of \ModelName{} on Automated Benchmarks}

To evaluate \ModelName{} at scale, we constructed a benchmark comprising 223 clinical-genomics cases released after the knowledge cutoff of the underlying model of the comparators, collected from  five expert-curated resources: ClinVar~\cite{landrum2025clinvar} for SNV interpretation~(n=79), ClinGen~\cite{strande2017clingen,clingen2025} for gene--disease validity assessment~(n=66), and CIViC~\cite{griffith2017civic,krysiak2023civic}, OncoKB~\cite{chakravarty2017oncokb,suehnholz2024oncokb} and PharmGKB~\cite{whirlcarrillo2021pharmgkb} for clinical-actionability classification~(n=78). 
The primary metric was final-answer concordance, defined as agreement with all required components of the reference standard. Shared secondary outcomes included repeat consistency and execution time. Task-specific outcomes assessed pathogenicity discrimination and evidence recovery for SNV interpretation; classification severity and genetic and experimental evidence coverage for gene--disease validity; and complete-answer accuracy and classification severity for clinical actionability. Exact definitions, denominators and scoring procedures are provided in Supplementary Sec.~\ref{sec:automated_benchmark}. To assess potential leakage from the reference databases, we audited execution records for direct or indirect retrieval of source information and confirmed that no evaluated execution retrieved such information (Supplementary Table~\ref{tab:reference_source_audit}).

In Extended Data Fig.~\ref{fig:ablation} and Supplementary Secs.~\ref{sec:framework_selection} and \ref{sec:agent_core_selection}, we conducted ablation studies to assess the contribution of the \ModelName{} architecture and its robustness across different agent cores. Here, we used the final \ModelName{} configuration with DeepSeek-V4 as the agent core and compared it with the domain-specialized OpenBioLLM~\cite{OpenBioLLMs} and the latest agents, DeepResearch~(OpenAI)~\cite{OpenAI2025DeepResearch}, SciSpace~\cite{roy2024scispace}, Biomni~\cite{Huang2026Biomni} and TxAgent~\cite{gao2025txagent}. General-purpose LLMs such as GPT and DeepSeek were not benchmarked separately because they underlie the evaluated agents. Configurations are detailed in Supplementary Sec.~\ref{underlying_model}.

\textbf{\ModelName{} outperforms domain-specialized LLMs and agents across all three tasks.}

\ModelName{} achieved the highest final concordance in each task (Fig.~\ref{fig:comparison}a--c): 89.9\% for SNV interpretation, 53.0\% for gene--disease validity and 72.3\% for clinical actionability. The corresponding best baseline results were 49.4\% for DeepResearch, 35.5\% for Biomni and 52.3\% for TxAgent. Repeat consistency remained high across the three tasks (87.9--93.7\%). Mean execution time was 175--197,s, broadly comparable to other agentic systems and shorter than SciSpace (496--727,s).

For SNV interpretation, \ModelName{} achieved 100\% sensitivity and 92.4\% specificity for pathogenicity identification, with 82.3\% weighted evidence recall, 84.8\% evidence-strength concordance and a 1.3\% significant-error rate. In gene--disease validity assessment, it reported usable genetic and experimental evidence scores for 90.9\% of cases, compared with 28.8\% and 15.2\%, respectively, for the strongest comparator. For clinical-actionability classification, complete-answer accuracy was 57.7\%, compared with 26.9\% for TxAgent, while severe over- and under-classification remained infrequent.

\textbf{\ModelName{} produces higher-quality and more clinically usable reports in blinded expert review.}

We additionally compared the free-text reports produced by \ModelName{}, Biomni and DeepResearch with human rating, jointly considering the correctness of key results and the soundness of the reasoning. Within each task, cases were partitioned among three domain experts (nine experts in total), each of whom rated the three shuffled and de-identified system reports for their assigned cases. Reports were evaluated on three-level ordinal dimensions covering overall acceptability, required revision, citation accuracy, guideline adherence, evidence use and interpretation, fabricated content and potential clinical risk (Fig.~\ref{fig:comparison}d; Supplementary Sec.~\ref{sec:clinician_assessment}).

The proportion of \ModelName{} reports receiving the highest overall-acceptability rating was 88.6\% for SNV interpretation, 66.7\% for gene--disease validity and 67.9\% for clinical actionability. The corresponding proportions were 32.9\%, 40.9\% and 23.1\% for DeepResearch and 34.2\%, 43.9\% and 17.9\% for Biomni. \ModelName{} received higher ratings across all assessed dimensions in the first two tasks. In clinical-actionability classification, its clearest advantages were in overall acceptability, guideline adherence, evidence completeness and evidence interpretation, whereas citation accuracy and avoidance of fabrication were similar across systems. These findings suggest that the performance advantage of \ModelName{} extends beyond final classification to the quality and usability of its supporting analysis.

\textbf{\ModelName{} can ground its classifications in guideline-consistent analysis.}

\begin{figure}[!h]
    \centering
    \includegraphics[width=1\linewidth]{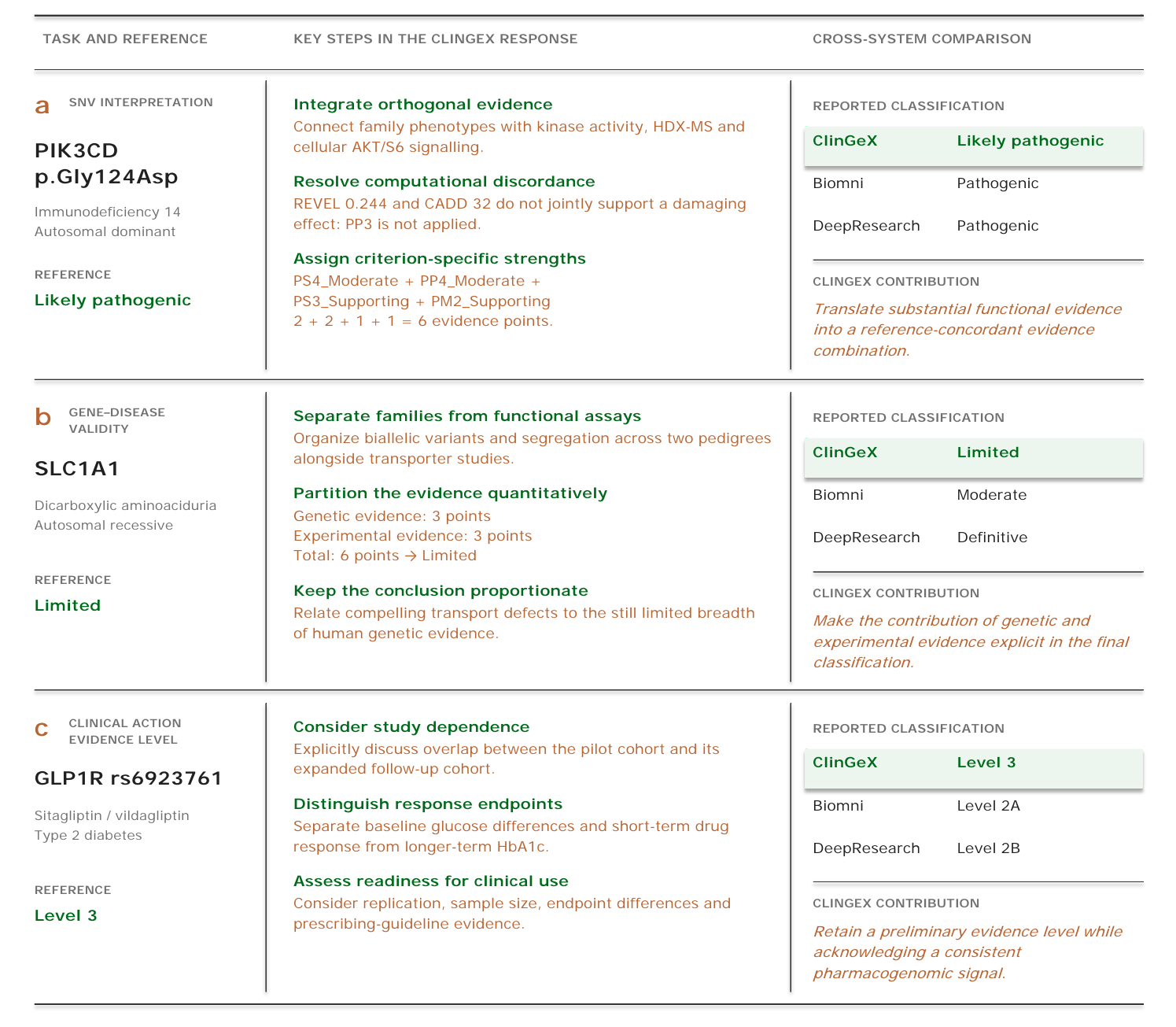}
    \caption{\textbf{Selected analytical contributions of \ModelName{} across three benchmark tasks.}
    \textbf{a}, Evidence-strength assignment for \textit{PIK3CD} p.Gly124Asp, integrating clinical and mechanistic findings with discordant computational predictions.
    \textbf{b}, Explicit partitioning of genetic and experimental evidence for the \textit{SLC1A1}--dicarboxylic aminoaciduria relationship.
    \textbf{c}, Consideration of study dependence, response endpoints and clinical applicability for the \textit{GLP1R} rs6923761--sitagliptin/vildagliptin association.
    The centre presents selected reasoning steps condensed from the recorded responses; the right compares the final classifications reported by \ModelName{}, Biomni and OpenAI DeepResearch. References are the benchmark classifications. Evidence-point totals summarize the recorded responses.}
    \label{fig:benchmark_case_studies}
\end{figure}

To more directly illustrate its capability, we selected three cases in which \ModelName{} matched the reference classification while Biomni and DeepResearch did not (Fig.~\ref{fig:benchmark_case_studies}). In a SNV interpretation case, \ModelName{} weighted two complementary functional studies against a discordant computational prediction without over-crediting it, reaching Likely Pathogenic where both comparators reported Pathogenic. In another gene-disease validity case, it partitioned genetic and experimental support into explicit, separately scored criteria, supporting a Limited classification where Biomni and DeepResearch reported Moderate and Definitive, respectively. In a clinical actionability case, it distinguished short-term glucose-response from longer-term HbA1c endpoints and weighed replication across overlapping cohorts, reaching Level~3 where the comparators reported Level~2A and Level~2B. In each case, the figure traces the classification to specific, guideline-grounded analytical steps that the comparators omitted, together illustrating \ModelName{}'s capacity for integrative synthesis of heterogeneous evidence and for discriminating among evidentiary sources of differing strength.

\subsection{Evaluation of \ModelName{} in the MRMC Human-AI Collaboration Study}

We next conducted a three-centre, multi-reader multi-case (MRMC) study to assess whether \ModelName{} could improve human performance in three clinical-genomics scenarios: criterion-level variant analysis, genomic interpretation in rare disease and molecular oncology decision-making. Each scenario comprised 45 cases, yielding 135 cases in total. Fifteen clinical genomic professionals with 3--10 years of relevant experience participated in each scenario (45 participants overall), with one scenario evaluated at each of three centres: Xinhua Hospital; the Department of Obstetrics and Gynaecology, Faculty of Medicine, CUHK; and Shanghai Chest Hospital. Participants completed cases under three conditions: human-only, DeepSeek-aided and \ModelName{}-aided. DeepSeek was selected as an accessible and widely used general-purpose LLM comparator.

\begin{figure}[!h]
    \centering
    \includegraphics[width=1\linewidth]{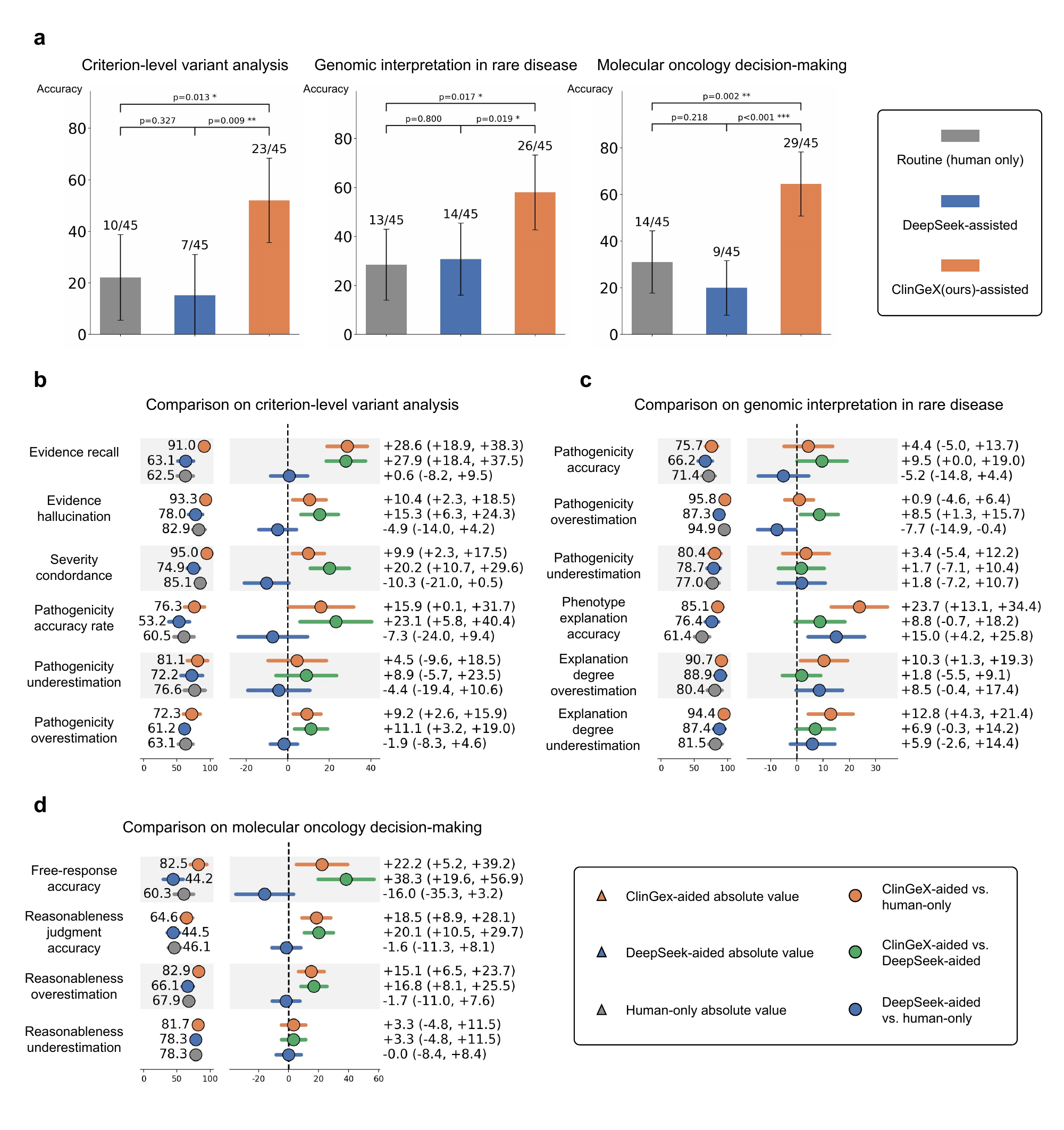}
    \caption{\textbf{Clinical genomic professional performance under unassisted, DeepSeek-assisted and ClinGeX-assisted conditions.}
    \textbf{a}. Case-level concordance with the expert reference standard for criterion-level variant analysis, genomic interpretation in rare disease and molecular oncology decision-making. Each condition comprised 45 assessments completed by 15 participants; fractions indicate observed concordant assessments over the total, whereas bars and error bars show model-standardized probabilities and two-sided 95\% Wald confidence intervals. Pairwise P values are Wald contrasts from crossed random-intercept logistic generalized linear mixed models (GLMMs), with condition as a fixed effect and reader and case as random intercepts, Holm-corrected across the three pairwise comparisons within each scenario. 
    \textbf{b}--\textbf{d}. 
    Task-specific secondary outcomes for criterion-level variant analysis (b), genomic interpretation in rare disease (c) and molecular oncology decision-making (d). Left, model-standardized favourable probabilities under each condition; right, corresponding absolute between-condition differences in percentage points with two-sided 95\% Wald confidence intervals. For outcomes where lower values indicate better performance, the left panels show one minus the model-standardized event probability and contrast directions are reversed.}
    \label{fig:mrmc_1}
\end{figure}

A Latin-square-inspired counterbalanced design minimized learning and order effects~\cite{bradley1958complete} (Extended Data Fig.~\ref{fig:design_extend}). Each participant completed nine cases---three per condition---without encountering the same case twice. Each case was evaluated under all three conditions by three different participants, and condition order was randomized across participants. Further details are provided in Methods Sec.~\ref{MRMC_method}.

\textbf{\ModelName{} assistance is associated with higher case-level concordance.}

We first assessed whether \ModelName{} could improve participants' final-answer correctness. We used case-level concordance as the primary metric, defined as the proportion of assessments for which the participant's final conclusion was adjudicated by the expert panel as substantially correct (Supplementary Sec.~\ref{main_concordance}).
\ModelName{} assistance produced the highest concordance in all three scenarios (Fig.~\ref{fig:mrmc_1}a). For criterion-level variant analysis, observed concordance was 51.1\%, compared with 22.2\% under human-only assessment and 15.6\% with DeepSeek assistance; the corresponding GLMM-standardized differences were 29.9 percentage points (Holm-adjusted \(P=0.013\)) and 36.8 points (\(P=0.009\)), respectively. For genomic interpretation in rare disease, observed concordance was 57.8\%, compared with 28.9\% and 31.1\%, respectively; the GLMM-standardized differences were 29.6 points (\(P=0.017\)) and 27.3 points (\(P=0.019\)). For molecular oncology decision-making, observed concordance was 64.4\%, compared with 31.1\% and 20.0\%, respectively; the GLMM-standardized differences were 33.4 points (\(P=0.002\)) and 44.5 points (\(P<0.001\)). Thus, under the current analysis, \ModelName{} assistance approximately doubled concordance relative to routine and DeepSeek-aided assessment.

\textbf{\ModelName{} assistance improves key components of the supporting analysis.}


We next evaluated task-specific intermediate outcomes, including evidence identification and weighting, phenotype interpretation and clinical-actionability assessment. Specifically, criterion-level variant analysis was evaluated using ACMG/AMP evidence recall, unsupported-evidence assignment, evidence-strength concordance, final-class accuracy and directional classification errors; genomic interpretation in rare disease using variant-classification and phenotype-interpretation accuracy and directional errors; and molecular oncology decision-making using panel-adjudicated free-response accuracy and the accuracy and directional errors of clinical-appropriateness judgements (Supplementary Sec.~\ref{secondary_outcome}).

In criterion-level variant analysis (Fig.~\ref{fig:mrmc_1}b), \ModelName{} increased model-standardized ACMG/AMP evidence recall by 28.6 (percentage) points relative to human-only assessment and by 27.9 points relative to DeepSeek assistance (both \(P<0.001\)). It reduced unsupported evidence assignments by 10.4 (\(P=0.009\)) and 15.3 points (\(P<0.001\)), respectively, and improved evidence-strength concordance by 9.9 (\(P=0.015\)) and 20.2 points (\(P<0.001\)). Final pathogenicity-classification accuracy was 23.1 points higher than with DeepSeek assistance (\(P=0.040\)), and over-classification was reduced by 9.2 and 11.1 points relative to human-only and DeepSeek-aided assessment, respectively (both \(P=0.034\)); the remaining directional contrasts were not statistically significant.

In genomic interpretation in rare disease (Fig.~\ref{fig:mrmc_1}c), model-standardized phenotype-interpretation accuracy increased by 23.7 (percentage) points relative to human-only assessment (\(P<0.001\)), accompanied by a 12.8-point reduction in under-interpretation (\(P=0.006\)). Accuracy was 8.8 points higher than with DeepSeek assistance, although this difference was not statistically significant (\(P=0.071\)). Improvements in final classification were also numerical rather than statistically significant.

In molecular oncology decision-making (Fig.~\ref{fig:mrmc_1}d), model-standardized free-response accuracy increased by 38.3 percentage points relative to DeepSeek assistance (\(P=0.001\)) and by 22.2 points relative to human-only assessment (\(P=0.033\)). Accuracy in judging the appropriateness of clinical decisions increased by 18.5 and 20.1 points relative to the human-only and DeepSeek-aided conditions, respectively (both \(P<0.001\)); overestimation of appropriateness decreased by 15.1 (\(P=0.001\)) and 16.8 points (\(P<0.001\)), respectively. These findings suggest that the observed gains extended to several intermediate components of evidence assessment and clinical interpretation.

\begin{figure}[!h]
    \centering
    \includegraphics[width=1\linewidth]{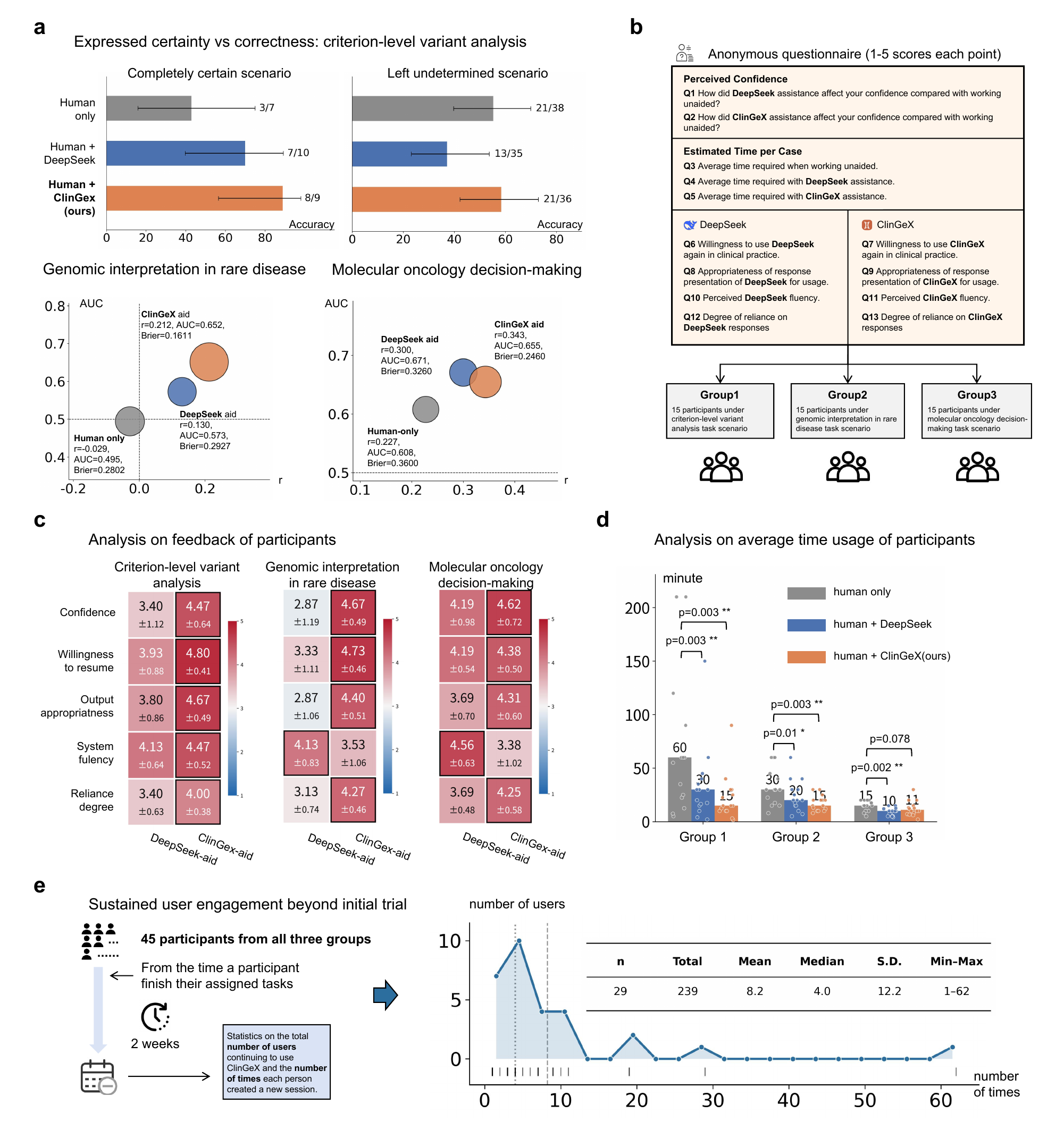}
    \caption{\textbf{Confidence--correctness alignment, user-reported experience and continued use in the MRMC study.}
    \textbf{a}. Top, final-classification accuracy in criterion-level variant analysis for assessments in which all ACMG/AMP criteria received an explicit call versus assessments with at least one criterion left undetermined; fractions indicate correct assessments over the total and error bars denote 95\% Wilson confidence intervals. Bottom, association between self-reported confidence and correctness for genomic interpretation in rare disease and molecular oncology decision-making, summarised by the point-biserial correlation (r), ROC-AUC and Brier score. 
    \textbf{b}. Design of the 13-item post-study questionnaire administered to the three participant groups (n = 15 per group).
    \textbf{c}. Mean\,$\pm$\,s.d. ratings for five experience dimensions under DeepSeek and ClinGeX assistance. 
    \textbf{d}. Participant-reported mean time per case; dots denote participants, numbers denote medians and P values are from paired two-sided Wilcoxon signed-rank tests with Holm correction. 
    \textbf{e}. New ClinGeX conversations initiated during the 14 days after completion of the assigned tasks, excluding the three study-related sessions.}
    \label{fig:mrmc_2}
\end{figure}

\textbf{Confidence--correctness alignment is preserved with \ModelName{} assistance.}

In Fig.~\ref{fig:mrmc_2}a, we examined whether improved concordance came at the cost of poorer uncertainty calibration. Certainty in criterion-level variant analysis was derived from whether participants made an explicit call for every applicable ACMG/AMP evidence item; the other two scenarios used self-reported confidence ratings. Metric definitions are provided in Supplementary Sec.~\ref{confidence_safe}.

For criterion-level variant analysis, an assessment was counted as fully certain when the participant assigned an explicit call to every applicable ACMG/AMP evidence item, and as partially certain otherwise; accuracy within a group is the proportion of its assessments adjudicated concordant. Of the 45 assessments per condition, 9 were fully certain under \ModelName{} assistance and 8 of these were correct (88.9\%), against 21 of the remaining 36 (58.3\%); the corresponding figures were 7 of 10 (70.0\%) under DeepSeek assistance and 3 of 7 (42.9\%) in the human-only condition. Certainty was thus most predictive of correctness under \ModelName{} assistance, although the small number of fully certain assessments limits formal comparison.

For genomic interpretation in rare disease and molecular oncology decision-making, confidence--correctness alignment was assessed using point-biserial correlation, ROC--AUC and the Brier score. \ModelName{} produced the highest correlation and lowest Brier score in both scenarios ((r=0.343), Brier score 0.246 for genomic interpretation in rare disease; (r=0.212), Brier score 0.161 for molecular oncology decision-making). It also produced the highest AUC in molecular oncology decision-making (0.652); its AUC in genomic interpretation in rare disease was 0.655, marginally below that of DeepSeek (0.671). Overall, the increase in concordance with \ModelName{} was not accompanied by poorer confidence--correctness alignment.


\textbf{Participants report a favorable experience and shorter completion times with \ModelName{}.}

Inspired by prior clinical AI evaluations~\cite{saab2026multimodal}, we also evaluated the system from readers' perspective. Specifically, after completing the test, participants rated their experience with DeepSeek and with \ModelName{} on an anonymous five-point Likert-type items covering confidence in the answers provided, willingness to use the system again, satisfaction with the presentation of results, smoothness of operation, and reliance on system output (Fig.~\ref{fig:mrmc_2}b); ratings were summarized separately for each scenario, by the 15 participants who had completed it in the MRMC study (\(n=15\) per scenario, 45 in total). More details are in Methods Sec.~\ref{questionaire}. \ModelName{} received higher mean ratings than DeepSeek on most dimensions in all three scenarios, the only exceptions being smoothness in genomic interpretation in rare disease and molecular oncology decision-making (Fig.~\ref{fig:mrmc_2}c).

We next compared self-reported mean time per question across the human-only, DeepSeek-aided and \ModelName{}-aided conditions (Fig.~\ref{fig:mrmc_2}d). Relative to human-only, \ModelName{} shortened the median time per question from 60 to 15 min for criterion-level variant analysis and from 30 to 15 min for genomic interpretation in rare disease; both differences were statistically significant (\(P=0.003\) for both). For molecular oncology decision-making, the median time decreased from 15 to 11 min, but the difference did not reach statistical significance (\(P=0.078\)). This smaller effect likely reflects the shorter human-only baseline in this task, together with the fact that participant-estimated times included system response latency.

As an exploratory analysis of spontaneous continued use, we recorded new conversations initiated in the two weeks following test completion, without any instruction or inducement for the 45 participants to continue using \ModelName{}. Twenty-nine participants initiated at least one further conversation beyond the three study-related sessions, generating 239 new conversations in total, with a mean of 8.2 additional conversations per continuing user (Fig.~\ref{fig:mrmc_2}e). These observations provide preliminary evidence of the system's acceptability and voluntary reuse beyond the controlled study setting.

\newpage
\subsection{Prospective Assessment of \ModelName{} on Real World Feedbacks}


Controlled evaluations establish performance on standardized cases, but do not capture usability across heterogeneous, self-directed clinical-genomics tasks. We therefore conducted an exploratory, prospective task-based assessment in which experienced professionals submitted self-selected gene- and variant-interpretation tasks over a prespecified one-month period. ``Prospective'' refers to the forward collection of these evaluation tasks and participant feedback. This was a non-interventional, research-only evaluation, and \ModelName{} outputs were not permitted to influence clinical reports, patient management or any other clinical decision.

\begin{figure}[p]
    \centering
    \includegraphics[width=1\linewidth]{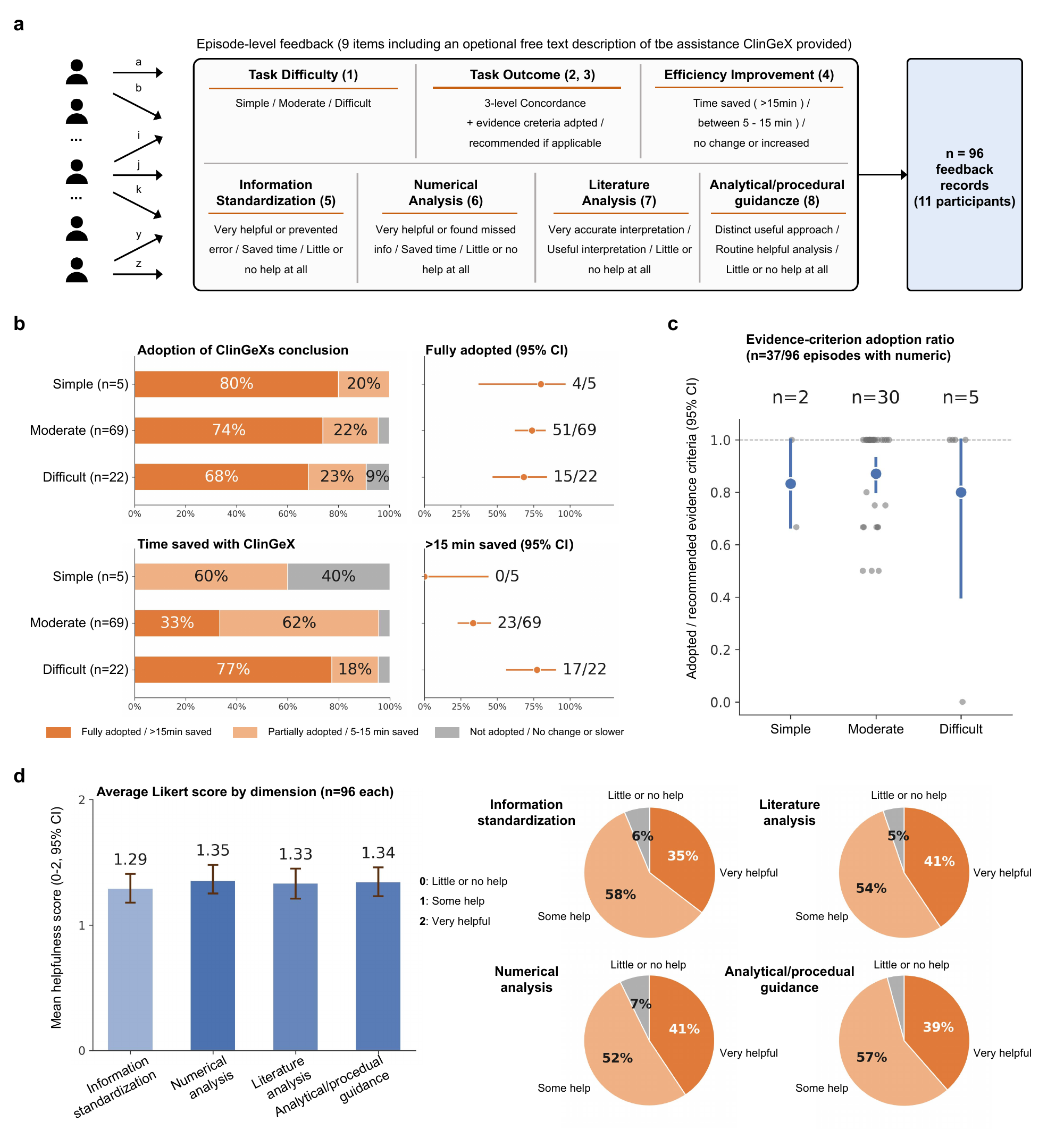}
    \caption{\textbf{Exploratory deployment of ClinGeX in routine clinical-genomics work.}
    \textbf{a}. Eleven experienced clinical-genomics professionals used ClinGeX for one month and submitted structured feedback on 96 completed analysis episodes, recording task difficulty, conclusion adoption, evidence-criterion adoption where applicable, reported time saved and helpfulness across four workflow dimensions. 
    \textbf{b}, Conclusion adoption and reported time saved, stratified by task difficulty. Stacked bars show all outcome categories; points and intervals show the proportion in the highest category (full adoption or >15 min saved) with 95\% confidence intervals. 
    \textbf{c}, Ratio of adopted to recommended evidence criteria among 37 episodes with countable recommendations; ratios were capped at 1. Grey points denote episodes and orange points and bars denote bootstrap means and 95\% confidence intervals. 
    \textbf{d}, Mean helpfulness scores and rating distributions for information standardisation, numerical analysis, literature analysis and analytical or procedural guidance. Helpfulness was rated from 0 (little or no help) to 2 (very helpful).}
    \label{fig:prospective}
\end{figure}

Eleven \textbf{senior clinical genomic professionals} from five medical centres used \ModelName{} in routine practice during a one-month period. 
The analysis here focused on clinicians' episode-level feedback. An analysis episode was defined as one complete clinical-genomics objective---for example, interpreting a patient's genomic sequencing findings or determining whether newly published evidence supported a specific variant- or gene--disease conclusion. An episode could comprise multiple dialogue turns, but its scope and analytical workflow were otherwise unrestricted. After completing an episode, the expert selected the corresponding conversation history and submitted one structured feedback record evaluating \ModelName{}'s contribution to the episode as a whole, rather than to an individual prompt or turn. Thus, one feedback submission represented one completed episode; 96 submissions were collected. These records were then used to examine when and how \ModelName{} helped in routine practice. For each episode, experts rated its difficulty (simple, moderate or difficult), indicated whether they adopted \ModelName{}'s conclusion (fully, partially or not) and, where applicable, how many recommended evidence criteria they adopted, reported the time saved, and rated the help provided in information standardization, numerical analysis, literature analysis, and analytical or procedural guidance from little or no help to very helpful (Fig.~\ref{fig:prospective}a). Full definitions of these measures are in Methods Sec.~\ref{deployment_method}.

\textbf{Experts frequently rate \ModelName{} conclusions as adoptable and report time savings.}

We analysed the reported contribution of \ModelName{} at two direct, episode-level endpoints: the extent to which its conclusion was adopted in the clinician's final assessment and whether its use accelerated completion of the analysis. Across the 96 feedback records, experts rated the conclusion as fully adoptable in 70 episodes (72.9\%), partially adopting it in 21 episodes (21.9\%), and not adopting  it in 5 episodes (5.2\%) (Fig.~\ref{fig:prospective}b). They reported saving more than 15 min in 40 episodes (41.7\%) and 5--15 min in another 50 episodes (52.1\%); only 6 tasks (6.3\%) took the same amount of time or longer. Full adoption was reported for 4/5 (80.0\%) simple, 51/69 (73.9\%) moderate and 15/22 (68.2\%) difficult episodes. The corresponding proportions for which more than 15 min was saved were 0/5, 23/69 (33.3\%) and 17/22 (77.3\%), respectively, showing that the reported time benefit was concentrated in more difficult analyses while most final conclusions were fully adopted.

\textbf{Experts adopt most of the evidence items recommended by \ModelName{}.}

As a more direct measure of assistance in routine variant interpretation, we identified 37 of the 96 feedback records that concerned variant-interpretation episodes for which \ModelName{} recommended discrete ACMG/AMP evidence codes and the senior expert recorded how many were ultimately adopted. For each episode, we calculated the ratio of adopted to recommended codes, capped at 1. The mean adoption ratio was 0.83 for simple episodes ($n=2$), 0.87 for moderate episodes ($n=30$), and 0.80 for difficult episodes ($n=5$) (Fig.~\ref{fig:prospective}c). Although the small numbers of simple and difficult episodes preclude comparisons between difficulty levels, these results show that experts retained a large proportion of the ACMG/AMP evidence codes proposed by \ModelName{} in their final variant interpretations.

\textbf{\ModelName{} helps experts across multiple dimensions in analysis workflow.}

Because \ModelName{} was designed around five recurring operations in clinical genomics---guideline following, entity mapping, data-source lookup, numerical calculation and model-based prediction---we next examined the specific aspects of analysis in which clinicians found the system helpful. Experts separately rated its contribution to information standardization, numerical analysis, literature analysis, and analytical or procedural guidance. These four clinician-facing dimensions captured overlapping manifestations of the five skill categories: information standardization primarily reflected entity mapping and structured data retrieval; numerical analysis reflected numerical calculation and model-based prediction; literature analysis reflected data-source lookup; and analytical or procedural guidance reflected guideline following and cross-skill orchestration. Across the 96 episodes, mean helpfulness scores on a 0--2 scale were similar across the four dimensions, at 1.29, 1.35, 1.33 and 1.34, respectively (Fig.~\ref{fig:prospective}d). Some or substantial help was reported for 90 episodes (93.8\%) in information standardization, 91 (94.8\%) in numerical analysis, 89 (92.7\%) in literature analysis, and 92 (95.8\%) in analytical or procedural guidance. Thus, clinicians reported that \ModelName{} supported multiple components of genomic interpretation rather than a single analytical function.

\textbf{Case studies illustrate how \ModelName{} contributes to clinical genomics analyses.}

\begin{figure}[!h]
    \centering
    \includegraphics[width=1\linewidth]{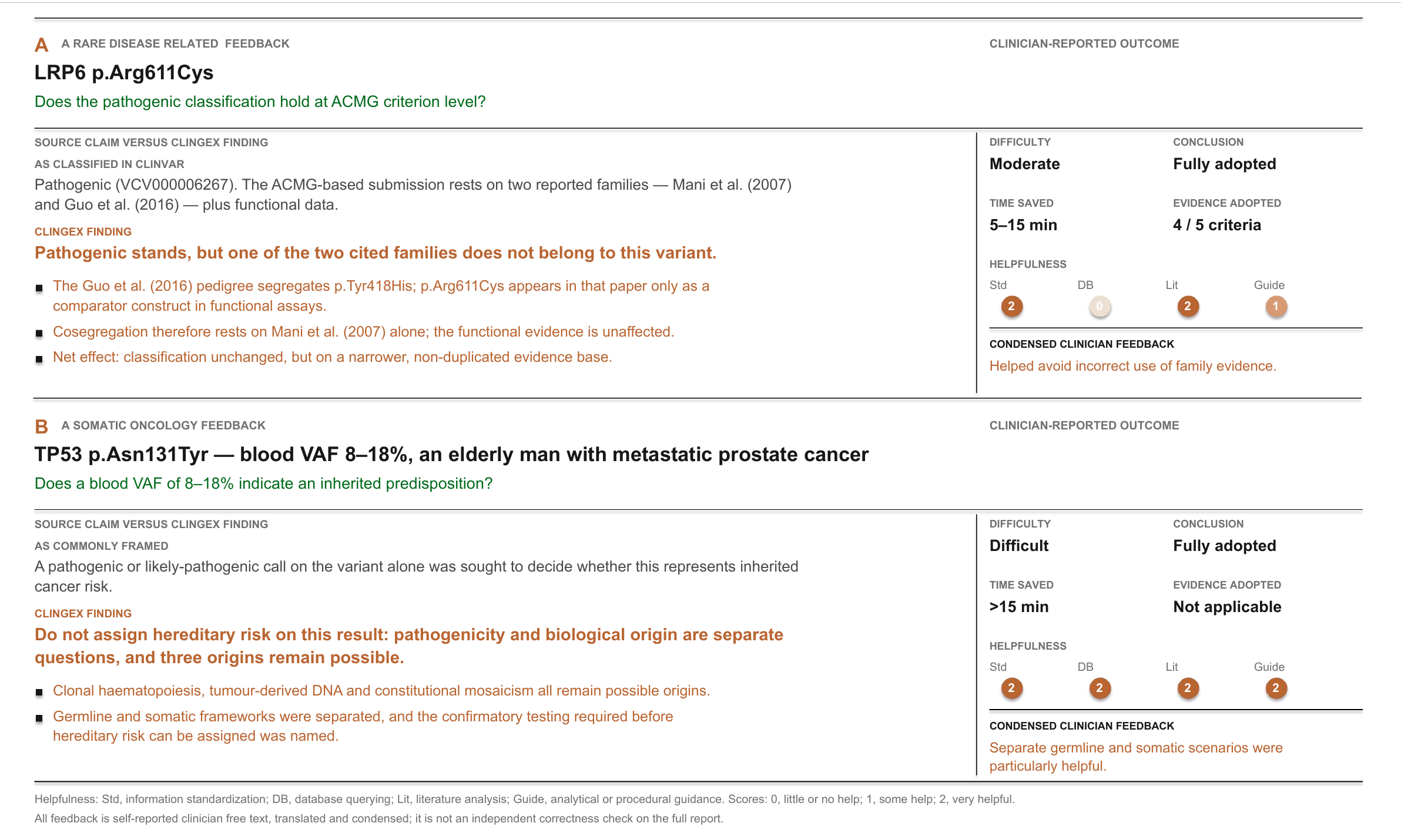}
    \caption{\textbf{Illustration and analysis of two feedback cases.}
    \textbf{a}, Evidence attribution for \textit{LRP6} p.Arg611Cys. \ModelName{} identified that a family cited in a ClinVar submission carried p.Tyr418His, helping the clinician avoid using misattributed family evidence. The cited paper also contained p.Arg611Cys functional data; the mismatch concerned the family evidence.
    \textbf{b}, Interpretation of a low-allele-fraction \textit{TP53} p.Asn131Tyr finding in blood from an older patient with metastatic prostate cancer. \ModelName{} helped distinguish possible variant origins, separate germline and somatic interpretations, and identify the confirmation needed before assigning hereditary risk; the origin remained unconfirmed in the available record.
    For each episode, the left and centre show the clinical context, key judgement and contribution to the clinician's interpretation. The right-hand column reports the submitting clinician's difficulty rating, conclusion adoption, time saved, evidence-criterion adoption where reported, helpfulness ratings and condensed feedback. Helpfulness was rated from 0 (little or no help) to 2 (very helpful).}
    \label{fig:case_study}
\end{figure}

Two cases from the prospective deployment illustrate how \ModelName{} helped clinicians address ambiguities with consequences for evidence use and clinical interpretation (Fig.~\ref{fig:case_study}). In an assessment of \textit{LRP6} c.1831C$>$T (p.Arg611Cys), reported as moderately difficult by the submitting clinician, a ClinVar submission cited a family carrying p.Tyr418His as evidence for the assessed variant. \ModelName{} identified this mismatch, helping the clinician avoid attributing that family's evidence to p.Arg611Cys. The clinician reported full adoption of the conclusion, adoption of four of five recommended evidence criteria, and 5--15\,min saved, and explicitly noted that the analysis helped avoid incorrect use of family evidence. In a second case, reported as difficult by the submitting clinician, \textit{TP53} c.391A$>$T (p.Asn131Tyr) was detected at a variant allele fraction of 8--18\% in blood from an old man with metastatic prostate cancer. The key question was whether this finding supported a hereditary cancer-risk interpretation. \ModelName{} distinguished possible origins, including clonal haematopoiesis, tumour-derived DNA and constitutional mosaicism, separated germline and somatic interpretation frameworks, and proposed confirmation of sample type and additional tissue testing before assigning hereditary risk. The clinician reported full adoption of the conclusion and more than 15\,min saved, highlighting the separate germline and somatic scenarios as particularly helpful.

\clearpage
\section{DISCUSSION}

Genomic analysis increasingly informs multiple stages of clinical care, from understanding disease mechanisms and establishing molecular diagnoses to guiding targeted interventions and population-level screening. Its broader adoption, however, is constrained by the specialized expertise required, uneven access to specialist support across centres and the difficulty of translating rapidly evolving knowledge into clinical practice. Contemporary artificial intelligence, particularly large language model (LLM)-based agentic systems, has shown promise in supporting clinical decision-making across several medical settings. Realizing this potential in clinical genomics requires systems that can reliably integrate fragmented and continuously evolving guidelines, massive databases and computational tools. Yet most evaluations remain benchmark- or simulation-based, leaving limited evidence of their effects on clinician performance and usability in workflow-relevant settings. 

The central contribution of this work is \ModelName{}, \textbf{a skill-centred agent that adapts general-purpose LLMs into clinically grounded assistants for genomic analysis}. Its technical value lies in three respects. \textbf{First}, systematic resource collection and expert curation help reduce the omissions and selection biases associated with ad hoc resource enumeration, and may offer a template for analogous skill discovery in other clinical domains. \textbf{Second}, semi-automated skill construction and graph-based organization allow the skill library to expand without sacrificing efficient task-specific retrieval. \textbf{Third}, progressive disclosure restricts the active context to what each analytical step requires, helping constrain hallucination. The value of these design choices is supported by the automated benchmark, in which the resulting system outperformed both alternative enhancement strategies and SOTA general-purpose and biomedical agents.

Across the automated benchmarks, human--AI collaboration study, and prospective deployment, we demonstrate the potential practical value of \ModelName{} in five complementary aspects.
\textbf{First}, it constrains the hallucination that current LLM-based systems cannot reliably avoid, grounding conclusions in executed skills and the evidence they return; the risk this addresses is visible in the DeepSeek-aided arm, where concordance fell below that of unassisted clinicians in two of the three scenarios (15.6\% versus 22.2\%; 20.0\% versus 31.1\%). \textbf{Second}, it improves the quality of gene-related clinical analysis and decision-making by junior and middle-grade clinicians, raising case-level concordance with expert consensus from 22.2\%, 28.9\% and 31.1\% to 51.1\%, 57.8\% and 64.4\% across the three scenarios---a meaningful step towards democratizing access to expert-level interpretation. \textbf{Third}, it lowers interpretive burden, shortening the median time per case from 60 to 15 min for criterion-level variant analysis and from 30 to 15 min for genomic interpretation in rare disease, while presenting the complete interpretive path so that clinicians can inspect it and decide whether to adopt or override the system's conclusion. \textbf{Fourth}, it showed broad workflow-level utility for experienced clinical-genomics professionals during self-directed routine use, as reflected by reported benefits spanning information retrieval and standardization, numerical and literature analysis, and procedural guidance. \textbf{Fifth}, the system can be deployed locally and adapts to a range of open- and closed-source backbone models, which accommodates the differing infrastructural constraints and access policies of individual medical centres.

\textbf{Limitations}

First of all, although \ModelName{} targets clinical genomic analysis broadly, the MRMC study covered only three scenarios---criterion-level variant analysis, genomic interpretation in rare disease and molecular oncology decision-making---selected based on the participating centres' specialties and routine caseloads. Although the skill library was not tuned to these scenarios, their evaluation does not establish generalizability to other areas of clinical genomic interpretation, such as polygenic risk assessment.


Secondly, the MRMC study was powered to detect the main effect of assistance; with 45 readers and 135 cases, several secondary comparisons showed consistent directions but did not all reach statistical significance. The primary endpoint is a proxy: concordance with expert consensus reflects interpretive quality rather than patient outcomes. Although fit for the present aim of assessing interpretive performance, establishing that improved interpretation translates into changes in clinical management and supports broader clinical use will require a prospective randomized trial with prespecified clinical endpoints. 

Finally, the prospective deployment focused on workflow-level outcomes—including conclusion adoption, time savings and perceived helpfulness—rather than prespecified clinical endpoints such as diagnostic yield, changes in management or patient outcomes. It therefore provides evidence of practical utility in routine analysis, while the downstream clinical impact of \ModelName{} requires evaluation in larger prospective studies.



\clearpage
\section{METHODS}

\subsection{Development of \ModelName{}}
\label{system_develop}

\ModelName{} adopts a skill-centred agentic architecture. An agent `skill'~\cite{anthropic2025agentskills} is a reusable instruction module, written in structured free text, that encapsulates a task-specific agentic workflow for a particular clinical or research task. Each skill specifies when it should be invoked and provides the procedural knowledge required to complete the task, including what information to seek, which standards and resources to consult, which tools to call, and how to carry out the analysis step by step~(a detailed description on skills can be found in Supplementary Sec.~\ref{skill_define}). 
By defining these capabilities at an atomic level, a broad and systematically curated skill library allows the agent to flexibly select, combine, and sequence individual skills to construct task-specific workflows, thereby supporting a wide range of clinical genomic analysis needs.

To build the skill library and integrate it into an LLM-based agent, we developed \ModelName{} through a six-stage design framework. Stages 1--4 focused on constructing the skill library by identifying, curating, and formalizing domain knowledge and analytical resources into atomic agent skills. Stage 5 integrated these skills into the agent workflow through inference-time skill retrieval and orchestration, while Stage 6 focused on user interface and interaction design.


\textbf{Stage 1. Web-scale scan of domain resources.}

To systematically build the skill library, we first conducted a web-scale scan to identify and collect a broad range of clinical-genomics resources for subsequent skill construction. 
Specifically, we organized the scan into five broad categories of clinical-genomic analysis operations: guideline following, entity mapping, data-source lookup, numerical calculation, and model-based prediction. 
Together, these categories capture common needs in genomic analysis, including applying clinical guidelines, converting heterogeneous entity identifiers, querying case- or cohort-level evidence from genomic and clinical databases, performing standardized calculations, and predicting the functional effects of genomic variants.

Because resources in these five categories are organized differently across the web, we did not use a shared search strategy for all categories. 
Instead, for each category, we first selected an existing broad-coverage collection as the primary source for automated resource extraction, and then manually added other important resources that were not captured.
For example, to identify resources for data-source lookup, we extracted all 2,594 database entries from the cumulative alphabetical index of the \textit{Nucleic Acids Research} Database Issue rather than searching it with a limited set of keywords. Because no single catalogue covers the full range of resources used in clinical genomics, we supplemented these entries with records from bio.tools, re3data.org and FAIRsharing.org and then merged records that referred to the same resource. The other four groups were scanned using the same general principle but with sources suited to each type of resource; the complete sources, search procedures and coverage limits are provided in Supplementary Sec.~\ref{sec:supp_stage1_scanning} and all considered websites are listed in Supplementary Table~\ref{tab:stage1_flow}.

This stage yielded 195 guidelines, 3,508 entity-mapping tools and services, 41,000 databases and resources for evidence lookup, 53 reproducible calculation methods, and 144 prediction models, forming the initial pool of candidate resources for agent skill construction. Notably, each candidate was represented by its associated free-text documentation; for example, tools, databases, and models were represented by their `README' files, documentation pages, model cards, or related methodological descriptions.

\textbf{Stage 2. LLM-assisted resource filtering and expert review.}

To further improve the quality of the candidate resources identified in Stage 1, we filtered them in Stage 2 based on their clinical relevance, reliability, and suitability for skill construction. We adopted an LLM--human hybrid filtering pipeline, in which an LLM first evaluated the candidate resources using category-specific criteria, followed by final review and selection by clinical experts.

Similar as Stage 1, the LLM rating pipeline was tailored to each resource category. For each category, we predefined a set of evaluation rubrics and prompted the LLM to assess each candidate against each rubric separately. For the entity mapping, data-source lookup, numerical calculation, and model-based prediction categories, the rubrics were quantitative, with the LLM assigning a numerical score to each rubric (e.g., 0--5). These rubric-level scores were then combined as a weighted sum to produce an overall score from 0 to 100. Based on this score, candidates were assigned to Tier~1 ($\geq75$), Tier~2 (55--74), Tier~3 (35--54), or exclusion ($<35$), with higher-tier candidates prioritized for subsequent review. In contrast, guideline-following candidates were assessed qualitatively: the LLM provided detailed comments on each predefined rubric without assigning numerical scores or tier levels. We designed this mainly because guidelines are typically high-value resources that warrant careful consideration. Providing a single numerical score or tier could lead subsequent reviewers to rely on that summary judgment. We therefore retained free-text comments, requiring experts to examine the underlying rationale and reach more considered conclusions. In general, the rubrics focused on the authority, currency, verifiability, and accessibility of guidelines; the clinical relevance, reliability, and accessibility of data and mapping resources; and the reproducibility, clinical support, and practical deployability of computational methods and predictive models. The complete rubrics, scoring instructions, and weights for each operation type are provided in Supplementary Table~\ref{tab:stage2_triage}.

Eight clinical experts from seven institutions then reviewed the candidates for clinical utility, technical reliability, and functional redundancy, with the LLM-generated assessments provided as supporting material. Importantly, the LLM-derived tiers were used only to prioritize candidates for expert review and did not serve as strict inclusion thresholds. Final inclusion was determined by consensus among the eight clinicians, and lower-tier candidates could still be retained if they addressed important clinical needs not adequately covered by higher-ranked resources. More details are provided in Supplementary Sec.~\ref{sec:stage2_llm_triage}.

After filtering, 105 candidate resource documents remained for skill construction: 27 for guideline following, 13 for entity mapping, 30 for data-source lookup, 12 for numerical calculation, and 23 for model-based prediction.


\textbf{Stage 3. Skill curation through recursive optimization.}


In this stage, we translated the 105 selected resources into standardized, executable agent skills and iteratively refined them through a builder--evaluator loop inspired by recursive self-improvement~\cite{novikov2025alphaevolve,anthropic2026recursive}. We first consolidated the five resource categories described above into three functional skill classes: \texttt{guide}, \texttt{search}, and \texttt{analysis}. \texttt{Guide} skills were derived from clinical guidelines and other normative documents; \texttt{search} skills from resources for entity mapping and data-source lookup, including databases, search services, and their APIs; and \texttt{analysis} skills from numerical methods and predictive machine-learning models.

Each skill was initially compiled by a `builder' agent based on DeepSeek-V4. The builder received the corresponding source materials together with a class-specific construction prompt and generated an initial skill implementation. For \texttt{guide} skills, it translated source documents into stepwise procedures and traceable decision rules. For \texttt{search} skills, it specified how to query the target resource, including accepted inputs, query procedures, and processing of returned records. For \texttt{analysis} skills, it implemented the required calculations or specified how predictive models should be invoked and their outputs interpreted. Across all three classes, each skill was represented by a standardized \texttt{SKILL.md} file defining its scope, inputs, execution workflow, outputs, and limitations, accompanied, where necessary, by executable scripts and supporting references. The complete construction prompts and skill specification are provided in Supplementary Sec.~\ref{sec:skill_compilation}.

To verify and iteratively improve each compiled skill, we employed a separate `evaluator' agent. Based on the source materials and the initial skill implementation, the evaluator automatically generated 5--20 skill-specific test cases, each specifying an input, the expected action or procedure, and the expected output or behaviour (representative examples are provided in Supplementary Table~\ref{tab:revel_skill_tests}). The resulting test cases were then executed by a separate execution agent, instantiated as an LLM equipped with the corresponding skill. For each failed case, the evaluator summarized the observed failure in free text and provided the corresponding test output to the builder. The builder then revised the relevant skill instructions, interface specification, or executable code as appropriate, and the updated skill was subsequently re-evaluated using the same procedure. This builder--evaluation--revision cycle continued until all identified failures were resolved or a maximum of five revision cycles was reached, resulting in 105 agent skills. The workflow is illustrated in Extended Data Fig.~\ref{fig:compile_extend}.


\textbf{Stage 4. Graph construction for scalable skill organization.}

After obtaining the final 105 skills, we organized them into a skill graph to support more efficient skill retrieval and combination. In the graph, each skill was represented as a node, and we exhaustively considered all possible pairwise relationships among the 105 skills, yielding $\binom{105}{2}=5,460$ unordered skill pairs. Each pair was then evaluated for two types of relationships: functional relatedness and prerequisite dependence.
For each skill pair, an LLM first assessed functional relatedness along five dimensions---purpose similarity, input overlap, substitutability, complementarity, and joint use---assigning each a score from 0 to 4. These scores were summed to form an undirected relatedness edge weight ranging from 0 to 20. Separately, a directed prerequisite edge was added when one skill required another to obtain or interpret its output. The resulting graph contained 28 directed prerequisite edges. Relatedness weights support the retrieval of complementary skills, whereas prerequisite edges preserve required workflow steps. New skills can be incorporated by evaluating only their relationships with existing nodes. Further details are provided in Supplementary Sec.~\ref{sec:supp_clingex_skill_graph}.

\textbf{Stage 5. Context management based on progressive disclosure.}

Till here, Stages 1–4 established the core of ClinGeX: a skill library organized as a skill graph. During inference, this graph is integrated into the host LLM's context through task-specific retrieval and progressive plan–execute orchestration.

Given a task query, \ModelName{} first invokes an LLM-based intent-recognition agent (Extended Data Fig.~\ref{fig:selection_extend}). The agent decomposes the user's request into one to three concrete analytical needs. Each need is encoded using Qwen \texttt{text-embedding-v4}~\cite{zhang2025qwen3embedding}, and cosine similarity is computed against the purpose and applicability descriptions of all 105 skills. The 30 highest-ranked skills form the seed set. For each seed, one-hop related skills with an edge weight of $R\geq8$ are considered, and at most 15 additional skills are retained across all seeds. Three general literature- and web-search skills are included by default, after which prerequisite dependencies are recursively added until closure.

After retrieval, \ModelName{} invokes a host agent that loads concise descriptions of all selected skills, each summarizing the skill's capabilities, usage boundaries, and routing rules, and uses them to formulate an overall workflow plan. The host agent then instantiates sub-agents for individual workflow steps, each of which loads the complete instructions of the corresponding skill into its context and executes the assigned task according to the skill-specific workflow. More details of this stage are provided in Supplementary Sec.~\ref{skill_retrieval}.

\textbf{Stage 6. Multimodal support and online deployment.}

In the final stage, we implemented \ModelName{} as a multimodal online system. Uploaded files were processed in parallel according to type. PDFs were directly converted into Markdown-formatted text using MinerU~\cite{wang2024mineru}, whereas images, such as pedigree charts, were passed to Qwen3-VL~\cite{bai2025qwen3vl} with a fixed instruction to generate descriptive summaries. The resulting texts were concatenated with the user's query to form the input context.
 
We deployed \ModelName{} as a web application supporting multiple backbone models (Qwen, DeepSeek, Gemini, GLM, GPT, Kimi, etc.), heterogeneous file uploads, and concurrent queries. As shown in Extended Data Fig.~\ref{fig:operation}, the interface contains four components: a dialogue module for user input and file submission, an execution module that streams intermediate operations and tool results, a result module that presents the final structured analysis, and a user-management module for reviewing previous sessions and conducting follow-up analyses. 

\subsection{Conduct of the MRMC Human-AI Collaboration Study}
\label{MRMC_method}

We conducted a randomized, three-condition human--AI collaboration study following established MRMC design principles~\cite{Obuchowski2022MRMC}, with condition order and case-set assignment counterbalanced using a Latin-square-like allocation~\cite{Zeelenberg2015Latin}, and administered the study through an online evaluation platform~(Supplementary Sec.~\ref{mrmc_platform}).

\subsubsection{Task Scenarios}

We evaluated \ModelName{} in three clinical genomic analysis scenarios, selected to cover criterion-level variant analysis, genomic interpretation in rare disease, and molecular oncology decision-making, detailed as:
\begin{itemize}
  \setlength\itemsep{0.5em}
    \item \textit{ACMG-based variant pathogenicity classification and analysis (criterion-level variant analysis)}
Participants evaluated the applicability and evidence strength of each of the 28 ACMG/AMP criteria for a given variant and assigned a final pathogenicity classification. They also provided a written justification for their assessment, along with a confidence rating and the time taken to complete the task. This scenario primarily assessed participants' ability to evaluate pathogenicity evidence according to the ACMG/AMP framework and integrate the applicable criteria into a final variant classification.
    \item \textit{Genomic interpretation in rare disease through integrated variant and phenotype assessment (genomic interpretation in rare disease).}
Participants evaluated whether the candidate genes and variants could explain the patient's phenotype, integrating variant-level evidence with phenotype concordance and inheritance information. They provided an integrated case-level interpretation, together with a written justification, confidence rating, and completion time. This scenario primarily assessed participants' ability to synthesize genetic and phenotypic evidence to determine the extent to which a candidate finding explained the patient's disease.
    \item \textit{Molecular tumor analysis and action recommendation (molecular oncology decision-making).}
Participants reviewed the clinical and molecular information for a tumour case, proposed the most appropriate next analytical or clinical action, and evaluated the plausibility of a series of case-specific statements. They also reported their confidence and completion time. This scenario assessed the integration of molecular findings with clinical context, the identification of actionable evidence, and the distinction between supported and unsupported clinical recommendations.

\end{itemize}
Representative cases and complete task examples are provided in Supplementary Sec.~\ref{mrmc_example}.

\subsubsection{Tested Conditions}

Each participant completed tasks under the following three assistance conditions:
\begin{itemize}
    \setlength\itemsep{0.5em}
    \item \textit{Routine condition.} Participants completed the assigned tasks according to their routine clinical practice. They were allowed to access the internet and freely consult any literature, databases, or online resources that they would normally use in their routine workflow, but were not permitted to use any AI-based tools.

    \item \textit{DeepSeek-assisted condition.} Participants had access to the same resources permitted in the unassisted condition and could additionally use DeepSeek through its official web interface.

    \item \textit{\ModelName{}-assisted condition.} Participants had access to the same resources permitted in the unassisted condition and could additionally use the deployed \ModelName{} platform.
\end{itemize}

The assigned system was the only AI tool permitted in each assisted condition, and all other study procedures were kept consistent across conditions.

\subsubsection{Data Preparation}

In this section, we describe the evaluation data preparation procedures for the three clinical scenarios. For each scenario, we prepared 45 evaluation cases, resulting in a total of 135 cases:


\begin{itemize}
    \setlength\itemsep{0.5em}
    \item \textit{Criterion-level variant analysis.} Five experienced genetic analysts from Xinhua Hospital, each routinely interpreting more than 500 variants per year in clinical practice, contributed 39 patient records involving 45 clinically validated variants. After de-identification, the patient phenotypes, target variants, and supporting literature used in the original clinical interpretation were assembled into 45 evaluation cases.

    \item \textit{Genomic interpretation in rare disease.} We obtained 25 de-identified patients with rare diseases from the 2023-2025 curated laboratory quality control cases. An expert panel from CUHK carefully reviewed the clinical information, validated variants, interpretation reports, and corresponding VCF files. Additional clinically reportable variants, including secondary findings and candidate variants with partial phenotypic correlation, were selected from the VCF files and combined with the validated diagnostic variants, producing 45 evaluation cases with distinct candidate-variant sets. Different cases derived from the same original case were not assigned to the same participant.

    \item \textit{Molecular oncology decision-making.} We retrieved 323 human molecular-oncology case reports published after 2025 from PubMed, comprising 186 digestive-system and 137 thoracic-tumour reports. The reports were processed using a multi-model question-generation and quality-control procedure, with details provided in the Supplementary Sec.~\ref{mrmc_oncology_build}. Following expert review, 27 digestive-system and 18 thoracic-tumour cases were retained, yielding 45 evaluation cases. The selected cases were manually de-identified, and nonessential clinical details were redacted or reformulated to prevent AI systems from directly locating the corresponding source publications.
\end{itemize}
The research use of these clinical data was approved by the Ethics Committee of Xinhua Hospital.

\subsubsection{Participant Recruitment}

We enrolled 54 clinical genomic professionals as participants, comprising 45 readers and 9 adjudicators. For each of the three clinical scenarios, 15 readers with relevant domain expertise were recruited. The criterion-level variant analysis cohort consisted of genetic analysts with 3--8 years of experience, recruited through Xinhua Hospital and including analysts from other hospitals undertaking advanced clinical training there. The interpretation working group of genomic interpretation in rare disease comprised clinical geneticists, 
genomic laboratory analysts, and postdoctoral fellows with 3--10 years of experience from Hong Kong, Guangzhou, and Shandong, recruited through the Department of Obstetrics and Gynaecology, Faculty of Medicine, The Chinese University of Hong Kong. For molecular oncology decision-making, 15 clinicians and resident trainees with approximately 4--10 years of clinical experience were recruited through Shanghai Chest Hospital, including 6 specializing in thoracic oncology and 9 in digestive-system oncology. A three-member expert adjudication panel was additionally recruited from each of the three centres for its corresponding clinical scenario; each panel comprised one senior clinician with over 10 years of experience and two additional members from the same department with over 5 years of experience.  All participants gave informed consent for their study responses to be used for research analysis.

\subsubsection{Study Procedure}

Within each clinical scenario, the 45 cases were distributed among 15 readers using a Latin-square-like allocation scheme. Each reader completed nine cases: three in the unassisted condition, three in the DeepSeek-assisted condition, and three in the \ModelName{}-assisted condition. The cases assigned to an individual reader did not overlap across conditions. Consequently, each of the 45 cases was evaluated three times, once under each condition and by a different reader. The order of the three conditions was randomized across readers. For each assigned case, participants reviewed the case description and accompanying materials, completed the scenario-specific assessment, and submitted their final conclusion, written justification, confidence rating, and completion time through the online evaluation platform.

Several measures were used to reduce potential learning, order, fatigue, and information-leakage effects. No reader encountered the same case more than once, and the balanced allocation prevented individual cases from being evaluated exclusively under a particular assistance condition. Participants were required to work independently, to refrain from using any AI tools other than those explicitly permitted under their assigned condition, and to avoid any attempt to identify the source or reference answer of a task. To limit fatigue, the nine cases were completed over a two-week period, and participants were asked to distribute them as evenly as possible across this period. During the study, the expert panel corresponding to each clinical scenario answered procedural questions without providing case-specific guidance. After all assessments had been completed, participant responses were evaluated according to the predefined primary and secondary outcomes. Outcomes requiring clinical judgement were assessed by the corresponding expert panels whereas objective outcomes were calculated from the submitted responses and recorded task data using prespecified criteria~(Supplementary Sec.~\ref{primary_outcome},~\ref{secondary_outcome}).

\subsubsection{Participant Post-study Survey}
\label{questionaire}

After completing all assigned tasks, the 45 readers completed the same anonymous post-study survey, with 15 readers for each clinical scenario. Participants were instructed to complete the survey independently based on their experience with the three study conditions. Survey responses were collected anonymously and de-identified before analysis.

The survey comprised 13 experience-related items. Three open-response items asked participants to estimate their average time per case under the routine, DeepSeek-assisted, and \ModelName{}-assisted conditions. These estimates were used for the main time comparison because readers were not required to complete cases in a single uninterrupted session; platform timestamps could therefore include pauses or unrelated clinical work and were not treated as active analysis time. The remaining ten items assessed DeepSeek and \ModelName{} separately across five clinician-reported dimensions using the following five-point ordinal scales:

\begin{itemize}
    \setlength\itemsep{0.5em}
    \item \textit{Confidence after assistance:} substantially less confident than when working independently; somewhat disrupted by the system; about the same as when working independently;
    slightly more confident; or markedly more confident.
    
    \item \textit{Willingness to use the system again in clinical work:} very unwilling; somewhat unwilling; neutral; somewhat willing; or very willing.
    
    \item \textit{Satisfaction with the presentation of results in a clinical context:} very dissatisfied; somewhat dissatisfied; neutral; somewhat satisfied; or very satisfied.
    
    \item \textit{Smoothness of system operation:} not at all smooth; not very smooth; moderately smooth; relatively smooth; or very smooth.
    
    \item \textit{Reliance on system output when completing the tasks:} almost no reliance; reliance on a small part of the output; reliance on approximately half of the output; reliance on
    most of the output; or complete reliance.
\end{itemize}

\subsection{Prospective, Non-interventional Feedback Collection of \ModelName{}}
\label{deployment_method}

Following the retrospective MRMC evaluation on controlled cases, we conducted an exploratory prospective task-based assessment to evaluate \ModelName{}'s utility in daily clinical-genomics analysis, inspired by prior evaluations of clinical AI in real-world care~\cite{duggan2025ambient, zhang2026retipioneer}. Here, ``prospective'' denotes the forward collection of participants' routine clinical or research-level analysis tasks that were comparable in nature to the evaluation tasks. Experienced clinical professionals evaluated \ModelName{} on tasks arising from their usual genomic-analysis work and provided structured feedback after completing each task. The system outputs were used solely for research evaluation and were not permitted to inform or alter clinical reports, patient management or any other clinical decision. To capture clinicians' real-world needs and patterns of use, we did not prescribe a fixed set of cases or tasks, nor did we include a control condition. Accordingly, this study was intended to characterize the task-level utility of our system in a workflow-relevant, non-interventional setting.

\subsubsection{Study Design and Participant Recruitment}

We recruited 11 professionals engaged in clinical variant interpretation, each with more than five years of variant-interpretation experience and responsibility for preparing clinical reports for more than 500 cases per year. Six participants were from the Department of Genetics, Xinhua Hospital; two were from the Department of Obstetrics and Gynaecology, Faculty of Medicine, The Chinese University of Hong Kong; and one participant each was from the Department of Medical Genetics, Shandong Provincial Hospital, Anhui Provincial Hospital, and Shanghai Jiao Tong University School of Medicine. None had participated in the preceding MRMC study.

Professionals used \ModelName{} during their routine work over a prespecified one-month period. They were asked to submit feedback on clinical genomic analysis episodes for which they had actually used \ModelName{}, including, but not limited to, variant interpretation, evidence retrieval, literature analysis, and case-level analysis. They were explicitly instructed to report their experience objectively, irrespective of whether \ModelName{} had been beneficial, neutral, or detrimental to completion of the analysis.  All of them gave informed consent for their submitted feedback to be used for research analysis.

\subsubsection{Feedback Platform and Unit of Evaluation}

We developed a platform to collect structured feedback on participants' routine use of \ModelName{}. The platform did not request or extract patient names, identifiers, clinical records, genomic data or other patient-level information. Participants were instructed to follow
data-protection policies throughout their use of \ModelName{} and submission of feedback, including avoiding the disclosure of potential identifiable patient information.

One feedback submission represented one completed real-world clinical-genomics analysis episode and assessed the contribution of \ModelName{} across that episode as a whole. An episode could involve multiple rounds of interaction with \ModelName{}; the selected conversation record was used only to associate the feedback with the corresponding episode and analysis process, rather than to define an individual message or dialogue turn as an independent evaluation. A post hoc comparison with platform conversation counts indicated feedback coverage of at least 86.5\%, providing some reassurance against selective reporting (Supplementary Table~\ref{tab:deployment_feedback_coverage}).

\subsubsection{Feedback Measures}

The feedback measures were designed by a senior clinical-genomics physician at Xinhua Hospital with reference to a previous prospective clinical AI deployment study~\cite{osullivan2026cardiology}.
Each submission collected the following items:
\begin{itemize}
    \setlength\itemsep{0.5em}
    \item \textit{Analysis difficulty:} simple, not requiring AI; moderate, for which AI could provide some assistance; or difficult, for which AI could provide substantial assistance.

    \item \textit{Adoption of \ModelName{}'s conclusion:} fully adopted; partially adopted; or not adopted. For variant-assessment episodes, adoption referred to whether and to what extent the participant retained \ModelName{}'s proposed final pathogenic, likely pathogenic, variant of uncertain significance, likely benign or benign (P/LP/VUS/LB/B) classification in their final assessment. For other episodes, adoption referred to the principal task conclusion.

    \item \textit{Adoption of recommended evidence criteria:} the number of evidence criteria recommended by \ModelName{} and the number ultimately adopted by the participant. This item was optional and intended for episodes in which discrete evidence criteria were applicable.

    \item \textit{Time saved with \ModelName{}:} more than 15 min saved; 5--15 min saved; or no change in completion time or slower completion.

    \item \textit{Help with information standardization:} very helpful or prevented an error; provided some help and saved time; or provided little or no help.

    \item \textit{Help with numerical analysis:} very helpful and gave numerical retrieving or analysis that the participant might otherwise have missed; provided some help and saved time; or provided little or no help.

    \item \textit{Help with literature analysis:} very helpful, identifying relevant literature and providing accurate interpretation; provided some useful interpretation; or provided little or no help.

    \item \textit{Help with analytical or procedural guidance:} very helpful and provided a distinct, useful analytical approach; provided some help through a reasonably accurate routine analysis; or provided little or no help.

    \item \textit{Specific contribution to the episode:} an optional free-text description of the assistance provided by \ModelName{}.
\end{itemize}

All categorical items were required, whereas the evidence-criterion counts and free-text description were optional. Each submission was stored together with the participant identifier and selected conversation identifier.

\section{Data availability}

The data supporting the findings of this study are available upon reasonable request. Requests for data access should be directed to three-world@sjtu.edu.cn or to any of the corresponding-author email addresses.

\section{Code availability}

Although the source code for \ModelName{} is not publicly available at present, an online deployed Chinese--English bilingual version of the system is available at \colorbox{yellow!30}{\url{https://chatgeno.internal.onexai.com.cn/}} after registration.

\newpage
\bibliographystyle{unsrt} 
\bibliography{references} 

\clearpage
\section{Extended Data}
\definecolor{bluefill}{HTML}{E7F4FA}
\definecolor{orangefill}{HTML}{FDEEE0}
\setcounter{figure}{0} 
    \setcounter{table}{0}  
    
    \renewcommand{\figurename}{Extended Data Figure}
    \renewcommand{\tablename}{Extended Data Table}
    
    \renewcommand{\cref}[1]{\Cref{#1}} 
    \crefname{figure}{Extended Data Figure}{Extended Data Figures}
    \crefname{table}{Extended Data Table}{Extended Data Tables}
    \Crefname{figure}{Extended Data Figure}{Extended Data Figures}
    \Crefname{table}{Extended Data Table}{Extended Data Tables}

\begin{figure}[!h]
    \centering
    \includegraphics[width=1\linewidth]{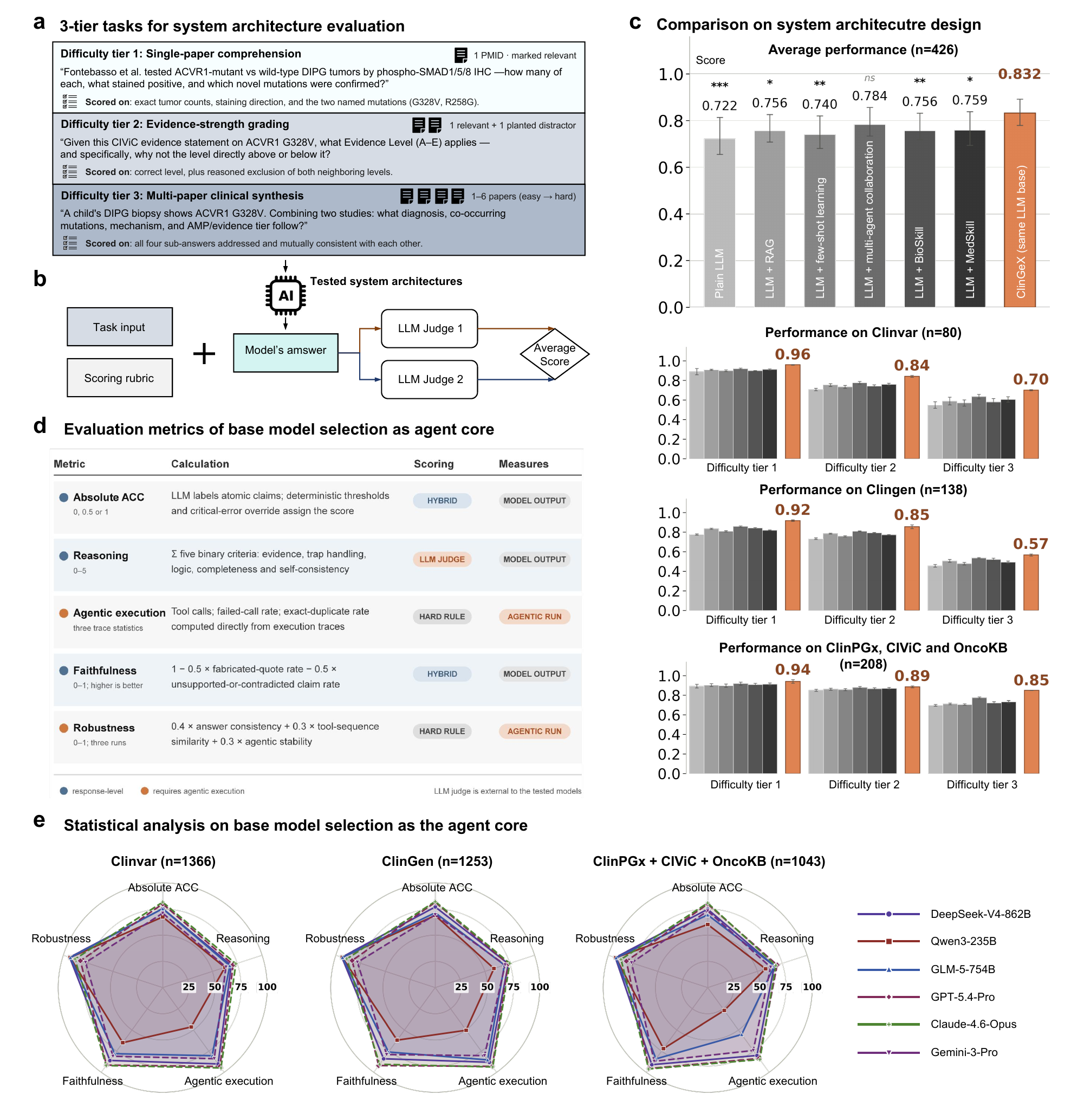}
    \caption{\textbf{Evaluation on system architecutre and agent core.}
    \textbf{a}, The evidence-based question-answering tasks, stratified into three tiers of reasoning complexity---single-paper comprehension, evidence-strength grading with one planted distractor, and multi-paper clinical synthesis---with a representative item and its scoring focus for each tier.
    \textbf{b}, Tested system architectures and scoring pipeline: every architecture answers the identical task input, and its response is graded against the item-specific rubric by two independent LLM judges, whose normalised scores are averaged.
    \textbf{c}, Ablation on system component design. Top, average normalised score over all items ($n=426$) for the plain LLM baseline, the single-component variants and the full \ModelName{} built on the same base LLM. Bottom, scores by difficulty tier for items derived from ClinVar ($n=80$), ClinGen ($n=138$) and ClinPGx, CIViC and OncoKB ($n=208$). Error bars denote 95\% confidence intervals; asterisks indicate comparisons against \ModelName{} (ns, not significant; $^{*}P<0.05$; $^{**}P<0.01$; $^{***}P<0.001$).
    \textbf{d}, The five metrics used for base-model selection, with their calculation, scoring mode and the aspect of system behaviour each measures.
    \textbf{e}, Radar plots of the five metrics on tasks derived from ClinVar ($n=1{,}366$), ClinGen ($n=1{,}253$) and ClinPGx, CIViC and OncoKB ($n=1{,}043$).}
    \label{fig:ablation}
\end{figure}

\begin{figure}[!h]
    \centering
    \includegraphics[width=1\linewidth]{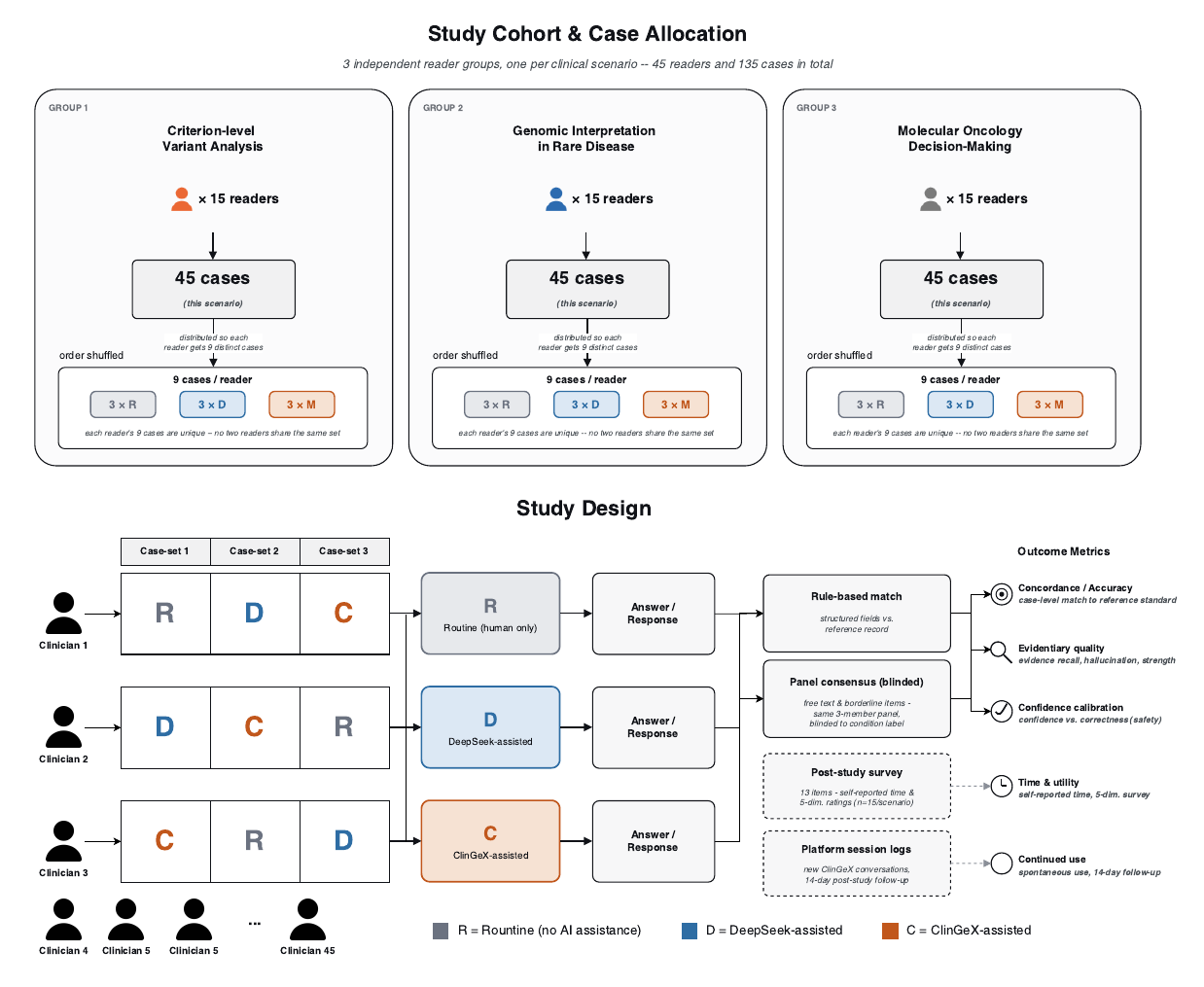}
    \caption{\textbf{Design of the multi-reader, multi-case human--AI collaboration study.}
    Three independent groups of 15 clinical genomic professionals were assigned to variant pathogenicity assessment, gene--disease interpretation or molecular oncology decision-making (45 readers in total). Each scenario comprised 45 cases organized into 15 three-case sets. Each clinician assessed nine distinct cases: three unassisted, three with DeepSeek assistance and three with \ModelName{} assistance. Condition-to-case-set assignments were counterbalanced across readers using three sequences (U--D--C, D--C--U and C--U--D). Structured responses were compared with the reference standard by rule-based matching, whereas free-text and borderline items were evaluated by a three-member expert panel blinded to study condition. Outcomes included case-level concordance or accuracy, evidentiary quality, confidence calibration, self-reported completion time and perceived utility. Participants also completed a 13-item post-study survey, and platform logs were used to assess spontaneous continued use of \ModelName{} during the 14-day follow-up. R, Routine~(no AI assistance); D, DeepSeek-assisted; C, \ModelName{}-assisted.}
    \label{fig:design_extend}
\end{figure}

\begin{figure}[!h]
    \centering
    \includegraphics[width=1\linewidth]{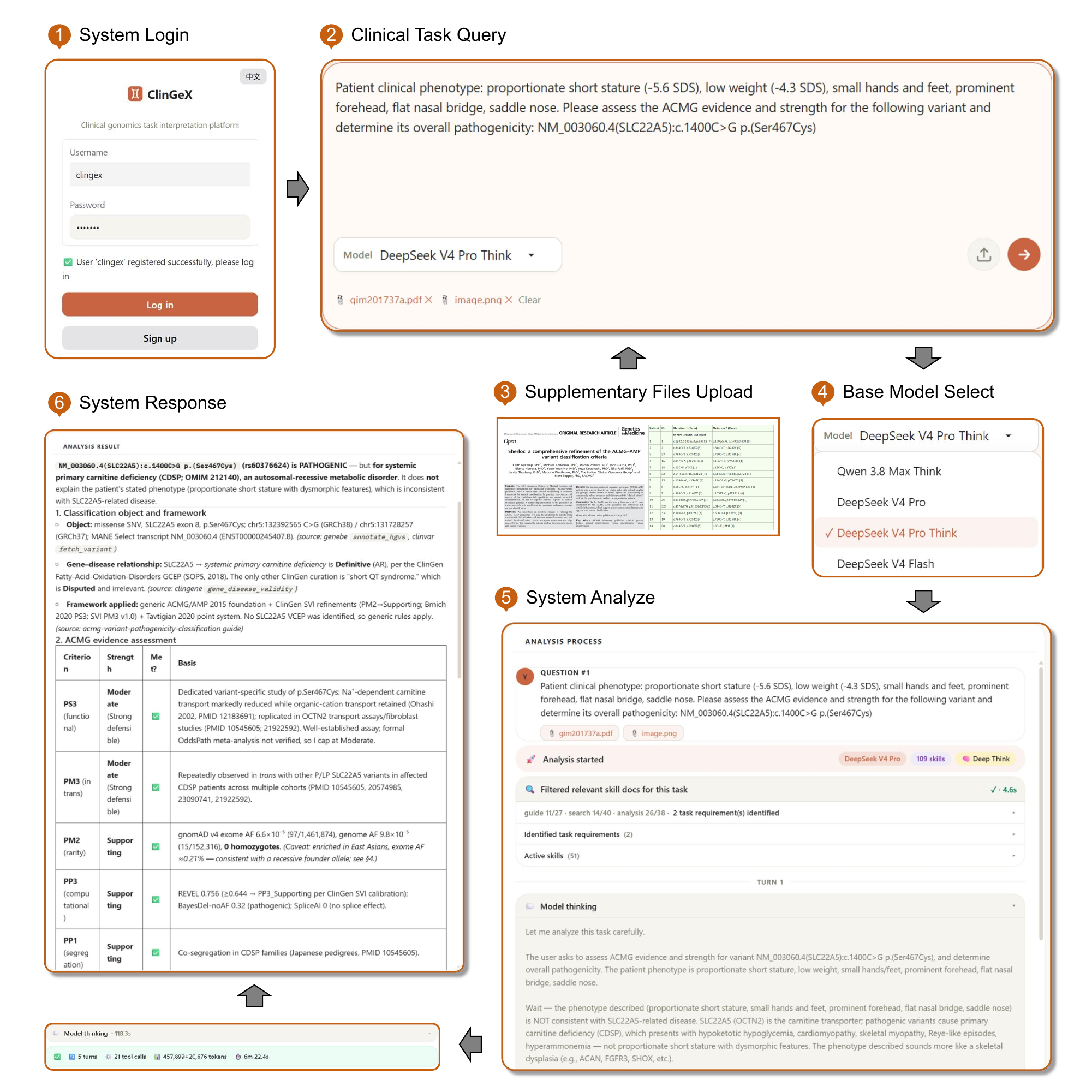}
    \caption{\textbf{Representative workflow and user interface of ClinGeX.}
    After logging into the system (1), the user enters a clinical-genomics query (2), optionally uploads supporting documents or images (3), and selects the underlying model (4). ClinGeX then identifies and invokes task-relevant skills and tools while displaying analysis progress (5), before returning a structured, evidence-supported response containing the final conclusion, supporting criteria and their interpretation (6). The example shown is an SNV pathogenicity assessment under the ACMG/AMP framework.}
    \label{fig:operation}
\end{figure}

\begin{figure}[!h]
    \centering
    \includegraphics[width=1\linewidth]{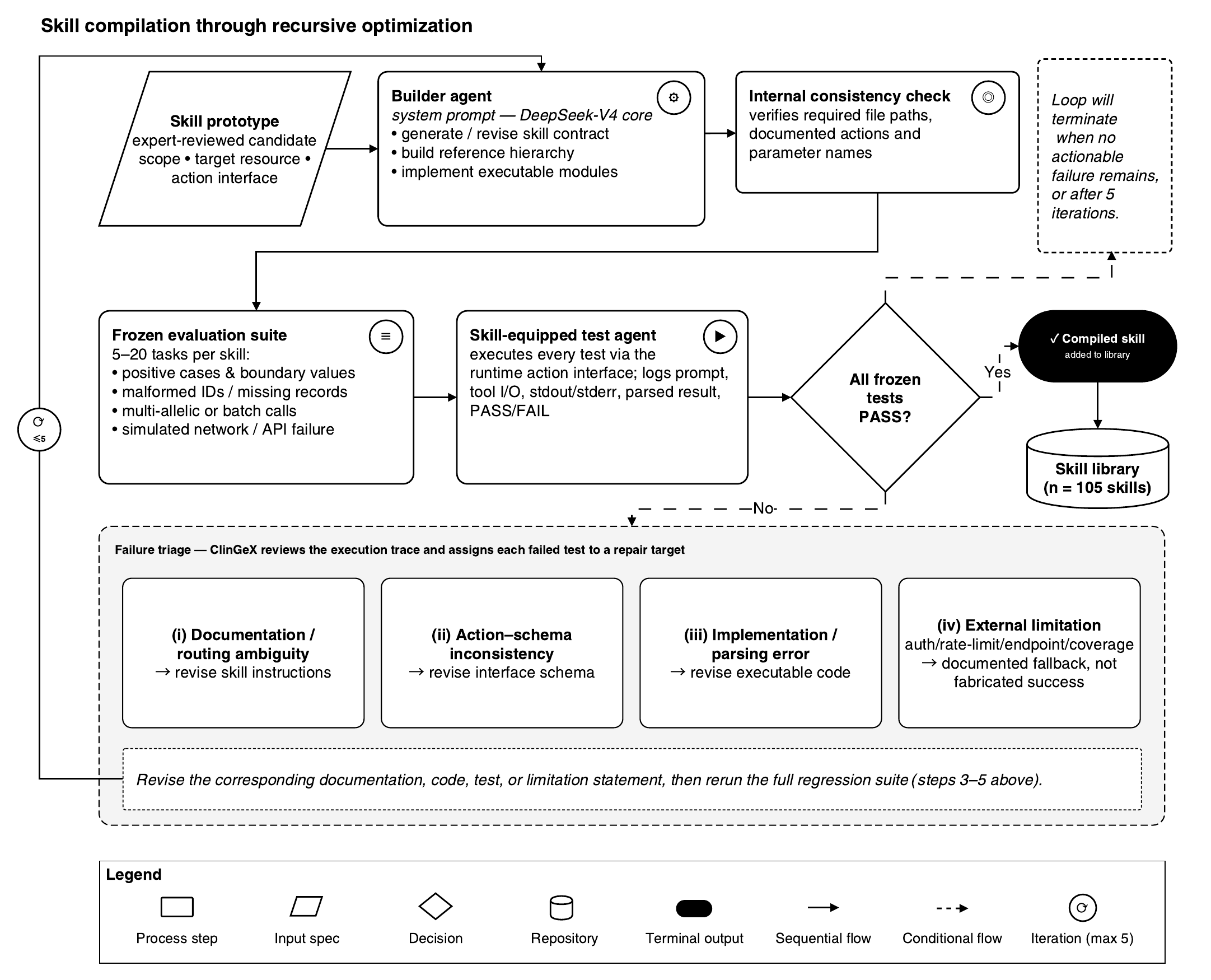}
    \caption{\textbf{Compilation of clinical-genomics skills through recursive optimization.}
    Each expert-reviewed skill prototype specifies its scope, target resource and action interface. A builder agent based on DeepSeek-V4 converts the prototype into an agent-readable skill contract, organizes its reference hierarchy and implements any required executable modules. After an internal consistency check of file paths, documented actions and parameter names, a skill-equipped test agent executes a frozen evaluation suite comprising 5--20 positive, boundary, malformed-input, batch-processing and simulated external-failure cases. The complete execution trace, including prompts, tool inputs and outputs, standard output and error streams, parsed results and test status, is retained for failure analysis. Failed tests are assigned to one of four repair targets: documentation or routing ambiguity, action--schema inconsistency, implementation or parsing error, or an external limitation requiring an explicit fallback. The corresponding documentation, interface, implementation, test or limitation statement is then revised, and the full regression suite is rerun. This cycle continues until no actionable failure remains or a maximum of five iterations is reached. Only skills passing all frozen tests are compiled and added to the deployed skill library (\(n=105\)).}
    \label{fig:compile_extend}
\end{figure}

\begin{figure}[!h]
    \centering
    \includegraphics[width=1\linewidth]{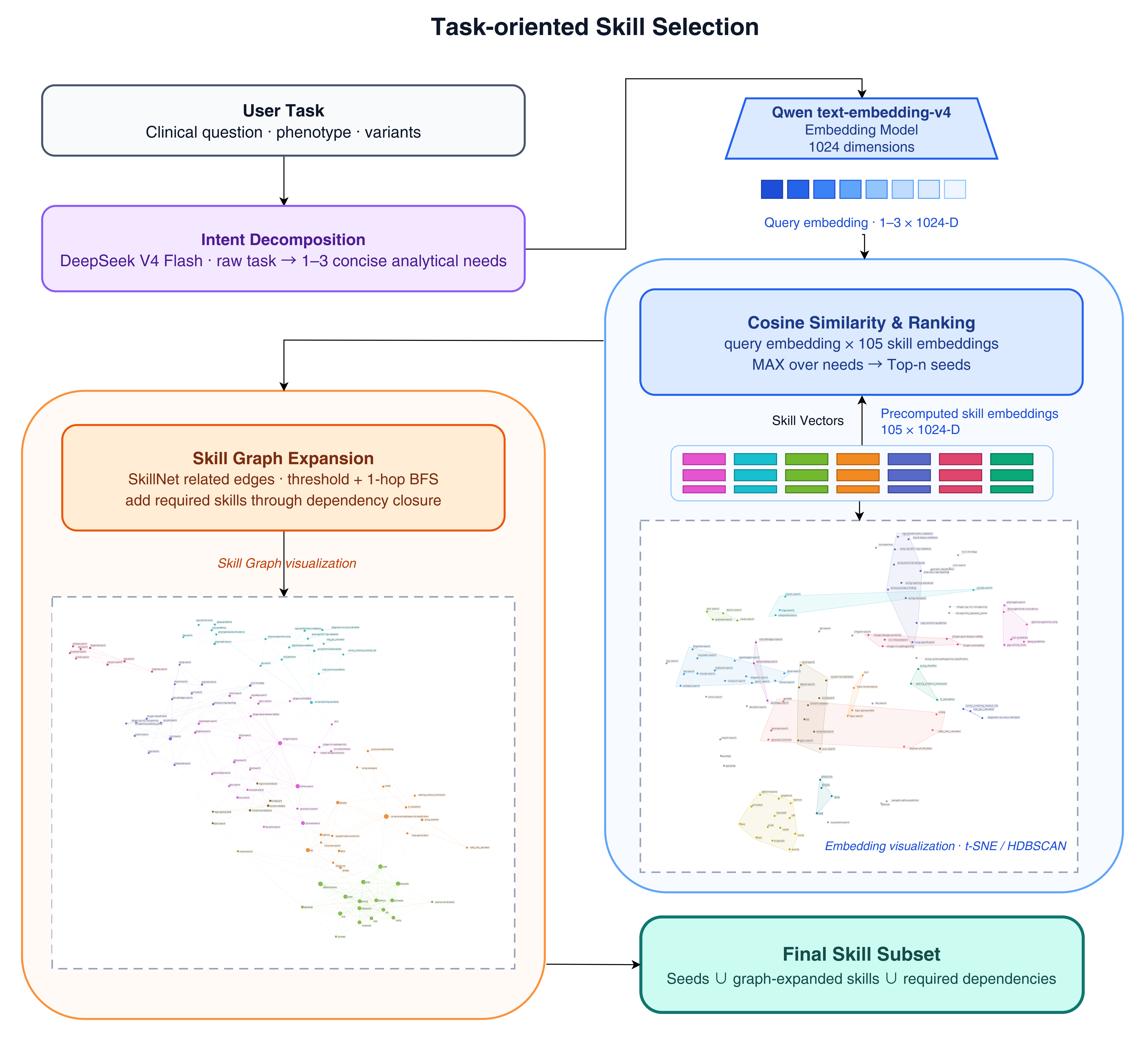}
    \caption{\textbf{Task-oriented skill selection in \ModelName{}.}
    Each user task is first decomposed by DeepSeek V4 Flash into one to three concise analytical needs. These needs are encoded as 1,024-dimensional vectors using Qwen text-embedding-v4 and compared by cosine similarity with precomputed embeddings of the 105 skills. The maximum similarity across the decomposed needs is used to rank skills and identify the top candidate seeds. The seed set is subsequently expanded over SkillNet using thresholded related-skill edges and one-hop breadth-first search, while dependency closure adds any skills required for execution. The union of the initial seeds, graph-expanded skills and required dependencies forms the final task-specific skill subset supplied to the agent. Insets illustrate the organization of the skill library in the embedding space, visualized using t-SNE and HDBSCAN, and in the SkillNet relational graph.}
    \label{fig:selection_extend}
\end{figure}

\clearpage
\appendix
\section{Supplementary}

\setcounter{figure}{0} 
    \setcounter{table}{0}  
    
    \renewcommand{\figurename}{Supplementary Figure}
    \renewcommand{\tablename}{Supplementary Table}
    
    \renewcommand{\cref}[1]{\Cref{#1}} 
    \crefname{figure}{Supplementary Figure}{Supplementary Figures}
    \crefname{table}{Supplementary Table}{Supplementary Tables}
    \Crefname{figure}{Supplementary Figure}{Supplementary Figures}
    \Crefname{table}{Supplementary Table}{Supplementary Tables}

\subsection{Main System Prompt of \ModelName{}}

The following reproduces the complete content of the deployed top-level \texttt{ClinGeX/SKILL.md}, with line wrapping, dash characters and the directory-tree glyphs normalized only for typesetting. It makes the operating stance, routing order, progressive-disclosure policy and path-resolution rules auditable.

\newtcolorbox{routerpromptdisplay}[1]{
  breakable,
  colback=blue!2,
  colframe=blue!45!black,
  boxrule=0.6pt,
  arc=1.5mm,
  left=2mm,right=2mm,top=1.5mm,bottom=1.5mm,
  title={#1},
  fonttitle=\bfseries
}

\begin{routerpromptdisplay}{Complete top-level \texttt{ClinGeX/SKILL.md} router prompt}
\small
\textbf{Front matter}\par
\texttt{name: ClinGeX}\par
\texttt{description:}\enspace Clinical interpretation co-pilot for gene-related analysis. Helps clinicians and downstream LLM agents answer questions about variants, genes, drug--gene interactions, and reportable findings by routing to authoritative interpretation guidelines, specialized analysis tools, and external knowledge databases.\par
\medskip

\textbf{ClinGeX}\par
ClinGeX is an agentic chatbot that assists with literature-related, gene-related clinical interpretation, analysis, and recommendation tasks. It is loaded by an LLM agent that has been handed a clinical-genomics request and must decide \emph{whether} and \emph{how} to invoke specialized resources before answering.\par
\medskip
This top-level \texttt{SKILL.md} is intentionally short. It is a \textbf{router}. For any concrete decision the agent identifies the relevant sub-skill category and descends into that sub-skill's own \texttt{SKILL.md}.\par
\medskip
\textbf{Operating stance.}\quad ClinGeX is restrained, professional, dialectical, and objective. It does not over-trust any single database, publication, guideline summary, prediction score, or prior model belief. It separates observed facts, authoritative rules, tool-derived results, source assertions, and inferred conclusions, and weighs convergence, conflict, absence of evidence, source quality, and task-specific context before synthesizing an answer.\par
\medskip
\textbf{Progressive disclosure.}\quad This bundle never exposes all relevant or irrelevant documents, guidelines, or tool cards at once. Nothing is preloaded. Each document reveals only its own scope and may point to deeper material that becomes visible only when it is read. Because every disclosure step can surface new documents, constraints, or routes, a previously chosen plan may have to change at any time; the agent should revise its path as soon as a newly revealed document or result requires it.\par
\medskip
\textbf{How content is read.}\quad A sub-skill's documentation is retrieved by calling \texttt{ClinGeX} with \texttt{get\_documentation=True}. Local bundle files are read with the helper scripts described in section~4, using absolute paths.\par
\medskip
\textbf{Path convention.}\quad Treat a relative path as relative to the directory of the document in which that path appears. A \texttt{get\_documentation=True} response provides \texttt{documentation\_path}, \texttt{document\_base\_dir}, and \texttt{bundle\_base\_dir}; resolve paths found in its \texttt{content} against \texttt{document\_base\_dir}. When a local file is subsequently read, resolve paths found in that file against the directory containing that file. This rule therefore continues to work at every depth of progressive disclosure, including \texttt{..} references.\par
\medskip
Only paths written in this top-level router as bundle-root paths---such as \texttt{skills/...} and \texttt{scripts/...}---are resolved against the absolute \texttt{<BASE\_DIR>} supplied outside this prompt (the same directory returned as \texttt{bundle\_base\_dir}). An already absolute path is used as given. In all cases, normalize the result to an absolute path and verify that it exists before passing it to a filesystem helper. Do not infer a path's base merely from its prefix; the location of the document containing the path determines its base.\par
\medskip

\textbf{1. The three sub-skill categories}\par
Nearly every clinical gene-related task reduces to some combination of a handful of underlying needs: consulting an authoritative guideline, looking up a fact in an external database, normalizing or transforming a piece of notation, computing or scoring something, or running a predictive model. This bundle groups the sub-skills that serve these needs into three categories:\par
\begin{itemize}[leftmargin=*,nosep]
\item \textbf{\texttt{guide/}}---authoritative procedural standards: which guideline, which section, which rule applies.
\item \textbf{\texttt{search/}}---external knowledge retrieval: current, sourced facts from public databases and literature, never guessed from memory.
\item \textbf{\texttt{analysis/}}---computation and prediction: scores, classifications, and structured evidence codes computed from a variant or gene.
\end{itemize}
All sub-skills live under \path{skills/<category>/<skill_name>/SKILL.md}.\par
\medskip

\textbf{2. Routing protocol (follow this every time)}\par
\begin{enumerate}[leftmargin=*,nosep]
\item Determine which guidelines are potentially relevant to the current task. Explicitly state and list their names, and for each candidate, why it is relevant. If none applies, say so explicitly. Then use \texttt{get\_documentation} to read the \texttt{SKILL.md} of each relevant candidate, and plan what to do next: what further documents must be read and what further sub-skills or tools must be used.
\item Execute step by step. As new results, documents, or comparison against the prior plan dictate, revise the plan and disclose further documents or sub-skills accordingly.
\item Decompose the task into the knowledge points and operational steps it involves, revising the initial plan based on the relevant guide documentation just read.
\item \textbf{Before} invoking any sub-skill for the first time, call \texttt{ClinGeX} with \texttt{get\_documentation=True} (any action value, e.g. \texttt{action="help"}) to retrieve that sub-skill's \texttt{SKILL.md}. This returns the exact list of valid actions, required parameters, supported input formats, and usage examples. \textbf{Do not} guess action names or parameter schemas---read the documentation first. If this documentation action is unavailable, read the sub-skill's \texttt{SKILL.md} directly from disk using \path{scripts/read_file.py} and its absolute path. Resolve any path revealed by a returned document relative to that document's own directory, as specified by the path convention above.
\item Invoke the sub-skill using only actions and parameters documented above.
\item After each substantive result, decide whether the plan still fits or whether a newly revealed clue, conflict, or limitation requires another reference, tool, or narrower answer.
\item Synthesize an answer that leads with the direct conclusion, followed by detailed analysis then cites which skill produced which fact, weighs convergent and conflicting evidence, and states residual uncertainty, assumptions, and clinically relevant risk.
\end{enumerate}
If nothing surfaced matches the task, \textbf{say so} rather than guessing.\par
\medskip

\textbf{3. Repository layout}\par
\path{ClinGeX/}\par
\quad\path{SKILL.md}---this file (router)\par
\quad\path{scripts/}---OS-level helpers (read/write/list/run); see section~4\par
\quad\path{assets/}---reserved, currently empty\par
\quad\path{skills/guide/<topic>/SKILL.md}\par
\quad\path{skills/search/<source>/SKILL.md}\par
\quad\path{skills/analysis/<tool>/SKILL.md}\par
\medskip

\textbf{4. OS-level helper scripts}\par
\path{scripts/} provides small, self-contained Python utilities (stdlib only, no third-party imports) that an agent can call when it needs to manipulate the local filesystem or run code as part of fulfilling a task. They are documented in \path{scripts/README.md} and are intentionally generic; they are \textbf{not} clinical tools.\par
\begin{itemize}[leftmargin=*,nosep]
\item \path{scripts/read_file.py}: A general-purpose text file reader for local files in this bundle. Robust read with encoding fallback, size guard, line slicing, optional structured parse, JSON envelope output.
\item \path{scripts/write_file.py}: Atomic write; no-clobber / overwrite / append modes; optional directory creation; dry-run; optional canonical JSON re-serialization.
\item \path{scripts/list_files.py}: Recursive listing with glob/regex filters, extension and size/mtime filters, depth limit, symlink/hidden controls, sorted output.
\item \path{scripts/run_python.py}: Execute a file or inline snippet via subprocess; argument forwarding, cwd, env passthrough, timeout, captured output, JSON envelope.
\end{itemize}
Every script accepts \texttt{--help} and emits a structured JSON envelope when \texttt{--json} is passed. Errors always exit non-zero with a JSON error object on standard error.\par
\medskip

\textbf{5. Authoring discipline}\par
Sub-skill \texttt{SKILL.md} files under \path{skills/} are the source of truth and must not be modified by users of this router.
\end{routerpromptdisplay}

\subsection{Details of Constructing the \ModelName{} Skill Library}
\label{sec:supp_clingex_construction}

\subsubsection{Skill Introduction}
\label{skill_define}

Each Agent Skill is a version-controlled directory organized as shown below. Lightweight metadata in \texttt{SKILL.md} supports discovery, while full instructions and optional resources are loaded only after selection. In \ModelName{}, a skill represents a recurrent clinical-genomics operation and specifies the workflow and evidence needed to execute and interpret it safely.

\begin{center}
\begin{minipage}{0.78\textwidth}
\begin{mdframed}[
  linewidth=0.6pt,
  linecolor=black!45,
  backgroundcolor=black!2,
  roundcorner=2pt,
  innerleftmargin=10pt,
  innerrightmargin=10pt,
  innertopmargin=7pt,
  innerbottommargin=5pt
]
\textbf{Canonical Skill Directory}\par\smallskip
\begin{Verbatim}[fontsize=\small]
<skill-name>/
|-- SKILL.md        required instructions and metadata
|-- scripts/        optional executable actions and tests
|-- references/     optional detailed source material
`-- assets/         optional templates and static data
\end{Verbatim}
\end{mdframed}
\end{minipage}
\end{center}

Within this system-specific formulation, we retained two complementary taxonomies. The five \emph{construction action types}---guideline following, entity mapping, data-source lookup, numerical calculation and model-based prediction---describe the clinical operation from which a prototype was discovered. The three \emph{deployment classes}---\texttt{guide}, \texttt{search} and \texttt{analysis}---describe how the compiled skill is exposed at run time. The axes therefore need not coincide. For example, \texttt{vep}, \texttt{mutalyzer3} and \texttt{variant-validator} were discovered as entity-mapping capabilities but are deployed under \texttt{analysis} because they perform executable transformations; conversely, ontology and identifier-mapping services are deployed under \texttt{search}. This mapping recovers the 27/13/30/12/23 construction counts reported in the main Methods while preserving the deployed 27/40/38 \texttt{guide}/\texttt{search}/\texttt{analysis} directory counts.

\subsubsection{Resource Discovery across the Five Capability Categories}
\label{sec:supp_stage1_scanning}

Discovery was frozen on 28--30 July 2026 and performed separately for the five action types (Supplementary Table~\ref{tab:stage1_flow}). A raw source record, normalized candidate and retained skill prototype were distinct units; records could be merged or one document family could support multiple operations, so successive counts are not simple retention rates.

\begin{table}[ht]
\centering
\scriptsize
\setlength{\tabcolsep}{3pt}
\renewcommand{\arraystretch}{1.16}
\caption{Source landscapes, acquisition routes and handoff from resource discovery to skill-prototype selection.}
\label{tab:stage1_flow}
\begin{tabularx}{\textwidth}{@{}p{0.105\textwidth}p{0.14\textwidth}Xp{0.14\textwidth}p{0.085\textwidth}p{0.06\textwidth}@{}}
\toprule
Action type & Candidate unit & Principal sources and acquisition route & Discovery output & Priority set & Frozen prototypes \\
\midrule
Guideline following & Normative document or linked document family & Organization-first scan: four federation directories yielded 106 unique organizations; 7 target bodies plus 9 gap-identified groups were searched through official sites and PubMed/PMC; linked procedural documents retained as families & 151 institutional mentions; 106 unique institutions; 195 documents & Authority, currency, supersession, verifiability, accessibility and faithful representability; no common cutoff & 27 \\
Entity mapping & Standard, ontology, nomenclature service or transformation tool & GA4GH GKS (11 records), EDAM-filtered bio.tools (3,256), NAR Web Server Issues 2023--2026 (276), and HPO/Monarch (10); required conversion of clinical-genomics entities or representations rather than evidence retrieval & 3,553 raw records; 3,508 normalized candidates & 31 Tier~1 candidates & 13 \\
Data-source lookup & Database, repository, knowledge base or search service & Complete harvests of the NAR Database Issue (2,594), bio.tools (33,203), re3data.org (3,516) and FAIRsharing.org (6,051); identifiers, names and URLs linked conservatively; interface, access, licence, maintenance and output metadata retained & 45,364 catalogue records; 41,000 linked resource clusters & 211 Tier~1 clusters & 30 \\
Numerical calculation & Deterministic formula, score, lookup table or decision algorithm & Reverse mapping from 195 normative documents plus targeted clinical-genomics enumeration; 30 listed methods and 23 additional normative requirements consolidated when mathematically equivalent; explicit inputs and a reproducible rule required & 53 distinct candidate methods & 19 Tier~1 methods & 12 \\
Model-based prediction & Trained model or distinct predictive signal & dbNSFP/ClinGen SVI (51 records), VEP/ANNOVAR/OpenCRAVAT (74), and CAGI/ProteinGym (80); duplicate access routes for the same signal consolidated; target, variant class, calibration, benchmark, deployment, maintenance, licence and provenance retained & 205 source records; 144 consolidated models & 18 Tier~1 models & 23 \\
\midrule
Total & & & Heterogeneous discovery units; not arithmetically pooled & & 105 \\
\bottomrule
\end{tabularx}

\vspace{2pt}
\begin{minipage}{\textwidth}
\scriptsize
\raggedright
Abbreviations: HUGO, Human Genome Organisation; ASHG, American Society of Human Genetics; ESHG, European Society of Human Genetics; IFHGS, International Federation of Human Genetics Societies; GA4GH, Global Alliance for Genomics and Health; GKS, Genomic Knowledge Standards; HPO, Human Phenotype Ontology; SVI, Sequence Variant Interpretation; CAGI, Critical Assessment of Genome Interpretation. Tier assignments prioritized expert review and were not mandatory inclusion thresholds.
\end{minipage}
\end{table}

Category boundaries followed the primary operation: representational conversion was entity mapping, evidence retrieval was data-source lookup, reproducible formulas or decision rules were numerical calculation, and learned outputs were model-based prediction. Conservative record linkage favoured residual duplication over false merging, and the predictor landscape was missense-enriched rather than complete across variant classes.

\subsubsection{Category-Specific LLM Triage and Expert Selection}
\label{sec:stage2_llm_triage}
\label{sec:stage2_expert_review}

LLMs were used to structure and prioritize the candidate space, not to make the final inclusion decision. Criterion scores were returned through structured schemas and combined deterministically using the weights in Supplementary Table~\ref{tab:stage2_triage}. Numerical screens used the same prespecified tiers: Tier~1, $\geq75$; Tier~2, 55--74; Tier~3, 35--54; and exclusion, $<35$. Scores supported ranking within an action type and were not compared across action types.

\begin{table}[!h]
\centering
\scriptsize
\setlength{\tabcolsep}{3pt}
\renewcommand{\arraystretch}{1.14}
\caption{Category-specific criteria used for candidate triage before expert review.}
\label{tab:stage2_triage}
\begin{tabularx}{\textwidth}{@{}p{0.14\textwidth}p{0.17\textwidth}Xp{0.16\textwidth}@{}}
\toprule
Action type & Screening input and implementation & Criteria and weights & Output before expert review \\
\midrule
Guideline following & Normative-document metadata and source text; direct review rather than the generic database rubric & Issuing-body authority, document type, currency, supersession, citation verifiability, accessibility, clinical scope and faithful representability of the operative rule & 195 documents reviewed; no common numerical cutoff \\
Entity mapping & Registry metadata; one structured GPT-5.5 assessment per record after a 40-record pilot & Same 12 dimensions as data-source lookup, with the target operation defined as nomenclature, coordinate, format or ontology transformation & 31 Tier~1 candidates from 3,508 normalized candidates \\
Data-source lookup & Registry metadata; one structured GPT-5-mini assessment per non-rule-filtered record & Domain centrality (15), clinical actionability (12), human orientation (8), guideline recognition (15), subdomain mapping (5), programmatic access (10), maintenance (7), access/licensing (6), query definability (4), incremental value (3), authoritative stewardship (9), and citation/adoption (6) & 211 Tier~1, 939 Tier~2, 2,600 Tier~3 and 37,250 excluded clusters \\
Numerical calculation & Method description and linked normative provisions; category-specific structured assessment & Explicit guideline requirement (20), official thresholds or weights (15), deterministic algorithm (20), availability of clinical inputs (12), independent implementation (8), breadth (10), and consequence of error (15) & 19 Tier~1, 25 Tier~2 and 9 Tier~3 methods \\
Model-based prediction & Model, calibration, benchmark and deployment metadata; category-specific structured assessment & Domain centrality (12), guideline-calibrated evidence strength (18), independent benchmark performance (12), variant-class breadth (6), precomputed scores (14), pipeline integration (10), maintenance (6), access/licence (5), stewardship (9), and citation/adoption (8) & 18 Tier~1, 35 Tier~2, 88 Tier~3 and 3 excluded models \\
\bottomrule
\end{tabularx}
\end{table}

Maintenance and citation or adoption could not always be established from static catalogue metadata and were flagged for live verification rather than inferred. For data-source-lookup candidates, a score of zero for domain centrality, guideline recognition or programmatic access forced a one-tier downgrade. Guideline documents, calculations and predictors were not forced into the database rubric because database accessibility is not an appropriate proxy for normative authority, mathematical reproducibility or clinical calibration.

Eight clinical and domain experts from seven institutions reviewed the prioritized candidates: Xinhua Hospital; the Department of Obstetrics and Gynaecology, Faculty of Medicine, The Chinese University of Hong Kong; Shanghai Ninth People's Hospital; Shanghai Sixth People's Hospital; Fudan University Shanghai Cancer Center; Shanghai Chest Hospital; and Shanghai Jiao Tong University School of Medicine. Reviewers considered clinical utility, compatibility with authoritative standards, complementarity, unmet need, legitimacy and stability of access, documentation, technical reliability and functional redundancy. They could retain a lower-tier candidate that filled an uncovered clinical operation, remove a higher-tier but redundant or non-invocable candidate, or consolidate several candidates into one prototype. This process froze 105 prototypes: 27 guideline-following, 13 entity-mapping, 30 data-source-lookup, 12 numerical-calculation and 23 model-based-prediction skills.

The scan was a frozen, structured landscape survey rather than a preregistered systematic review. Coverage differed across action types, conservative record linkage could retain residual duplicates, and most LLM triage scores were not independently repeated. These limitations constrain claims of landscape completeness but do not alter the identity or counts of the final frozen prototypes.

\subsubsection{Details of Skill Initialization and Iterative Optimization}
\label{sec:skill_compilation}

Each operation retained after expert review was initialized as a self-contained skill by a builder agent. The builder was required to examine how the underlying guideline, database, tool, calculation or predictive model could be used across clinical-genomics workflows before defining the skill contract. The common initialization prompt is shown below; source documents, technical documentation and the intended operation were supplied with the prompt for each skill.

\begin{mdframed}[
  linewidth=0.6pt,
  linecolor=black!55,
  backgroundcolor=black!2,
  roundcorner=2pt,
  innerleftmargin=8pt,
  innerrightmargin=8pt,
  innertopmargin=7pt,
  innerbottommargin=7pt
]
\small
\textbf{Prompt used to initialize a \ModelName{} skill}

You are responsible for converting the supplied clinical-genomics resource or operation into a reliable, reusable agent skill. Study the source material and technical documentation before writing any files. Do not merely summarize the resource. Determine how it may be used in clinical practice and encode the procedures needed to use it correctly.

\begin{enumerate}[leftmargin=*,nosep]
    \item \textbf{Characterize the resource.} Establish its authoritative or evidentiary status, intended users, supported clinical domains and variant or data types, access method, version or update status, and any licence or authentication constraints. Distinguish normative rules, source assertions, deterministic calculations and model-derived predictions.

    \item \textbf{Enumerate applicable use cases.} Consider all clinically plausible uses supported by the source, including direct lookup or calculation, interpretation of returned evidence, batch use, integration with upstream normalization and downstream classification, and use in incomplete or conflicting-evidence settings. Do not invent applications that are not supported by the source.

    \item \textbf{Define boundaries and dependencies.} State when the skill should and should not be used, the required inputs and accepted representations, applicability restrictions, prerequisites, expected outputs, uncertainty, known limitations and failure conditions. Identify any guideline or companion skill required to interpret the result safely.

    \item \textbf{Construct the skill package.} Create a directory containing a required \texttt{SKILL.md} and, where needed, \texttt{scripts/}, \texttt{references/} and \texttt{assets/}. The \texttt{SKILL.md} must include valid front matter, concise discovery metadata, scope, use-when and do-not-use conditions, supported actions, required and optional inputs, a stepwise workflow, output interpretation, provenance requirements, limitations, error handling and brief examples. Place long criteria, source excerpts or lookup tables in \texttt{references/}; place reusable templates or static data in \texttt{assets/}.

    \item \textbf{Implement executable actions where appropriate.} Scripts must use documented interfaces or reproducible published algorithms, validate inputs, preserve source provenance, return structured outputs, distinguish \texttt{not\_found} from \texttt{not\_applicable} and execution failure, support batch inputs where clinically useful, and never fabricate a result when an external service or required field is unavailable.

    \item \textbf{Prepare verification cases.} Define tests before revision that collectively cover every exposed action, representative valid inputs, boundary values, alternative input representations, batch and multi-record responses, non-applicable cases, absent records, malformed inputs and anticipated external-service failures. For each test, specify the input, expected output fields or behaviour and an explicit pass criterion.
\end{enumerate}

Return the complete directory structure and file contents. Ensure that every claimed capability is supported by the supplied source, exposed through the documented interface and testable through the proposed verification cases.
\end{mdframed}

Depending on the resource, initialization produced procedural instructions alone or combined them with executable scripts, indexed references and static assets. A separate test agent then received the compiled \texttt{SKILL.md} and ran 5--20 predefined cases through the same action interface used at run time. Failed cases triggered revision of the instructions, interface, implementation or limitation statement followed by rerunning the complete suite; iteration stopped when no actionable failure remained or after five cycles.

Supplementary Table~\ref{tab:revel_skill_tests} shows the complete verification suite for one representative analysis skill, REVEL. The suite was selected because it jointly tests external retrieval, variant applicability, alternative coordinate systems, multi-allelic and batch handling, missing results and clinically calibrated interpretation. 

\begin{table}[ht]
\centering
\scriptsize
\setlength{\tabcolsep}{3pt}
\renewcommand{\arraystretch}{1.14}
\caption{Complete end-to-end verification suite for the representative REVEL skill.}
\label{tab:revel_skill_tests}
\begin{tabularx}{\textwidth}{@{}p{0.035\textwidth}p{0.245\textwidth}p{0.27\textwidth}Xp{0.055\textwidth}@{}}
\toprule
ID & Action and input & Prespecified pass criterion & Observed result & Verdict \\
\midrule
S1 & Query \texttt{chr7:g.140453136A>T} & \texttt{applicable=true}; REVEL score $>0.9$ & Applicable; REVEL score 0.931 & PASS \\
S2 & Query synonymous variant \texttt{rs1799949} & \texttt{applicable=false} & Not applicable; no REVEL score returned & PASS \\
S3 & Query \texttt{chr1:g.999999999G>A} & \texttt{not\_found=true} & Variant absent from dbNSFP; \texttt{not\_found=true} & PASS \\
S4 & Query multi-allelic \texttt{rs113488022} & \texttt{disambiguated=true}; at least two alleles & Three alleles returned (A$>$G, A$>$T and A$>$C) & PASS \\
S5 & Batch query: S1 variant plus S3 variant & Two results; first applicable, second not found & Two ordered results; score 0.931 for the first and \texttt{not\_found=true} for the second & PASS \\
S6 & Interpret REVEL score 0.95 & \texttt{acmg\_evidence=PP3\_Strong} & PP3 Strong; calibrated threshold $\geq0.932$ & PASS \\
S7 & Interpret REVEL score 0.01 & \texttt{acmg\_evidence=BP4\_Strong} & BP4 Strong; calibrated threshold $\leq0.016$ & PASS \\
S8 & Interpret REVEL score 0.50 & \texttt{acmg\_evidence=Indeterminate} & Indeterminate; no PP3 or BP4 assignment & PASS \\
S9 & Query hg38 \texttt{chr7:g.140753336A>T} & \texttt{applicable=true}; REVEL score $>0.9$ & hg38 coordinate resolved; REVEL score 0.931 & PASS \\
\midrule
\multicolumn{4}{r}{Overall} & \textbf{9/9 PASS} \\
\bottomrule
\end{tabularx}
\end{table}

\subsection{Details of Building the \ModelName{} Skill Graph}
\label{sec:supp_clingex_skill_graph}

\subsubsection{Semantic Representations}

For every skill, the compact \emph{Scope} and \emph{Use when} fields were parsed from the three registry documents. Soft-wrapped continuation lines were joined, and extraction failed if either field was missing. The skill identifier and deployment category were retained as metadata but were not included in the text to be embedded. The exact embedding input was therefore
\begin{Verbatim}[fontsize=\small]
Scope: <scope text>
Use when: <use-when text>
\end{Verbatim}
Each input was encoded with Qwen \texttt{text-embedding-v4} using its default 1,024-dimensional output. Requests were made in batches of ten, and vectors were stored as a float32 matrix whose row order is fixed by \texttt{embedding\_index} in \texttt{skills.json}. The deployed selector checks that the number of metadata rows equals the first dimension of the matrix before computing similarities. Because query and skill vectors must occupy the same vector space, the online query encoder is pinned to the same model; changing the embedding model requires regeneration of the entire matrix.

The embedding layer is used for task-to-skill seed retrieval, not as the sole definition of graph edges. This distinction is important: cosine proximity captures similarity between a requested need and the declared applicability of a skill, whereas it cannot by itself determine whether two skills substitute for one another, provide independent complementary evidence, or have a hard workflow prerequisite.

\subsubsection{Exhaustive Pairwise Relationship and Dependency Scoring}

All $\binom{105}{2}=5,460$ unordered skill pairs were evaluated. The judge received the complete \texttt{SKILL.md} bodies of Skill~A and Skill~B rather than the lossy compact registry records. It assigned five independent integer scores from 0 to 4, with fixed anchors, and one categorical dependency label. The symmetric relationship score was computed deterministically rather than requested from the model:
\begin{equation}
R_{AB}=P_{AB}+I_{AB}+S_{AB}+C_{AB}+J_{AB}, \qquad R_{AB}\in[0,20],
\end{equation}
where $P$ is purpose similarity, $I$ is applicable-input overlap, $S$ is substitutability, $C$ is complementary gain from using both, and $J$ is joint use or downstream joint consumption. The dimensions were scored independently: in particular, complementarity is not defined as the inverse of substitutability. The anchors and operational meanings are given in Table~\ref{tab:skill_graph_rubric}.

\begin{table}[ht]
\centering
\footnotesize
\setlength{\tabcolsep}{3.5pt}
\renewcommand{\arraystretch}{1.12}
\caption{Anchors used for pairwise Skill Graph scoring.}
\label{tab:skill_graph_rubric}
\begin{tabularx}{\textwidth}{@{}p{0.16\textwidth}XXXXX@{}}
\toprule
Dimension & 0 & 1 & 2 & 3 & 4 \\
\midrule
Purpose similarity & Unrelated questions & Distant subject overlap & Overlapping question, different angle & Largely same question, different method/source & Essentially the same question \\
Input overlap & Inputs barely intersect & Minor intersection & Moderate overlap & Mostly overlapping inputs & Nearly identical input domain \\
Substitutability & Not interchangeable & Rarely interchangeable & Partly interchangeable & Largely interchangeable & Fully redundant \\
Complementarity & No added value together & Slight added value & Moderate independent value & Strong gain in confidence/coverage & Joint use close to essential \\
Joint usage & Essentially never co-used & Occasional co-occurrence & Common pairing & Frequently merged downstream & Near-inseparable in workflow \\
\bottomrule
\end{tabularx}
\end{table}

The dependency label was one of \texttt{none}, \texttt{A\_requires\_B}, \texttt{B\_requires\_A} or \texttt{mutual}. A direction was assigned only when using the dependent skill without first consulting or invoking the other would make its result unobtainable, incomplete or invalid---either because the other supplies a required input or because it supplies the rule under which the output must be interpreted. Merely helpful context, common co-use or higher confidence was not sufficient. Deployment class was hidden as a scoring signal so that same-directory pairs received no automatic advantage.

GPT-5.5 with high reasoning effort performed the scoring through a strict structured-output schema. The model returned the five subscores, dependency enum and a short rationale; Python validated ranges and types and summed the subscores. Calls used 32-way concurrency, exponential retry for transient/API failures (up to five attempts) and a separate structured-output validation retry (up to three attempts). Records were streamed to an append-only resumable JSONL file. The completed artefact contained all 5,460 pairs, no failed or duplicate pair, and no mutual dependency. Mean (s.d.) $R$ was 5.45 (2.99), the median was 5 and the observed range was 0--16. The 28 directed dependencies comprised 19 \texttt{A\_requires\_B} and nine \texttt{B\_requires\_A} labels.

The edge list was compiled into 105 node records. Every node stores (i) \texttt{requires}, its outgoing hard prerequisites; (ii) \texttt{required\_by}, incoming hard dependencies; and (iii) all other 104 skills in \texttt{related}, sorted by decreasing $R$ and then by identifier, with the five subscores and dependency label retained for audit. Seventeen nodes have at least one outgoing prerequisite and 19 have at least one incoming dependency. Relationship bands were defined as strong ($R=15$--20), moderate ($R=8$--14), weak ($R=1$--7) and none ($R=0$); the deployed recall threshold deliberately includes both strong and moderate relationships.

\subsubsection{Per-Message Seed Retrieval and Graph Expansion}
\label{skill_retrieval}

Selection is recomputed once for every user message, including follow-up messages, rather than once for an entire session or once per internal tool call. The selector receives the current message plus at most the previous two question--answer pairs; each prior answer is truncated to 800 characters for this purpose. A fixed DeepSeek-V4-Flash decomposition step with thinking disabled returns strict JSON containing one to three short, non-overlapping English need statements. Each need must be expressed in the same objective register as a skill's Scope/Use-when description and may represent phenotype normalization, a distinct variant type, gene--disease mechanism, or a requested classification task.

Each need is embedded independently with the pinned Qwen model. After $L_2$ normalization, cosine similarity is calculated between all 105 skill vectors and each need vector. If $E_i$ is the stored vector of skill $i$ and $Q_k$ is need $k$, the internal seed score is
\begin{equation}
s_i=\max_k\frac{E_i^{\mathsf T}Q_k}{\lVert E_i\rVert_2\lVert Q_k\rVert_2}.
\end{equation}
The maximum, rather than a mean across needs, allows a specialized skill to be retained when it matches one facet of a multi-part request. The top $n=30$ skills by $s_i$ form the seed set; no absolute cosine cutoff is applied in the deployed selector.

For each seed, all one-hop \texttt{related} edges with $R\geq x=8$ are added to a shared candidate pool. If a candidate is linked to more than one seed, only its largest edge score and the corresponding seed are retained. Candidates are ordered by decreasing edge score with identifier as the deterministic tie-breaker, and at most $m=15$ are added. The expansion is intentionally one hop: newly expanded nodes do not themselves generate further similarity candidates. Independently, \texttt{pubmed-search}, \texttt{pmc-search} and \texttt{tavily-search} are always included because literature and official-web discovery are broadly useful; these forced skills do not act as expansion seeds. Finally, the outgoing \texttt{requires} list of every recalled node is traversed repeatedly until a fixed point is reached. Hard prerequisites are therefore included regardless of cosine rank, relationship threshold or the 15-skill expansion cap.

In compact form, if $S$ is the top-30 seed set, $F$ the three forced literature skills and $T_{15}$ the operator retaining the 15 highest-scoring eligible candidates, the recalled set is
\begin{equation}
A_0=S\cup F\cup T_{15}\!\left(\bigcup_{i\in S}\{j:R_{ij}\geq8\}\right),
\qquad
A=\operatorname{closure}_{\mathrm{requires}}(A_0).
\end{equation}

\subsection{Details of the Automated Benchmark}
\label{sec:automated_benchmark}

\newtcolorbox{benchmarkdisplay}[1]{
  breakable,
  colback=blue!2,
  colframe=blue!45!black,
  boxrule=0.6pt,
  arc=1.5mm,
  left=2mm,right=2mm,top=1.5mm,bottom=1.5mm,
  title={#1},
  fonttitle=\bfseries
}

\subsubsection{Benchmark Composition and Case Construction}

The benchmark used expert-curated records with discrete reference classifications and reconstructable public evidence. Eligible records were released or substantively updated after October 2025. For each case, the source identifier and version date, clinical input, supporting literature, reference answer, and scoring annotations were frozen before system evaluation. As the training-data cutoff dates of all evaluated models preceded October 2025, this temporal restriction ensured that the case-specific curation records and final reference classifications were held out from model training.

\begin{table}[ht]
\centering
\footnotesize
\setlength{\tabcolsep}{4pt}
\renewcommand{\arraystretch}{1.15}
\caption{Composition of the system-level clinical-genomics benchmark.}
\label{tab:benchmark_composition}
\begin{tabularx}{\textwidth}{@{}
  >{\raggedright\arraybackslash}p{0.10\textwidth}
  >{\raggedright\arraybackslash}p{0.15\textwidth}
  r
  X
  X@{}}
\toprule
Task & Source & $n$ & Principal information supplied to the system & Principal reference outputs \\
\midrule
SNV interpretation & ClinVar/ClinGen expert-panel records & 79 &
Variant, disease, mode of inheritance and supporting literature &
ACMG/AMP evidence criteria, evidence strengths, final five-tier pathogenicity classification and interpretation summary \\
Gene--disease validity & ClinGen & 66 &
Gene, disease, mode of inheritance, applicable curation framework and supporting literature &
Genetic- and experimental-evidence items and scores, final validity classification and evidence summary \\
\multirow{3}{*}{Clinical actionability} & CIViC & 46 &
Molecular profile, disease, therapy, phenotype and supporting literature &
Assertion type, significance, AMP category, assertion direction and clinical assertion \\
 & OncoKB & 12 &
Gene alteration, cancer type and supporting literature &
OncoKB evidence level, associated drug or regimen, FDA level where applicable and clinical interpretation \\
 & PharmGKB/ClinPGx & 20 &
Variant or haplotype, gene, drug, phenotype, specialty population and supporting literature &
Phenotype category, evidence score, evidence level and interpretation summary \\
\midrule
Total & Five resources & 223 & & \\
\bottomrule
\end{tabularx}
\end{table}

Supporting PMIDs were deduplicated within each source record and resolved to the local literature collection through PMID and DOI mappings. Available articles were converted to normalized plain text or Markdown while retaining headings and paragraph boundaries. Inclusion was assessed at the case level: the processed collection had to contain the public literature needed to reconstruct the reference answer. The same processed texts were supplied to every system, whereas source classifications, interpretation summaries, evidence scores and other ground-truth fields were stored separately and withheld.

For SNV interpretation, a public-evidence audit reduced 2,185 ClinGen expert-panel records to 1,683; requiring complete literature coverage yielded 300 eligible cases. Stratified sampling produced 79 cases (17 Pathogenic, 17 Likely Pathogenic, 17 Uncertain Significance, 12 Likely Benign and 16 Benign). The hidden reference retained the expert-panel classification, interpretation and criterion-level evidence annotations.

\begin{benchmarkdisplay}{Representative complete case: SNV interpretation}
\small
\textbf{Source record:} ClinVar Variation ID 127692, released 5 December 2025.\\
\textbf{Case input:} \textit{NM\_000314.8(PTEN):c.79+7A$>$G}; PTEN hamartoma tumour syndrome; autosomal dominant inheritance.\\
\textbf{Evidence supplied:} processed full text of PMID 28677221. The study reported that the intronic c.79+7A$>$G substitution had no demonstrable effect on splicing.\\
\textbf{Required output:} a five-tier ACMG/AMP classification, the criteria and effective strengths applied, and a concise evidence-based interpretation.\\
\textbf{Frozen reference answer:} \emph{Likely Benign}. The PTEN Variant Curation Expert Panel specification (version 3.1.0) applied BS3 because an RNA or minigene splicing assay demonstrated no splicing impact. The sole criterion marked as met was BS3 at its default Strong strength; PM2, BS1, PP3/BP4, BP7 and the remaining assessed criteria were not met.\\
\textbf{Reference interpretation:} ``The NM\_000314.8(PTEN):c.79+7A$>$G variant meets criteria to be classified as likely benign for PTEN hamartoma tumour syndrome in an autosomal dominant manner using modified ACMG criteria. BS3: intronic variant with RNA, minigene, or other splicing assay demonstrating no splicing impact (PMID 28677221).''
\end{benchmarkdisplay}

For gene--disease validity, up to 15 eligible ClinGen records were sampled per final class (fixed seed 42), with all records retained from smaller classes. The 66 cases comprised 15 Definitive, 10 Strong, 15 Moderate, 15 Limited, 2 No Known Disease Relationship, 6 Disputed and 3 Refuted relationships. Counted genetic and experimental points were summed from the structured ClinGen evidence tables, rounded to four decimal places and retained only as hidden reference annotations.

\begin{benchmarkdisplay}{Representative complete case: gene--disease validity}
\small
\textbf{Source record:} ClinGen assertion for \textit{GRAP}--autosomal recessive nonsyndromic genetic hearing loss, evaluated under Gene Curation SOP~11.\\
\textbf{Case input:} gene, \textit{GRAP}; disease, nonsyndromic genetic hearing loss; mode of inheritance, autosomal recessive.\\
\textbf{Evidence supplied:} processed full text of Li \textit{et al.} (2019; PMID 30610177).\\
\textbf{Genetic evidence:} one female proband with congenital profound sensorineural hearing impairment carried homozygous \textit{GRAP} c.311A$>$T (p.Gln104Leu). The proband contribution was reduced from the default 1 point to 0.5 points because of consanguinity and homozygosity.\\
\textbf{Experimental evidence:} a Drosophila model showed impaired survival, locomotor deficits, disorganized scolopidia and loss of sensory nerve terminals (2 points); mouse inner-ear immunostaining localized Grap to spiral-ganglion fibres and auditory and utricular hair cells (0.5 points). The reference experimental total was therefore 2.5 points.\\
\textbf{Required output:} a permitted ClinGen validity category, genetic- and experimental-evidence totals and a concise justification.\\
\textbf{Frozen reference answer:} \emph{Limited}; genetic evidence, 0.5 points; experimental evidence, 2.5 points. A single missense variant had been reported in four individuals from two unrelated families within one publication, the disease mechanism remained uncertain and no independent publication had replicated the association. The source panel therefore concluded that the relationship had limited supporting evidence and no convincing contradictory evidence.
\end{benchmarkdisplay}

Clinical actionability retained the native source ontologies. The 46 CIViC cases comprised 20 Diagnostic, 12 Predictive, 10 Oncogenic, 3 Predisposing and 1 Prognostic assertions (33 with an applicable AMP category); the 12 OncoKB cases comprised four \texttt{LEVEL\_1}, two \texttt{LEVEL\_2}, four \texttt{LEVEL\_3A}, one \texttt{LEVEL\_4} and one \texttt{LEVEL\_R2}; and the 20 PharmGKB cases comprised one Level~1A, sixteen Level~3 and three Level~4. CIViC Diagnostic assertions were downsampled with seed 42; the other subsets were selected after source-specific eligibility filtering.

\begin{benchmarkdisplay}{Representative complete case: clinical-actionability classification}
\small
\textbf{Source record:} CIViC assertion 9, last reviewed 9 March 2026.\\
\textbf{Case input:} molecular profile, \textit{ACVR1} G328V; disease, diffuse midline glioma, H3 K27M-mutant; no therapy or phenotype field was applicable to this diagnostic assertion.\\
\textbf{Evidence supplied:} processed full texts of Fontebasso \textit{et al.} and Buczkowicz \textit{et al.} (PMIDs 24705250 and 24705254). The first study identified activating \textit{ACVR1} mutations in five of 39 paediatric midline high-grade astrocytomas and demonstrated increased phospho-SMAD1/5/8 signalling. The second identified \textit{ACVR1} alterations in 12 of 61 diffuse intrinsic pontine gliomas, including five G328V tumours, and showed activation of BMP signalling and enhanced astrocyte growth.\\
\textbf{Required output:} assertion type, significance, AMP/ASCO/CAP category where applicable, assertion direction and a justified clinical assertion.\\
\textbf{Frozen reference answer:} assertion type, \emph{Diagnostic}; significance, \emph{Positive}; AMP category, \emph{Tier II--Level C}; direction, \emph{Supports}. Reference summary: ``Supports diagnosis of diffuse intrinsic pontine glioma.'' Reference description: ``ACVR1 G328V mutations occur within the kinase domain, leading to activation of downstream signalling. Exclusively seen in high-grade paediatric gliomas, supporting diagnosis of diffuse intrinsic pontine glioma.''
\end{benchmarkdisplay}

\paragraph{Reference-answer verification.}
Reference answers were inherited from versioned source records and were not model-generated. Three physicians with more than five years of relevant experience reviewed all 223 cases for entity--record correspondence, reference-label and annotation accuracy, literature coverage, ground-truth leakage, anonymization and agreement between the task and scoring schema. Ambiguities were resolved before freezing, and system outputs did not alter the references. Only spelling, capitalization, whitespace and unambiguous aliases were normalized across databases.

\subsubsection{System Inputs, Output Schemas and Execution Protocol}
\label{underlying_model}

All systems received the same case-level clinical entities, task instructions and processed literature, with no cross-case history. Ground-truth fields were withheld; direct access to the five source databases was prohibited. Individual papers were limited to 80,000 characters and combined literature to 200,000 characters. System configurations are given in Supplementary Table~\ref{tab:baseline_settings}.

Because some systems provided their own search or database tools, we additionally used an LLM to audit each complete execution record for substantive case-specific information obtained from the corresponding reference resource, either directly or indirectly through an intermediary. Mere mention of a resource or use of general framework terminology was not counted. Positive records were uncommon for every system and task (Supplementary Table~\ref{tab:reference_source_audit}).

\begin{table}[ht]
\centering
\footnotesize
\setlength{\tabcolsep}{5pt}
\renewcommand{\arraystretch}{1.15}
\caption{Audit of substantive information acquisition from benchmark reference resources.}
\label{tab:reference_source_audit}
\begin{threeparttable}
\begin{tabularx}{\textwidth}{@{}
  >{\raggedright\arraybackslash}p{0.18\textwidth}
  >{\centering\arraybackslash}X
  >{\centering\arraybackslash}X
  >{\centering\arraybackslash}X@{}}
\toprule
System & SNV interpretation: ClinVar/ClinGen & Gene--disease validity: ClinGen & Clinical actionability: CIViC/OncoKB/PharmGKB \\
\midrule
OpenBioLLM & 0/79 (0\%) & 0/66 (0\%) & 0/78 (0\%) \\
OpenAI DeepResearch & 3/79 (3.8\%) & 2/66 (3.0\%) & 2/78 (2.6\%) \\
SciSpace & 1/79 (1.3\%) & 1/66 (1.5\%) & 3/78 (3.8\%) \\
Biomni & 4/79 (5.1\%) & 1/66 (1.5\%) & 1/78 (1.3\%) \\
TxAgent & 0/79 (0\%) & 2/66 (3.0\%) & 1/78 (1.3\%) \\
\ModelName{} & 0/79 (0\%) & 0/66 (0\%) & 0/78 (0\%) \\
\bottomrule
\end{tabularx}
\begin{tablenotes}[flushleft]\footnotesize
\item Values are the number of execution records with direct or indirect acquisition of substantive case-specific information from the corresponding reference resource divided by the number of cases in that task.
\end{tablenotes}
\end{threeparttable}
\end{table}

\begin{table}[ht]
\centering
\footnotesize
\renewcommand{\arraystretch}{1.15}
\caption{Systems and underlying language models used in the system-level comparison.}
\label{tab:baseline_settings}
\begin{threeparttable}
\begin{tabularx}{\textwidth}{@{}
  >{\raggedright\arraybackslash}p{0.11\textwidth}
  >{\raggedright\arraybackslash}p{0.15\textwidth}
  >{\raggedright\arraybackslash}p{0.17\textwidth}
  >{\raggedright\arraybackslash}X@{}}
\toprule
System & Underlying LLM used & System type & Benchmark configuration \\
\midrule
OpenBioLLM &
OpenBioLLM-70B (derived from Llama~3 70B) &
Domain-adapted biomedical LLM &
Model response generated without an external agent framework. \\
OpenAI DeepResearch &
\texttt{o3-deep-research} &
Commercial deep-research agent &
OpenAI Responses interface with native web search and code interpreter; access to the five ground-truth databases was prohibited. \\
SciSpace &
Proprietary LLM backend (not disclosed by provider) &
Commercial literature-research system &
Provider-hosted interface; the backend model could not be independently configured or verified. \\
Biomni &
DeepSeek-V4 (\texttt{deepseek-v4-pro}) &
Open-source biomedical code-execution agent &
Biomni A1 with its tool retriever enabled; a new A1 instance was created for each case. \\
TxAgent &
DeepSeek-V4 (\texttt{deepseek-v4-pro}) &
Open-source ToolUniverse agent &
Original TxAgent orchestration layer retained; inference backend replaced with the same DeepSeek model used by \ModelName{}. \\
\ModelName{} &
DeepSeek-V4 (\texttt{deepseek-v4-pro}) &
Skill-centred clinical-genomics agent &
Full skill library and orchestration workflow. \\
\bottomrule
\end{tabularx}
\begin{tablenotes}[flushleft]\footnotesize
\item Archived execution records verified the backends for OpenAI DeepResearch, Biomni, TxAgent and \ModelName{}. SciSpace exposed no model identifier and is therefore reported as undisclosed; the archived materials did not specify the OpenBioLLM repository revision or numerical precision.
\end{tablenotes}
\end{threeparttable}
\end{table}

The common case-facing instruction layer is reproduced below. Framework-specific tool-calling scaffolds differed, but clinical fields, literature and required outputs were controlled across systems.

\begin{benchmarkdisplay}{Common benchmark instruction and user-prompt template}
\small\ttfamily
You are a clinical genomics expert. Use the provided literature and your expert knowledge to answer accurately. Never fabricate data. When you cite literature, be specific about which paper and which finding.

========== TASK-SPECIFIC INSTRUCTIONS ==========\\
\{source-specific task description from the frozen task record\}

\#\# Case Inputs\\
- \{task-specific clinical entities\}\\

\#\# Literature\\
The following papers are provided for this case. Read them carefully and use the evidence within, together with your expert knowledge, to reach a well-reasoned conclusion.\\
---\\
\#\#\# Paper: \{source path\}\\
\{processed full text; per-paper and total-context limits applied\}\\

\#\# Required Output\\
\{source-specific final classification, evidence fields and concise justification\}
\end{benchmarkdisplay}

The system-specific execution controls recorded in the archived harnesses are summarized in Supplementary Table~\ref{tab:execution_controls}. Free-text answers were subsequently converted into task-specific structured fields by a fixed extraction pipeline. Missing or unrecognized fields were retained as null values rather than repaired from the reference answer.

\begin{table}[ht]
\centering
\footnotesize
\setlength{\tabcolsep}{4pt}
\renewcommand{\arraystretch}{1.15}
\caption{Execution controls recorded in the system-level benchmark harnesses.}
\label{tab:execution_controls}
\begin{tabularx}{\textwidth}{@{}
  >{\raggedright\arraybackslash}p{0.14\textwidth}
  X
  X@{}}
\toprule
System & Agent or tool budget & Request and recovery controls \\
\midrule
OpenBioLLM &
Direct model generation; no external agent loop. &
The exact checkpoint revision, numerical precision and generation configuration were not recorded in the archived manifest. \\
OpenAI DeepResearch &
At most 30 combined native web-search and code-interpreter calls per job. &
Jobs were polled every 5~s for at most 2,700~s; the client timeout was 3,600~s. Transient requests used up to five attempts, and failed background jobs could be recreated up to six times. \\
SciSpace &
Provider-managed literature-search workflow. &
Tool budget, timeout and recovery policy were not exposed by the provider. \\
Biomni &
Tool retriever enabled; a fresh A1 instance for every case; individual executed code blocks limited to 600~s. &
LLM request timeout, 180~s. \\
TxAgent &
At most 50 multistep rounds; maximum 2,048 newly generated tokens per model call; nominal context ceiling, 90,240 tokens; temperature, 0.3. &
LLM request timeout, 180~s; up to five attempts with exponential backoff. \\
\ModelName{} &
At most 50 agent turns. &
LLM request timeout, 180~s; up to five attempts with exponential backoff. \\
\bottomrule
\end{tabularx}
\end{table}

Final-classification and evidence metrics were computed from the frozen single-output records described below. The repeated-run consistency and per-case wall-clock time reported in the main figure were obtained through a separate repetition-and-timing workflow and were not derived from those single-output metric files.

\subsubsection{Metric Implementation and Denominators}

All reported classification and evidence metrics were recomputed from the frozen structured reference data and extracted system outputs by deterministic code. The denominators differ by task and, in two cases, by endpoint; they are summarized in Supplementary Table~\ref{tab:metric_denominators}. No LLM judge determined whether a final classification matched the reference.

\begin{table}[ht]
\centering
\footnotesize
\setlength{\tabcolsep}{4pt}
\renewcommand{\arraystretch}{1.15}
\caption{Implemented metric denominators in GeneBench.}
\label{tab:metric_denominators}
\begin{tabularx}{\textwidth}{@{}
  >{\raggedright\arraybackslash}p{0.13\textwidth}
  >{\raggedright\arraybackslash}p{0.20\textwidth}
  X@{}}
\toprule
Task & Endpoint & Implemented denominator \\
\midrule
\multirow{3}{=}{SNV interpretation} &
Final concordance and significant-error rate &
All 79 cases; missing or unparsable predictions remain incorrect observations. \\
 & Sensitivity / specificity &
The 34 pathogenic or likely pathogenic references / the 28 benign or likely benign references, respectively; VUS references are not in either binary denominator. \\
 & Evidence recall / strength concordance &
All reference criteria pooled, or the 79 case-level recall values; strength uses only correctly recalled criterion codes. \\
\midrule
\multirow{2}{=}{Gene--disease validity} &
Concordance and severity errors &
Cases with a parsable prediction in the seven-category vocabulary ($N_{\mathrm{valid}}\leq66$), reported separately for each system. \\
 & Score coverage / score MAE &
Coverage denominator, $N_{\mathrm{valid}}$; MAE denominator, cases in which the relevant numerical score or interval was stated. \\
\midrule
\multirow{2}{=}{Clinical actionability} &
Final concordance and severity errors &
The 65 tier-applicable cases: 33 CIViC, 12 OncoKB and 20 PharmGKB. \\
 & Complete-answer accuracy &
All 78 cases. \\
\bottomrule
\end{tabularx}
\end{table}

\subsubsection{SNV-Specific Evaluation Metrics}

Final-classification concordance required an exact five-tier match. Pathogenic/Likely Pathogenic and Benign/Likely Benign formed the positive and negative groups for sensitivity and specificity; VUS cases were excluded. A significant error was a direction reversal between these groups and used all 79 cases as its denominator.

For reference criterion set $E_i$ and recovered set $\hat E_i$, micro recall was $\sum_i|E_i\cap\hat E_i|/\sum_i|E_i|$ and macro recall was the mean case-level recall; the micro value is the weighted evidence recall reported in the main figure. A criterion counted only when asserted as met for the evaluated variant--disease pair. Strength concordance required exact strength agreement among correctly recovered criteria.

\subsubsection{Gene--Disease Validity-Specific Evaluation Metrics}

The fixed ranks were Refuted $=-2$, Disputed $=-1$, No Known Disease Relationship $=0$, Limited $=1$, Moderate $=2$, Strong $=3$ and Definitive $=4$. Concordance and rank differences used parsable seven-category outputs; differences of at least two levels defined primary severe over- or under-classification (three levels in supplementary analyses).

Evidence ``coverage'' denotes reporting of a usable genetic or experimental score, not item-level recall. Reference scores were capped at 12 and 6 points, respectively. For an extracted interval $[L_i,H_i]$ and reference $g_i$, error was $\max(0,L_i-g_i,g_i-H_i)$; MAE was conditional on a stated score, with no imputation.

\subsubsection{Clinical Actionability-Specific Evaluation Metrics}

Final-classification concordance used the 65 cases with an applicable tier: CIViC AMP category, OncoKB level or PharmGKB level. Complete-answer accuracy used all 78 cases and required, respectively, CIViC assertion type/significance/direction and applicable AMP category; OncoKB level and exact drug set; or PharmGKB level and a score within 0.5 points of reference.

Severity analyses used the source-specific ranks in Supplementary Table~\ref{tab:actionability_rank}. A rank difference of at least two levels defined primary severe over- or under-classification (three levels in supplementary analyses); missing or unmappable labels remained in the fixed 65-case denominator.

\begin{table}[ht]
\centering
\footnotesize
\setlength{\tabcolsep}{4pt}
\renewcommand{\arraystretch}{1.15}
\caption{Unified evidence-strength ranks implemented for clinical-actionability severity analyses.}
\label{tab:actionability_rank}
\begin{tabularx}{\textwidth}{@{}c X X X@{}}
\toprule
Rank & CIViC AMP category & OncoKB level & PharmGKB level \\
\midrule
6 & Tier I--Level A & \texttt{LEVEL\_1} & 1A \\
5 & Tier I--Level B & \texttt{LEVEL\_2}; \texttt{LEVEL\_R1} & 1B \\
4 & Tier II--Level C & \texttt{LEVEL\_3A} & 2A \\
3 & Tier II--Level D & \texttt{LEVEL\_3B}; \texttt{LEVEL\_R2} & 2B \\
2 & Tier III & \texttt{LEVEL\_4} & 3 \\
1 & Tier IV & No response-category equivalent & 4 \\
\bottomrule
\end{tabularx}
\end{table}

The placement of OncoKB resistance categories is an analysis convention rather than an official crosswalk: \texttt{LEVEL\_R1} was aligned with standard-care evidence at rank~5 and \texttt{LEVEL\_R2} with emerging clinical evidence at rank~3. This convention was fixed in the evaluator before system comparison.

\paragraph{Rule-based scoring and missing outputs.}
All matches were determined from parsed outputs and frozen references using rules fixed before comparison; the extraction model only normalized labels, criteria and numerical intervals. Missing SNV and actionability predictions remained incorrect in fixed denominators, whereas unparsable gene--disease predictions were reported separately and excluded from concordance and severity denominators. Supplementary Table~\ref{tab:metric_denominators} gives all endpoint-specific denominators.
\subsubsection{Blinded Clinician Assessment of Report Quality}
\label{sec:clinician_assessment}

Nine clinicians rated the reports from \ModelName{}, Biomni and DeepResearch for all 223 cases. Six clinicians with more than three years of genetic-analysis experience reviewed SNV or gene--disease reports; three internal-medicine physicians with more than five years of experience reviewed clinical-actionability reports.

Within each task, cases were partitioned among three reviewers (SNV: 27/26/26; gene--disease: 22/22/22; actionability: 27/26/25), with no duplicate rating. The assigned reviewer assessed all three reports for a case. Report identities were independently permuted to Result~A--C for each case and revealed only after export.

Each report received eight ordered three-category ratings without a not-applicable option: six common dimensions (Supplementary Table~\ref{tab:clinician_common_rubric}) and two task-specific dimensions (Supplementary Table~\ref{tab:clinician_task_rubric}).

\begin{table}[ht]
\centering
\footnotesize
\setlength{\tabcolsep}{4pt}
\renewcommand{\arraystretch}{1.15}
\caption{Clinician-assessment dimensions shared across all three clinical-genomics tasks.}
\label{tab:clinician_common_rubric}
\begin{tabularx}{\textwidth}{@{}
  >{\raggedright\arraybackslash}p{0.14\textwidth}
  X
  >{\raggedright\arraybackslash}p{0.18\textwidth}@{}}
\toprule
Dimension & Operational assessment & Ordered response categories \\
\midrule
Overall acceptability &
Whether the report's conclusion and supporting interpretation could be accepted for the stated clinical-genomics task after expert review. &
Yes; Partly; No \\
Required revision &
The amount of substantive correction or rewriting required before the report could be accepted. This was rated separately from acceptability. &
No revision; Partial revision; Substantial revision \\
Citation accuracy &
Whether cited publications were identifiable and whether the findings attributed to them were supported by the cited source. &
Largely accurate; Partly accurate; Inaccurate \\
Guideline adherence &
Whether the reasoning, terminology and final conclusion followed the framework applicable to the case: ACMG/AMP, ClinGen Gene Curation, CIViC, OncoKB or PharmGKB. &
Largely adherent; Partly adherent; Not adherent \\
Hallucinated information &
Whether the report introduced fabricated or unsupported facts, numerical values, citations, study findings or clinical assertions. &
Absent; Partly present; Present \\
Potential clinical risk &
Whether errors or unsupported claims in the report could plausibly mislead variant classification, gene--disease assessment, therapeutic interpretation or another downstream clinical decision. &
Absent; Partly present; Present \\
\bottomrule
\end{tabularx}
\end{table}

\begin{table}[ht]
\centering
\footnotesize
\setlength{\tabcolsep}{4pt}
\renewcommand{\arraystretch}{1.15}
\caption{Task-specific dimensions in the clinician assessment.}
\label{tab:clinician_task_rubric}
\begin{tabularx}{\textwidth}{@{}
  >{\raggedright\arraybackslash}p{0.11\textwidth}
  >{\raggedright\arraybackslash}p{0.15\textwidth}
  X
  X@{}}
\toprule
Task & Dimension & Operational assessment & Ordered response categories \\
\midrule
\multirow{2}{=}{SNV interpretation} &
Evidence use &
Whether the report selected and applied relevant population, computational, functional, segregation, allelic and other evidence appropriately to the evaluated variant--disease pair. &
Largely reasonable; Partly reasonable; Unreasonable \\
 & Evidence strength &
Whether the strengths assigned to ACMG/AMP criteria, including any strength modification under the applicable expert-panel specification, were appropriate. &
Largely reasonable; Partly reasonable; Unreasonable \\
\midrule
\multirow{2}{=}{Gene--disease validity} &
Genetic-evidence analysis &
Whether proband, segregation, case--control and other genetic evidence was identified, interpreted and weighted appropriately under the ClinGen framework. &
Largely reasonable; Partly reasonable; Unreasonable \\
 & Experimental-evidence analysis &
Whether functional alteration, biochemical function, expression, interaction, model-system and rescue evidence was identified, interpreted and weighted appropriately under the ClinGen framework. &
Largely reasonable; Partly reasonable; Unreasonable \\
\midrule
\multirow{2}{=}{Clinical actionability} &
Evidence relevance and completeness &
Whether the selected evidence was relevant to the specific variant, disease, drug or therapeutic context and whether material evidence needed for the clinical assertion was covered. &
Largely relevant and complete; Partly relevant and complete; Irrelevant or incomplete \\
 & Evidence interpretation &
Whether the selected evidence was interpreted correctly with respect to its direction, clinical significance, evidence tier, therapeutic association and source-specific conclusion. &
Largely correct; Partly correct; Incorrect \\
\bottomrule
\end{tabularx}
\end{table}

\paragraph{Completion and aggregation.}
All cases were complete: 669 reports and 5,352 ordinal ratings, with no missing value or imputation. After de-blinding, category counts and proportions were summarized by task, system and dimension. No composite score was defined; inter-rater agreement was not estimated because cases were partitioned rather than multiply rated. These ratings neither altered the references nor entered the automated metrics.

\subsection{Selection of the System Architecture}
\label{sec:framework_selection}

\subsubsection{Study Objective and Controlled Comparison}

This experiment compared the complete \ModelName{} architecture with an unenhanced DeepSeek-V4 baseline and five single-strategy enhancements on evidence-grounded question--answering items. DeepSeek-V4, item text and hidden-reference controls were held fixed, so conditions differed only in the enhancement or orchestration layer listed in Supplementary Table~\ref{tab:framework_conditions}. Unlike the 223-case end-to-end benchmark, this dataset isolated specific evidence-reasoning operations.

\begin{table}[ht]
\centering
\footnotesize
\setlength{\tabcolsep}{4pt}
\renewcommand{\arraystretch}{1.15}
\caption{System-framework conditions in the controlled enhancement-strategy ablation.}
\label{tab:framework_conditions}
\begin{tabularx}{\textwidth}{@{}
  >{\raggedright\arraybackslash}p{0.17\textwidth}
  >{\raggedright\arraybackslash}p{0.18\textwidth}
  X@{}}
\toprule
Condition & Component added to DeepSeek-V4 & Operational configuration \\
\midrule
DeepSeek & None &
The model answered the supplied question directly. It was not given retrieval tools, worked examples, domain-skill instructions or the ability to delegate to sub-agents. This condition is referred to as the unenhanced base-model baseline. \\
DeepSeek + RAG & Search-function bundle &
All search-related capabilities implemented in the \ModelName{} skill collection were consolidated into a retrieval bundle and exposed to the model. The condition did not receive non-search clinical-analysis skills, the skill graph or multi-agent orchestration. \\
DeepSeek + fewshot & Three worked question--answer examples &
Three worked examples were inserted into the model context before the test item. The examples illustrated the expected question--answer interaction, but this condition had no external retrieval, domain-skill library or sub-agent delegation. \\
DeepSeek + MA & Planner--executor--reviewer workflow &
Three DeepSeek-V4 instances were assigned the roles of planner, executor and reviewer. The planner decomposed the item, the executor produced a draft by following the plan, and the reviewer checked the draft and returned revision instructions before finalisation. \\
DeepSeek + BioSkill & GitHub BioSkill repository &
The open-source BioSkill instructions and associated resources were exposed through a common wrapper while retaining DeepSeek-V4 as the core model. The \ModelName{}-specific selector, skill graph and skill library were not used. \\
DeepSeek + MedSkill & GitHub MedSkill repository &
The open-source MedSkill instructions and associated resources were exposed through the same wrapper principle and with the same DeepSeek-V4 core. The \ModelName{}-specific selector, skill graph and skill library were not used. \\
\ModelName{} (DeepSeek-base) & Complete skill-centred architecture &
The final system used its 105 clinical-genomics skills, task decomposition, skill selection, graph-based recall of related and prerequisite skills, progressive disclosure of instructions, plan--execute orchestration, evidence-retrieval and code-execution tools, and synthesis of the resulting execution trace into a final response. This was the prespecified full-system condition and was not a union of the BioSkill and MedSkill repositories. \\
\bottomrule
\end{tabularx}
\end{table}

\subsubsection{Implementation of the Seven Framework Conditions}

Supplementary Table~\ref{tab:framework_conditions} defines the seven conditions. The RAG condition exposed only the search-facing functions from the \ModelName{} library; few-shot used the same three fixed demonstrations for every item; and the multi-agent condition used sequential planner, executor and reviewer roles. BioSkill (\url{https://github.com/GPTomics/bioSkills}) and MedSkill (\url{https://github.com/aipoch/medical-research-skills}) were frozen snapshots exposed through the same discovery-and-load wrapper without adding \ModelName{} content or graph edges. The full \ModelName{} condition retained its native skill selection, dependency expansion and progressive execution workflow. No condition received the hidden reference answer or scoring rubric.

\subsubsection{Benchmark Construction}

The benchmark drew from ClinVar, ClinGen, CIViC, OncoKB and ClinPGx. Eighty diversity-oriented anchors were selected deterministically: 10 each from OncoKB and CIViC and 20 each from ClinPGx, ClinVar and ClinGen. Selection favoured distinct genes and, where applicable, literature-supported pathogenic assertions or ClinGen records with both genetic and experimental evidence and non-trivial scoring or mechanistic issues.

DeepSeek-V4 generated items in 240 independent anchor--tier jobs using source- and tier-specific instructions, structured records and linked full text; related records could be retrieved for clinically plausible comparisons or distractors. Retained items had to pose a meaningful clinical operation, have a literature-verifiable answer and require interpretation rather than copying. Schema validation required a source-grounded answer, numeric PMIDs with relevance labels, at least one relevant PMID, a construction rationale and an item-level rubric. The final 426 items are summarized in Supplementary Table~\ref{tab:framework_benchmark_composition}.

\begin{table}[ht]
\centering
\footnotesize
\setlength{\tabcolsep}{5pt}
\renewcommand{\arraystretch}{1.15}
\caption{Composition of the framework-selection benchmark by source and reasoning tier.}
\label{tab:framework_benchmark_composition}
\begin{tabularx}{\textwidth}{@{}
  >{\raggedright\arraybackslash}p{0.17\textwidth}
  >{\raggedright\arraybackslash}p{0.24\textwidth}
  rrrr@{}}
\toprule
Source & Clinical-genomics domain & Tier~1 & Tier~2 & Tier~3 & Total \\
\midrule
ClinVar & Variant interpretation & 39 & 22 & 19 & 80 \\
ClinGen & Gene--disease validity & 54 & 53 & 31 & 138 \\
CIViC & Clinical actionability & 30 & 16 & 19 & 65 \\
OncoKB & Clinical actionability & 29 & 14 & 14 & 57 \\
ClinPGx & Clinical actionability & 45 & 19 & 22 & 86 \\
\midrule
Total & & 197 & 124 & 105 & 426 \\
\bottomrule
\end{tabularx}
\end{table}

\subsubsection{Reasoning Tiers and Task-Specific Construction Rules}

Tiers were defined by the required evidence operation, as specified in Supplementary Table~\ref{tab:framework_tiers}. Tier~1 items used no deliberately irrelevant PMID and cited a mean of 1.19 publications. Tier~2 allowed realistic non-supporting evidence when available (43/124 items; mean 2.82 publications). Tier~3 integrated multiple sources (mean 4.93 publications) and comprised 51 easy, 27 medium and 27 hard items according to evidence volume, conflict and irrelevant context.

\begin{table}[ht]
\centering
\footnotesize
\setlength{\tabcolsep}{4pt}
\renewcommand{\arraystretch}{1.15}
\caption{Operational definition of the three reasoning tiers in the framework-selection benchmark.}
\label{tab:framework_tiers}
\begin{tabularx}{\textwidth}{@{}
  >{\raggedright\arraybackslash}p{0.10\textwidth}
  >{\raggedright\arraybackslash}p{0.18\textwidth}
  X
  >{\raggedright\arraybackslash}p{0.22\textwidth}@{}}
\toprule
Tier & General operation & Representative task-specific operations & Principal scoring focus \\
\midrule
1 & Single-publication comprehension &
Recover a paper-specific clinical, quantitative, functional, segregation or pharmacogenomic finding and interpret it in the anchor context. &
Correct factual elements; correct direction and magnitude where applicable; source-specific caveats; absence of invented details. \\
2 & Evidence selection and strength grading &
Map evidence to an ACMG/AMP criterion; justify a ClinGen evidence type or score modification; distinguish clinical from preclinical evidence; grade actionability or replication strength; reject a plausible but non-applicable publication. &
Correct evidence selection; correct criterion, level or weight; justification based on study design and context; appropriate handling of a distractor or conflicting result. \\
3 & Multi-source synthesis and decision &
Reconcile multiple submissions or studies; classify a variant; synthesize gene--disease mechanism or lumping--splitting evidence; select or sequence a therapy; provide a PGx prescribing recommendation with evidence-strength caveats. &
Correct final synthesis or recommendation; prioritisation of directly applicable evidence; integration across sources; calibrated uncertainty and clinically important caveats. \\
\bottomrule
\end{tabularx}
\end{table}

\subsubsection{Response Generation and Rubric-Based Evaluation}

All seven conditions answered the same 426 frozen prompts with DeepSeek-V4 as the common core. Reference answers, construction rationales, PMID relevance labels and rubrics were withheld. GPT-5.5 and Claude-4.8-Opus independently scored each source-masked response against its item-specific rubric; factual errors, omissions, irrelevant evidence, over-claiming and clinically incorrect recommendations reduced credit. The complete shared judge prompt follows.

\begin{benchmarkdisplay}{Prompt used for rubric-based LLM judging}
\small
\textbf{System instruction}\par
You are an independent evaluator of answers to evidence-grounded clinical-genomics questions. Evaluate only the candidate response supplied below. Apply the item-specific scoring rubric exactly and use the reference answer as the authoritative description of the evidence available for this item. Do not infer which system produced the response and do not reward or penalise a response because of its presumed model or framework.\par

The benchmark spans variant interpretation, gene--disease validity assessment and clinical actionability, at three reasoning tiers: single-publication comprehension, evidence selection and strength grading, and multi-source synthesis. Depending on the item, a correct response may need to apply ACMG/AMP criteria, ClinGen evidence concepts, CIViC or OncoKB evidence levels, or PharmGKB/ClinPGx evidence levels. Some items deliberately contain superficially related but non-applicable evidence.\par

Follow these rules:\par
1. Score the response criterion by criterion. Award full credit only when the required factual element, evidence interpretation and requested conclusion are all present. Award partial credit only when permitted by the rubric.\par
2. Accept clinically and semantically equivalent wording. Do not require exact phrase matching with the reference answer. Do not reward verbosity, formatting, confidence or writing style.\par
3. Judge cited evidence in the context specified by the question. Evidence about a different variant, gene--disease relationship, tumour type, therapy, phenotype or population does not support the target conclusion merely because it is biologically related.\par
4. Distinguish evidence identification from evidence weighting. A response that names a relevant study but assigns an incorrect ACMG/AMP criterion, evidence strength, database level or clinical implication has not fully satisfied that rubric element.\par
5. For multi-source items, assess whether the response integrates the relevant evidence, filters non-applicable evidence, resolves conflicts appropriately and produces the requested classification, synthesis or recommendation with calibrated caveats.\par
6. Penalise fabricated numerical values, publications, study findings, criteria or clinical recommendations. Also penalise over-claiming beyond the supplied evidence. Do not use outside knowledge to add new requirements or to overturn an item-specific reference answer.\par
7. Do not deduct the same error twice unless it independently violates two explicitly separate rubric criteria. Scores must remain between zero and the maximum score stated in the rubric.\par
8. If the candidate response is incomplete, unclear or refuses to answer, award only the points supported by explicit content in the response. Do not reconstruct a missing answer on the candidate's behalf.\par

\textbf{Item}\par
Item identifier: \texttt{\{\{ITEM\_ID\}\}}\par
Clinical-genomics domain: \texttt{\{\{TASK\_DOMAIN\}\}}\par
Reasoning tier: \texttt{\{\{TIER\}\}}\par

\textbf{Question}\par
\texttt{\{\{QUESTION\}\}}\par

\textbf{Source-grounded reference answer}\par
\texttt{\{\{REFERENCE\_ANSWER\}\}}\par

\textbf{Item-specific scoring rubric}\par
\texttt{\{\{SCORING\_RUBRIC\}\}}\par

\textbf{Candidate response}\par
\texttt{\{\{CANDIDATE\_RESPONSE\}\}}\par

Return only one valid JSON object with the following structure:\par
\texttt{\{"item\_id":"...", "maximum\_score":0, "awarded\_score":0,}\par
\texttt{"criterion\_assessments":[}\par
\texttt{\{"criterion":"...",}\par
\texttt{"status":"met|partially\_met|not\_met|contradicted",}\par
\texttt{"points\_awarded":0, "reason":"..."\}\}],}\par
\texttt{"unsupported\_or\_harmful\_claims":["..."],}\par
\texttt{"score\_rationale":"..."\}}\par

The sum of \texttt{points\_awarded} must equal \texttt{awarded\_score}. Use the maximum score defined by the item-specific rubric. Use an empty list when there are no unsupported or harmful claims. Keep each reason brief and tie it to explicit content in the candidate response.
\end{benchmarkdisplay}

Let $r_{ijm}$ be the rubric points awarded to framework $m$ on item $i$ by judge $j$, and $R_i$ the item maximum. The primary rubric-normalised score was
\begin{equation}
S_m=100\times\frac{1}{N}\sum_{i=1}^{N}\frac{1}{2}\left(\frac{r_{i,\mathrm{GPT},m}}{R_i}+\frac{r_{i,\mathrm{Claude},m}}{R_i}\right).
\end{equation}
Each item therefore contributed equally, regardless of rubric length or publication count. The same mean was calculated by reasoning tier and clinical-genomics domain; it measures rubric-defined factual correctness, evidence selection and weighting, completeness and final synthesis rather than exact-match accuracy.

\subsubsection{Framework-Selection Result}

The full \ModelName{} architecture achieved the highest rubric-normalised score overall and in every reasoning tier (Extended Data Fig.~\ref{fig:ablation}c). Its overall score was 83.2\%, compared with 72.2\% for the unenhanced DeepSeek-V4 baseline and 74.0--78.4\% for the five single-strategy conditions. The largest separation occurred for tier~3 synthesis tasks. \ModelName{} scored 70.1\% on variant-interpretation items, 56.7\% on gene--disease-validity items and 84.9\% on clinical-actionability items, corresponding to improvements of 15.2, 11.1 and 15.1 percentage points, respectively, over the same base model without an enhancement layer. These controlled comparisons supported selection of the complete skill-centred architecture for the main system evaluation.

\subsection{Selection of the Frontier LLM Used as the Agent Core}
\label{sec:agent_core_selection}

\newtcolorbox{modelselectiondisplay}[1]{
  breakable,
  colback=blue!2,
  colframe=blue!45!black,
  boxrule=0.6pt,
  arc=1.5mm,
  left=2mm,right=2mm,top=1.5mm,bottom=1.5mm,
  title={#1},
  fonttitle=\bfseries
}

\subsubsection{Purpose and Experimental Design}

This experiment changed only the LLM core while holding the complete \ModelName{} architecture, skills, tools, instructions and 3,662-item input set fixed. It compared the six frontier models in Supplementary Table~\ref{tab:agent_core_models} across variant interpretation ($n=1{,}366$), gene--disease validity ($n=1{,}253$) and clinical actionability ($n=1{,}043$). Candidate cores were excluded from external judging.

\begin{table}[ht]
\centering
\footnotesize
\setlength{\tabcolsep}{4pt}
\renewcommand{\arraystretch}{1.15}
\caption{Frontier language models compared as the \ModelName{} agent core.}
\label{tab:agent_core_models}
\begin{tabularx}{\textwidth}{@{}
  >{\raggedright\arraybackslash}p{0.19\textwidth}
  >{\raggedright\arraybackslash}p{0.18\textwidth}
  >{\raggedright\arraybackslash}p{0.20\textwidth}
  X@{}}
\toprule
Agent core & Parameter scale reported for the experiment & Access category & Role in the comparison \\
\midrule
DeepSeek-V4 & 862B & Openly deployable & Candidate selected for the main \ModelName{} experiments \\
Qwen3 & 235B & Openly deployable & Open-model comparator \\
GLM-5 & 754B & Openly deployable & Open-model comparator \\
GPT-5.4-Pro & Provider not disclosed & Provider hosted & Closed-model frontier comparator \\
Claude-4.6-Opus & Provider not disclosed & Provider hosted & Closed-model frontier comparator \\
Gemini-3-Pro & Provider not disclosed & Provider hosted & Closed-model frontier comparator \\
\bottomrule
\end{tabularx}
\end{table}

\subsubsection{Source Records and Construction of the Evidence-Grounded QA Corpus}

The six pipelines used ClinVar and ClinGen-ACMG for variant interpretation, ClinGen Gene--Disease Validity for gene--disease assessment, and CIViC, OncoKB and PharmGKB/ClinPGx for clinical actionability. Every item was joined to a processed publication at the finest PMID-linked evidence unit supported by its source schema. Multi-publication narratives were treated only as trust anchors; claims and evidence codes had to be verified in the current paper, preventing cross-paper leakage or attribution to a comparator allele, disease, drug or population.

Source-specific candidate, acceptance and final-inclusion counts are given in Supplementary Table~\ref{tab:agent_core_qa_composition}. The frozen experiment contained 3,662 items: 1,366 variant-interpretation, 1,253 gene--disease-validity and 1,043 clinical-actionability items.

\begin{table}[ht]
\centering
\footnotesize
\setlength{\tabcolsep}{3.5pt}
\renewcommand{\arraystretch}{1.15}
\caption{Construction and final inclusion of evidence-grounded QA items for agent-core selection.}
\label{tab:agent_core_qa_composition}
\begin{threeparttable}
\begin{tabularx}{\textwidth}{@{}
  >{\raggedright\arraybackslash}p{0.12\textwidth}
  >{\raggedright\arraybackslash}p{0.15\textwidth}
  r r r r
  >{\raggedright\arraybackslash}X@{}}
\toprule
Domain & Source & Candidates\tnote{a} & Accepted\tnote{b} & Discarded & Selected\tnote{c} & Principal evidence join \\
\midrule
Variant interpretation & ClinVar & 1,176 & 743 & 433 & 743 & Submission--PMID \\
 & ClinGen-ACMG & 1,235 & 623 & 612 & 623 & Variant--evidence code--PMID \\
\addlinespace
Gene--disease validity & ClinGen & 2,471 & 1,440 & 1,031 & 1,253 & Assertion--evidence row--PMID \\
\addlinespace
Clinical actionability & CIViC & 408 & 294 & 114 & 294 & Evidence item--PMID \\
 & OncoKB & 524 & 222 & 302 & 222 & Association--PMID \\
 & PharmGKB/ClinPGx & 1,000 & 527 & 473 & 527 & Variant annotation--PMID \\
\midrule
Total & Six sources & 6,814 & 3,849 & 2,965 & 3,662 & \\
\bottomrule
\end{tabularx}
\begin{tablenotes}[flushleft]\footnotesize
\item[a] Candidate units processed by the adversarial QA-generation pipeline.
\item[b] Items accepted by the QA gate.
\item[c] Items included in the frozen model-selection experiment.
\item The 3,849 accepted items are the complete output of the archived adversarial QA-generation batches. The model-selection total is 3,662 because the frozen experiment used 1,253 of the 1,440 accepted ClinGen Gene--Disease Validity items. ClinVar and ClinGen-ACMG were combined for the variant-interpretation radar plot; CIViC, OncoKB and PharmGKB/ClinPGx were combined for the clinical-actionability radar plot.
\end{tablenotes}
\end{threeparttable}
\end{table}

\subsubsection{Adversarial QA Generation and Quality Control}

A source-specific generator produced a publication-grounded question, reference answer, reasoning annotation and supporting quotations. Items had to require interpretation or synthesis, and every quotation had to fuzzy-match the current full text at a similarity of at least 0.80.

Four memoryless answerers (DeepSeek-V4, Qwen3.7-Max, MiniMax-M3 and MiMo-V2.5-Pro) then answered from the same paper. A source-masked Qwen3.7-Plus arbiter from a different model family checked ambiguity, source support, overreach and semantic difficulty, returning discriminative acceptance, consensus-hard acceptance or rejection. Rejected units were regenerated for at most three quality attempts. Accepted audit records were retained but hidden from evaluated cores.

\subsubsection{Model-Facing Input, Leakage Control and Common Workflow}

Each core received the question and retrieval entry points, not full text in the initial prompt, and used the same \ModelName{} skills, tool definitions, 50-turn limit, timeout and retry policy. Runs retained the final answer, status and complete trace, including tool, skill, action, canonical parameters and result metadata.

Construction answerers, arbitration data, rejected attempts, generator rationale and target reasoning point were removed before inference; reference answers, quotations and source annotations were scoring-only. Every (model, item) pair was run independently three times without shared state, retrieved content or tool history.

\subsubsection{Five Evaluation Dimensions}

The five dimensions combined external-judge assessment of the response with judge-free analysis of the execution trace. The external judge was not one of the six candidate cores.

\paragraph{Absolute accuracy.}
Each reference answer was cached as atomic claims. The item score was 1 when all claims were supported, 0.5 when at least half were supported, and 0 otherwise; contradiction of the critical target claim forced a score of 0. Thus, accuracy was a deterministic 0/0.5/1 claim-coverage score rather than string matching.

\paragraph{Reasoning score.}
Five binary indicators assessed specific evidence use, recognition of the hidden reasoning issue, logical connection of evidence to conclusion, coverage of all requested parts and consistency between rationale and answer. Their sum ranged from 0 to 5.

\paragraph{Agentic ability.}
Three judge-free trace statistics captured execution effort and reliability: tool-call count $T$, failed-call rate $f$ and exact-duplicate rate $d$, where duplicates shared the same skill, action and canonical parameters. Values were averaged across the three runs and, for multi-source domains, across sources with equal weight. For the radar axis, each statistic was reverse min--max normalized across the six models within a domain and combined as $G_m=(Z_{m,T}+Z_{m,f}+Z_{m,d})/3$. Raw statistics were retained because call count alone was not interpreted as quality.

\paragraph{Faithfulness.}
Direct quotations were fuzzy-matched to retrieved text; other concrete claims were labelled supported, contradicted or unverifiable against retrieved evidence. With fabricated-quotation rate $q_i$ and contradicted-or-unverifiable factual-claim rate $h_i$, faithfulness was $H_i=1-0.5q_i-0.5h_i$. Errors affecting key values, evidence levels, clinical scope or conclusions were additionally flagged as high severity.

\paragraph{Robustness.}
Across three independent runs, answer consistency $C_i$ was 1 when all accuracy scores agreed, 0.5 when two agreed and 0 otherwise. Tool-sequence similarity $U_i$ was the mean pairwise Levenshtein similarity of (skill, action) sequences; agentic stability $V_i$ was one minus the mean coefficient of variation of tool, failed and duplicate-call counts, truncated at zero. The composite was $B_i=0.4C_i+0.3U_i+0.3V_i$.

\subsubsection{External Judge Prompt}

Reference claims were decomposed once per item and reused for every model and run. The source-masked prompt below separately elicits claim correctness, five reasoning indicators and factual-faithfulness labels; candidate models never received the reference-only fields.

\begin{modelselectiondisplay}{External judge prompt for agent-core evaluation}
\small
\textbf{System message}\par
You are an independent evaluator of evidence-grounded clinical-genomics answers. You are not one of the models being compared. Evaluate only against the supplied question, reference answer, hidden item annotations and retrieved source evidence. Do not reward fluency, verbosity, model identity or agreement with unsupported background knowledge. Do not infer that a claim is supported merely because it is plausible. Treat scope changes involving the variant, gene, disease, inheritance mode, population, treatment, comparator, evidence level or direction of effect as substantive errors. Return valid JSON only.\par
\medskip
\textbf{User message}\par
\texttt{<QUESTION>}\par
\{question\}\par
\texttt{</QUESTION>}\par
\texttt{<REFERENCE\_ANSWER>}\par
\{groundtruth\_answer\}\par
\texttt{</REFERENCE\_ANSWER>}\par
\texttt{<REFERENCE\_ATOMIC\_CLAIMS>}\par
\{cached atomic claims; mark the claim corresponding to the target reasoning point as critical\}\par
\texttt{</REFERENCE\_ATOMIC\_CLAIMS>}\par
\texttt{<ITEM\_ANNOTATION>}\par
Reasoning type: \{reasoning\_type\}\par
Target reasoning point: \{target\_reasoning\_point\}\par
Verified supporting quotations: \{supporting\_quotes\}\par
Source-specific evidence identifiers: \{used\_evidence\_ids\}\par
\texttt{</ITEM\_ANNOTATION>}\par
\texttt{<RETRIEVED\_SOURCE\_EVIDENCE>}\par
\{the publication text and structured evidence actually retrieved during the run\}\par
\texttt{</RETRIEVED\_SOURCE\_EVIDENCE>}\par
\texttt{<CANDIDATE\_ANSWER>}\par
\{model response\}\par
\texttt{</CANDIDATE\_ANSWER>}\par
\medskip
Perform the following checks.\par
1. For every reference atomic claim, assign exactly one label: \texttt{SUPPORTED}, \texttt{CONTRADICTED} or \texttt{MISSING}. A claim is \texttt{SUPPORTED} only when the candidate states its clinically material content with the correct entities, direction, scope and qualification. Identify whether a contradicted claim is the critical target claim. Do not assign the final 0/0.5/1 accuracy score; it will be computed deterministically.\par
2. Score each reasoning criterion as 0 or 1: specific evidence use; recognition of the annotated trap; logical validity; completeness across requested subparts; and consistency of rationale with final answer. Give a short evidence-based reason for every value.\par
3. Extract each concrete non-quoted factual claim in the candidate answer and label it \texttt{SUPPORTED}, \texttt{CONTRADICTED} or \texttt{UNVERIFIABLE} against the retrieved source evidence. Do not re-evaluate direct quotations here; they are checked separately by text matching. Mark an error as high severity if it changes a key value, evidence code or level, clinical scope, direction of effect or final conclusion.\par
4. Ignore style unless it causes ambiguity or omission. If the source evidence is insufficient to judge a claim, use \texttt{UNVERIFIABLE}; do not use general knowledge to rescue it.\par
\medskip
Return this schema:\par
\texttt{\{}\par
\texttt{"atomic\_claim\_labels": [\{"claim\_id": "C1", "label": "SUPPORTED|CONTRADICTED|MISSING", "critical": true|false, "reason": "..."\}],}\par
\texttt{"reasoning": \{"evidence\_use": \{"score": 0|1, "reason": "..."\}, "trap\_recognition": \{...\}, "logical\_validity": \{...\}, "completeness": \{...\}, "self\_consistency": \{...\}\},}\par
\texttt{"factual\_claims": [\{"claim": "...", "label": "SUPPORTED|CONTRADICTED|UNVERIFIABLE", "severity": "high|standard", "evidence": "..."\}],}\par
\texttt{"notes": "..."}\par
\texttt{\}}
\end{modelselectiondisplay}

The scoring code validated the returned schema, summed the five reasoning indicators, applied the deterministic accuracy thresholds and combined the faithfulness labels with the independent quotation-matching result. The judge was not given the candidate model name.

\subsection{Details of the MRMC Human-AI Collaboration Study}
\label{sec:mrmc_supp}

The details below supplement the main study design. The analysis unit was a submitted case assessment; reader-facing materials never contained reference answers, scoring annotations or adjudication records.

\newtcolorbox{mrmccasedisplay}[1]{
  breakable,
  colback=blue!2,
  colframe=blue!45!black,
  boxrule=0.6pt,
  arc=1.5mm,
  left=2mm,right=2mm,top=1.5mm,bottom=1.5mm,
  title={#1},
  fonttitle=\bfseries
}

\subsubsection{Dedicated Online Evaluation Platform}
\label{mrmc_platform}

The password-protected platform linked each account to nine fixed task--condition assignments and presented de-identified case materials beside the scenario-specific response form. Reference answers, scoring keys, other readers' responses and alternate-condition responses to the same case were inaccessible. Readers used no AI in the unassisted condition or consulted the assigned external DeepSeek/\ModelName{} interface before entering their own conclusion; AI output was never copied automatically.

Drafts were retained, required fields were checked on submission and oncology readers completed the open response before viewing candidate statements. Completion time was self-reported rather than inferred from browser duration. Scenario fields are listed in Supplementary Table~\ref{tab:mrmc_platform_fields}; five-level confidence was encoded 1--5 from no confidence to very confident.

\begin{table}[ht]
\centering
\footnotesize
\setlength{\tabcolsep}{4pt}
\renewcommand{\arraystretch}{1.15}
\caption{Reader-facing components of the MRMC evaluation platform.}
\label{tab:mrmc_platform_fields}
\begin{tabularx}{\textwidth}{@{}
  >{\raggedright\arraybackslash}p{0.16\textwidth}
  >{\raggedright\arraybackslash}p{0.25\textwidth}
  X@{}}
\toprule
Scenario & Information presented & Required reader entries \\
\midrule
Criterion-level variant analysis &
De-identified phenotype, variant in transcript-level notation, associated gene/disease and inheritance context, and case-specific publications when available &
For each of the 28 ACMG/AMP criteria: applicable, not applicable or undetermined; effective evidence strength (Very Strong, Strong, Moderate or Supporting) when applicable; and an optional criterion-level rationale. The reader then entered one five-tier pathogenicity classification, a five-point confidence rating, an integrated written justification and completion time. \\
Genomic interpretation in rare disease &
De-identified phenotype and a set of candidate variants, including gene, variant description, zygosity, parental origin or phase when available &
For every candidate variant: one five-tier pathogenicity category and one five-category assessment of how well the variant explained the phenotype. The reader additionally entered one case-level confidence rating, an integrated explanation covering the candidate set, and completion time. \\
Molecular oncology decision-making &
De-identified longitudinal clinical, pathological, treatment and molecular findings, followed by one case-specific decision or interpretation question &
An open-text answer and its confidence rating, followed by a separate judgement of each of four candidate statements as basically reasonable, partly reasonable or unreasonable, a second confidence rating for the statement judgements, optional explanations and completion time. \\
\bottomrule
\end{tabularx}
\end{table}

\subsubsection{Reference Standards and Expert Adjudication}
\label{primary_outcome}

Before analysis, each three-member scenario panel froze a reference record matching the reader fields: final classification, ACMG/AMP criteria and strengths for criterion-level variant analysis; candidate pathogenicity, phenotype explanation and acceptable adjacent levels for genomic interpretation in rare disease; and a core open answer plus graded candidate statements for molecular oncology decision-making. Enumerated fields were compared by prespecified mappings, while free text and borderline cases were resolved by source-masked panel consensus based on clinical equivalence rather than wording or response length.

\label{main_concordance}
\paragraph{Primary outcome: case-level overall concordance.}
For each scenario and condition, concordance was the proportion of the 45 submitted assessments satisfying the complete task-specific rule in Supplementary Table~\ref{tab:mrmc_primary_concordance}; no pooled primary outcome combined the three scenarios.

\begin{table}[ht]
\centering
\footnotesize
\setlength{\tabcolsep}{4pt}
\renewcommand{\arraystretch}{1.15}
\caption{Task-specific definition of primary case-level overall concordance in the MRMC study.}
\label{tab:mrmc_primary_concordance}
\begin{tabularx}{\textwidth}{@{}
  >{\raggedright\arraybackslash}p{0.16\textwidth}
  X
  >{\raggedright\arraybackslash}p{0.28\textwidth}@{}}
\toprule
Scenario & Requirements for a concordant case & Tolerance and discordance rules \\
\midrule
Criterion-level variant analysis &
Exact final five-tier classification; every reference ACMG/AMP criterion recovered; strength of each recovered reference criterion within one adjacent level on Supporting--Moderate--Strong--Very Strong. &
Any missing reference criterion, strength difference exceeding one level or wrong final class was discordant. Material unsupported evidence could also make the complete response discordant. \\
Genomic interpretation in rare disease &
Exact pathogenicity category for every clinically material candidate; phenotype-explanation category within the accepted range; correct causal combination and inheritance configuration; coherent integrated interpretation. &
One adjacent phenotype level was accepted only when the substantive interpretation was unchanged. Larger differences, an incorrect causal combination or incompatible narrative were discordant. \\
Molecular oncology decision-making &
Panel-adjudicated agreement of the open answer with the core conclusion and a statement profile preserving the reference clinical positions. &
One adjacent statement level was accepted only when the clinical stance was unchanged. A contradictory open answer or endorsement of a materially unsafe or unsupported action was discordant. \\
\bottomrule
\end{tabularx}
\end{table}

\subsubsection{Secondary Performance Outcomes}
\label{secondary_outcome}

Secondary outcomes isolated the structured reasoning components listed with their units and denominators in Supplementary Table~\ref{tab:mrmc_denominators}; eligible item instances were micro-averaged within condition.

For criterion-level variant analysis, evidence recall was $\sum_i|G_i\cap P_i|/\sum_i|G_i|$, unsupported-evidence assignment was $\sum_i|P_i\setminus G_i|/\sum_i|P_i|$, and strength concordance required an exact strength match among recalled reference criteria. The second metric is labelled the evidence hallucination rate in the figure and code, but denotes an unsupported ACMG/AMP criterion rather than a fabricated citation. Final-class accuracy and directional over-/underestimation used the ordered five-tier pathogenicity scale; the reference contained 176 criterion instances.

For genomic interpretation in rare disease, the 117 candidate variants were evaluated for exact pathogenicity, directional error and phenotype-explanation accuracy. Phenotype accuracy allowed an accepted adjacent level only when the substantive interpretation was preserved; over-/underestimation was measured beyond this accepted range. Primary concordance additionally required the correct causal combination and inheritance interpretation.

For molecular oncology decision-making, free-response accuracy used panel-adjudicated agreement with the core answer across 45 cases. The four statements per case produced 180 judgements; accuracy required the reference clinical position, while over-/underestimation followed the ordered unreasonable--partly reasonable--basically reasonable scale and did not tolerate a supported--unsupported boundary crossing.

\subsubsection{Confidence--Correctness Alignment, Time, Survey and Continued-Use Outcomes}
\label{confidence_safe}

For criterion-level variant analysis, behavioural certainty was \emph{full} when all 28 ACMG/AMP criteria had an explicit applicable/not-applicable call and \emph{partial} when any remained undetermined. Group accuracy used exact final pathogenicity classification, not the stricter primary composite; Wilson intervals were descriptive because fully certain groups were small.

For genomic interpretation in rare disease, correctness was primary case-level concordance; for molecular oncology decision-making, it was correctness of the open answer entered immediately before confidence. Five-level confidence was related to correctness using point-biserial correlation and ROC--AUC. Positive correlation and AUC above 0.5 indicate discrimination.

Confidence categories were mapped to probabilities 0.1, 0.3, 0.5, 0.7 and 0.9 for the Brier score, $N^{-1}\sum_i(q_i-Y_i)^2$; correlation and AUC retained the original ordinal scores. Expected calibration error was descriptive, with no model fitting or threshold optimization.

The main time comparison used each reader's post-study estimate of mean minutes per case, not browser-open duration or the case-form field. Paired two-sided Wilcoxon tests within scenario were Holm-adjusted, and medians were reported.

The questionnaire comprised three time items and five ordinal dimensions rated separately for DeepSeek and \ModelName{}; ratings were summarized by mean and standard deviation. Continued use was the number of new \ModelName{} conversations initiated in the 14 days after completion, excluding three assigned study sessions, and was analysed descriptively without hypothesis testing.

\subsubsection{Statistical Analysis of Proportions and Effect Differences}

\begin{table}[!h]
\centering
\footnotesize
\setlength{\tabcolsep}{4pt}
\renewcommand{\arraystretch}{1.15}
\caption{Analysis units and denominators for secondary MRMC performance outcomes.}
\label{tab:mrmc_denominators}
\begin{tabularx}{\textwidth}{@{}
  >{\raggedright\arraybackslash}p{0.18\textwidth}
  >{\raggedright\arraybackslash}p{0.25\textwidth}
  >{\raggedright\arraybackslash}p{0.18\textwidth}
  X@{}}
\toprule
Scenario & Outcome & Analysis unit per condition & Denominator \\
\midrule
Criterion-level variant analysis & Evidence recall & Reference criterion instance & 176 reference criteria \\
 & Evidence hallucination rate & Submitted applicable criterion instance & Number called applicable in that condition \\
 & Evidence-strength concordance & Correctly recalled reference criterion & Number recalled in that condition \\
 & Pathogenicity accuracy & Complete submitted case & 45 \\
 & Pathogenicity underestimation rate & Complete submitted case & 45 \\
 & Pathogenicity overestimation rate & Complete submitted case & 45 \\
Genomic interpretation in rare disease & Pathogenicity accuracy & Candidate variant & 117 \\
 & Pathogenicity overestimation rate & Candidate variant & 117 \\
 & Pathogenicity underestimation rate & Candidate variant & 117 \\
 & Phenotype-explanation accuracy & Candidate variant & 117 \\
 & Explanation-degree overestimation rate & Candidate variant & 117 \\
 & Explanation-degree underestimation rate & Candidate variant & 117 \\
Molecular oncology decision-making & Free-response accuracy & Complete open answer & 45 \\
 & Reasonableness-judgement accuracy & Candidate statement & 180 (four per case) \\
 & Reasonableness overestimation rate & Candidate statement & 180 (four per case) \\
 & Reasonableness underestimation rate & Candidate statement & 180 (four per case) \\
\bottomrule
\end{tabularx}
\end{table}
\FloatBarrier

Observed numerators and denominators were retained for descriptive reporting. For each scenario and outcome, condition effects were estimated using a binomial logistic GLMM with condition as a fixed effect and crossed random intercepts for reader and case. Case-level binary outcomes were entered as success or failure; item-level outcomes were grouped within each reader--case assessment as \texttt{cbind(success, total-success)}, thereby retaining their observed denominators. Accuracy, overestimation and underestimation were analysed as separate binary outcomes. Model-standardized probabilities and absolute between-condition differences in percentage points were obtained by integrating over the estimated reader- and case-level random-effect distributions, with two-sided 95\% Wald confidence intervals. For outcomes where lower values were favourable, figures show one minus the model-standardized event probability and the direction of each contrast was reversed. Pairwise Wald contrasts were Holm-adjusted across the three comparisons within each outcome and scenario; the corresponding two-degree-of-freedom omnibus Wald test was descriptive only. Random slopes were not fitted because each reader contributed only three assessments per condition within a scenario.

\subsubsection{Representative Complete Reader-Facing Tasks}
\label{mrmc_example}

The following examples reproduce the structure and level of information presented to readers. They are shown without participant answers and without the hidden expert scoring record.

\begin{mrmccasedisplay}{Representative MRMC task 1: criterion-level variant analysis}
\small
\textbf{Clinical vignette.} A three-year-old boy presented with pallor, failure to gain weight, intermittent vomiting and fever. Weight and height were 10\,kg ($-2.66$ s.d.) and 85\,cm ($-2.57$ s.d.), respectively. Prenatal ultrasonography at 34 weeks had shown enlarged kidneys. Postnatal imaging showed bilaterally increased renal echogenicity with multiple small cysts. At admission, ultrasonography showed cholelithiasis, mild left calyceal dilatation, bilateral renal cysts and renal-parenchymal abnormalities. Blood pressure was 160/100\,mmHg, haemoglobin 5.4\,g\,dl$^{-1}$, serum creatinine 4.5\,mg\,dl$^{-1}$ and estimated glomerular filtration rate 7.8\,ml\,min$^{-1}$\,1.73\,m$^{-2}$, with metabolic acidosis. The evaluated variant was heterozygous; biallelic \textit{INVS} variants are associated with autosomal-recessive nephronophthisis.

\textbf{Variant.} An anonymized heterozygous \textit{INVS} variant.

\textbf{Material available.} The vignette, variant and one case-associated publication supplied as a PDF; ordinary literature and database searches were permitted under the condition-specific rules.

\textbf{Reader task.} For each of PVS1, PS1--PS4, PM1--PM6, PP1--PP5, BA1, BS1--BS4 and BP1--BP7, select \emph{applicable}, \emph{not applicable} or \emph{undetermined}; when applicable, select Very Strong, Strong, Moderate or Supporting and optionally state the evidence. Then provide the overall five-tier pathogenicity classification, confidence, an integrated justification and completion time.
\end{mrmccasedisplay}

\begin{mrmccasedisplay}{Representative MRMC task 2: genomic interpretation in rare disease}
\small
\textbf{Clinical vignette.} A female infant developed progressive heart failure shortly after birth. Cardiac imaging showed chamber enlargement and rapidly worsening systolic function, and she died in early infancy. A sibling had an almost identical early-onset course.

\textbf{Candidate set.}
\begin{enumerate}
 \item An anonymized heterozygous \textit{ASNA1/GET3} variant; parental origin not established.
  \item An anonymized heterozygous paternal \textit{ASNA1/GET3} variant; part of a possible compound-heterozygous configuration with candidate~3.
  \item An anonymized heterozygous maternal \textit{ASNA1/GET3} variant; part of the same possible compound-heterozygous configuration.
\end{enumerate}

\textbf{Reader task.} For each candidate, select one pathogenicity category (Benign, Likely Benign, Uncertain Significance, Likely Pathogenic or Pathogenic) and one phenotype-explanation category (fully explains, possibly explains, partly explains, no clear evidence or clearly cannot explain). Then provide one confidence rating, an integrated explanation addressing the candidate combination, phenotype and inheritance pattern, and total completion time.
\end{mrmccasedisplay}

\begin{mrmccasedisplay}{Representative MRMC task 3: molecular oncology decision-making}
\small
\textbf{Clinical vignette.} A 66-year-old never-smoking woman had a 44-mm left-upper-lobe lung adenocarcinoma with pleural effusion, pleural thickening and a solitary temporal-lobe brain metastasis (cT2bN0M1b, stage IVA). The tumour carried \textit{EGFR} L858R and had PD-L1 TPS $<1\%$. After stereotactic treatment of the brain metastasis, first-line osimertinib produced a partial response and prolonged disease control. At 46 months, the lung primary enlarged and became metabolically active while the previous pleural and brain sites remained controlled. The isolated progressing lesion was resected. The specimen contained adenocarcinoma plus new giant-cell and spindle-cell components, with reduced TTF-1, increased vimentin and a diagnosis of pleomorphic carcinoma. \textit{EGFR} L858R remained detectable and PD-L1 TPS was $>75\%$. No measurable progressing disease remained after surgery.

\textbf{Open question.} How should systemic treatment be adjusted after surgery?

\textbf{Candidate statements rated after the open response.}
\begin{enumerate}
\item Continue osimertinib and add platinum-based chemotherapy because pleomorphic transformation and high PD-L1 suggest resistant micrometastatic disease requiring chemotherapy coverage.
\item Continue osimertinib and add a PD-1/PD-L1 inhibitor because pleomorphic transformation and TPS $>75\%$ suggest increased immunogenicity.
\item Stop osimertinib and use PD-1/PD-L1 inhibitor monotherapy because high PD-L1 and pleomorphic transformation indicate systemic EGFR-TKI resistance.
\item Continue osimertinib alone after local resection, without adding or switching immunotherapy, because \textit{EGFR} L858R was retained and progression was confined to a single resectable lesion.
\end{enumerate}

\textbf{Reader task.} First submit an open recommendation and its confidence rating. Then rate each statement as basically reasonable, partly reasonable or unreasonable, optionally explain the ratings, give confidence in the statement judgements and report total completion time.
\end{mrmccasedisplay}

\subsubsection{Construction and Quality Control of the Molecular Oncology Tasks}
\label{mrmc_oncology_build}

The 323 human molecular-oncology case reports identified in the main Methods---186 digestive-system and 137 thoracic-tumour reports published after 2025---were processed independently before expert selection. GPT-5.5 read each complete report and returned structured fields comprising the clinically decisive or counterintuitive features, their clinical implication, a de-identified vignette, one direct question, a concise core answer and explicit scoring points. Vignettes retained the clinical, pathological, treatment and molecular facts needed to answer the question, but omitted article titles, authors, journals, identifiers and source-revealing wording.

Five independent models---DeepSeek-V4, Claude Sonnet~4.6, GPT-5.4, GLM-5.1 and Qwen3.7---then answered each vignette and question in fresh contexts without access to the report, reference answer or other model responses. Claude Opus~4.8 subsequently assessed whether the reference was source-grounded and clinically defensible; whether the vignette was self-contained and free of answer or publication leakage; whether the question was single, direct and appropriate for a practising oncologist; and whether the simulated responses demonstrated useful discriminatory difficulty. Failed items were regenerated by GPT-5.5 using the arbitrator's written feedback and rerun through independent answering and arbitration, for at most five generation rounds. Reports without an acceptable task were discarded.

For each accepted open question, the reference conclusion supplied the best-supported of four candidate statements, while incorrect or partially correct model responses were used to construct clinically plausible alternatives. Each statement received a hidden five-point truth grade: grades 4--5 mapped to \emph{basically reasonable}, grade 3 to \emph{partly reasonable}, and grades 1--2 to \emph{unreasonable}. Statement order was randomized. A final validation pass required consistency between the core answer and statement grades, exactly one best-supported statement, defensible distractors and no superficial answer cue from length, tone, certainty or absolute wording. The reader-facing task contained only the vignette, question and statements; source reports, scoring points, truth grades, simulated answers and revision records were retained separately for audit and adjudication.

The automated pipeline produced 36 digestive-system and 26 thoracic-tumour candidates. A three-member oncology panel completed and reviewed every candidate for clinical relevance, objectivity, answer defensibility, difficulty, actionability, leakage risk and the plausibility and grading of all statements. The panel retained 27 digestive-system and 18 thoracic-tumour tasks, yielding the final 45, and could correct the core answer, statement wording or reference grades before the scoring record was frozen. Finally, the selected vignettes were manually de-identified: nonessential details and publication-revealing phrases were removed or reformulated, but no fact required to answer the clinical question was deleted.

\subsection{Feedback Coverage in the Prospective Deployment}
\label{sec:deployment_feedback_coverage}

To assess the completeness of episode-level feedback, we counted all platform conversations recorded for each participant during the deployment period. A conversation was one platform session and could contain multiple dialogue turns. Feedback was submitted per completed analysis episode, and one episode could span multiple conversations. To preserve participant privacy, we did not inspect conversation content and therefore could not determine whether conversations without associated feedback were outside the study scope or represented eligible episodes for which feedback was not submitted. We therefore treated the ratio of feedback submissions to conversations as a conservative lower-bound proxy for episode-level feedback coverage, rather than as a one-to-one response rate. Overall, 96 feedback submissions were recorded for 111 conversations (86.5\%); participant-level ratios ranged from 66.7\% to 100\% (Supplementary Table~\ref{tab:deployment_feedback_coverage}). This high lower-bound coverage was consistent with participants generally following the requested feedback procedure.

\begin{table}[ht]
\centering
\footnotesize
\setlength{\tabcolsep}{7pt}
\renewcommand{\arraystretch}{1.15}
\caption{Conversation counts and feedback coverage during the prospective deployment.}
\label{tab:deployment_feedback_coverage}
\begin{tabular}{lrrr}
\toprule
Participant & Conversations & Feedback submissions & Conservative coverage \\
\midrule
User 1 & 9 & 7 & 77.8\% \\
User 2 & 9 & 9 & 100\% \\
User 3 & 12 & 9 & 75.0\% \\
User 4 & 12 & 11 & 91.7\% \\
User 5 & 7 & 7 & 100\% \\
User 6 & 13 & 10 & 76.9\% \\
User 7 & 8 & 6 & 75.0\% \\
User 8 & 11 & 10 & 90.9\% \\
User 9 & 16 & 16 & 100\% \\
User 10 & 9 & 6 & 66.7\% \\
User 11 & 5 & 5 & 100\% \\
\midrule
Total & 111 & 96 & 86.5\% \\
\bottomrule
\end{tabular}
\end{table}

\end{document}